\documentclass{article}

\usepackage[preprint]{neurips_2024}

\usepackage[utf8]{inputenc} 
\usepackage[T1]{fontenc}    
\usepackage{hyperref}       
\usepackage{url}            
\usepackage{booktabs}       
\usepackage{amsfonts}       
\usepackage{nicefrac}       
\usepackage{microtype}      
\usepackage{xcolor}         

\usepackage[normalem]{ulem}
\usepackage{cancel}
\usepackage{enumerate}
\usepackage{paralist}
\usepackage{algorithm, algpseudocode}

\usepackage{amssymb}
\usepackage{amsthm}
\usepackage{amsmath}
\usepackage{mathtools}
\usepackage{topsection}
\usepackage[title]{appendix}
\usepackage{multirow}
\usepackage{graphicx,wrapfig,lipsum}

\PassOptionsToPackage{round}{natbib}

\theoremstyle{plain}
\newtheorem{thm}{Theorem}
\newtheorem*{thm*}{Theorem}
\theoremstyle{definition}
\newtheorem{dfn}{Definition}

\newcommand{\argmax}{\mathop{\rm arg~max}\limits}
\newcommand{\argmin}{\mathop{\rm arg~min}\limits}

\title{Robust off-policy Reinforcement Learning via Soft Constrained Adversary}

\author{%
  Kosuke Nakanishi\thanks{Kyoto University} \thanks{Honda R\&D Corp.} \\
  \And
  Akihiro Kubo\footnotemark[1]\\
  \And
  Yuji Yasui\footnotemark[2]\\
  \And
  Shin Ishii\footnotemark[1]
}

\begin{document}

\maketitle

\begin{abstract}

Recently, robust reinforcement learning (RL) methods against input observation have garnered significant attention and undergone rapid evolution due to RL's potential vulnerability. 
Although these advanced methods have achieved reasonable success, 
there have been two limitations when considering adversary in terms of long-term horizons. 
First, the mutual dependency between the policy and its corresponding optimal adversary limits the development of off-policy RL algorithms; 
although obtaining optimal adversary should depend on the current policy, 
this has restricted applications to off-policy RL. Second, these methods generally assume perturbations based only on the $L_p$-norm, 
even when prior knowledge of the perturbation distribution in the environment is available. 
We here introduce another perspective on adversarial RL: an f-divergence constrained problem with the prior knowledge distribution. 
From this, we derive two typical attacks and their corresponding robust learning frameworks. 
The evaluation of robustness is conducted and the results demonstrate that our proposed methods achieve excellent performance in sample-efficient off-policy RL. 
\end{abstract}

\section{Introduction}

In recent years, advancements in computational technology, coupled with the practical successes
of deep neural networks (DNNs) \citep{krizhevsky2012imagenet, simonyan2014vgg, he2016deep}, 
have fueled expectations for automated decision-making and control in increasingly complex environments \citep{kober2013reinforcement, levine2016end, kiran2021deep}. 
Deep reinforcement learning (DRL) is a promising framework for such applications, demonstrating performance that surpasses human
capabilities by acquiring high-dimensional representational power through function approximation
\citep{mnih2015human, silver2017mastering}. 
However, in real-world applications, significant performance degradation in control due to
adverse perturbations raises practical concerns \citep{huang2017adversarial, lin2017tactics}. Therefore, the development and testing of
algorithms that consider such challenges are crucial.

Recent research \citep{zhang2021robust,sun2021strongest,liang2022efficient} identifies 
there are two types of vulnerabilities in DRL. 
The first is related to the smoothness of the policy function, 
which primarily arises from the function approximation properties of DNNs. 
The second vulnerability stems from the dynamics of the environment and is considered within the framework of Markov Decision Processes (MDPs). 
To understand the latter case, imagine a situation where you are going to cross over a deep valley and there are two bridges. 
The one is short length but a narrow bridge and the other is a large bridge but has a little bit longer path.
If you have clear vision, you may prefer the former, but if you get noisy vision in foggy conditions, 
the latter path is the best choice to achieve your goals without the risk of falling.
To realize such comprehensive decision makings, 
we need to consider long-term reward appropriately into adversaries and robustness for an RL problem, 
rather than the temporal DNN smoothness or consistency of output as in the supervised learning. 

\cite{zhang2021robust,sun2021strongest} theoretically prove that optimal (worst-case) adversaries for a policy, 
capable of estimating long-term horizons, can be learned as an RL agent. 
\cite{zhang2021robust} propose an approach where the (victim) policy and the corresponding (optimal) adversary are trained alternately, 
resulting in a robust agent. 
They refer to this framework as \textit{Alternating Training with Learned Adversaries} (ATLA). 
\cite{sun2021strongest} extend this framework to high-dimensional state spaces and strong attacks by dividing the optimal adversary into two parts: searching for a perturbing direction in policy space (Director) and crafting a perturbation in the state space through numerical calculation (Actor), 
referring as \textit{Policy Adversarial Actor Director} (PA-AD).

These methods establish robust RL frameworks on state observations with a solid foundation; however, two main issues remain unresolved.
The first issue lies in the algorithms for continuous action spaces, which rely on on-policy Actor-Critic algorithms, 
making them unable to be applied to recently developed off-policy algorithms with good sample-efficiency and performance. 
To learn optimal adversary, \cite{zhang2021robust,sun2021strongest} use the policy's trajectories under the corresponding optimal adversary, 
but during training phase, RL's policy is updated constantly, resulting in the fact the gathered trajectories cannot be used for the adversary training. 
The second point is that these robust learning methods assume perturbations under the constraint of $L_p$-norm ball in the worst-case scenario, 
making it difficult to consider phenomena such as Gaussian noise commonly assumed in natural sciences and engineering. 
While Gaussian noise could be considered across the entire state space, how should we consider adversarial constraints?

In this study, we treat the search for the optimal adversary as an $f$-divergence constrained optimization problem, 
informed by the prior knowledge of perturbation distribution. 
From this, we derive two typical adversarial attacks: the Soft Worst-Case Attack (SofA) and the Epsilon Wosrt-Case (EpsA). 
To consider updates of the action-value function and the policy under such adversarial conditions, 
we have developed theoretically sound algorithms, from the perspective of contractions and policy improvements.
At the conclusion of this introduction, we outline our contributions, which are twofold:
by introducing $f$-divergence constrained methods, 
\textbf{(1)} we expand the application of robust DRL methods to recent innovative off-policy Actor-Critic algorithms, 
with a particular focus on the vulnerability that arises from MDPs, 
and \textbf{(2)} we introduce more arbitrary and realistic adversaries than those typically constrained by the conventional $L_{\infty}$-norm.

\section{Related Work}
\label{secRelatedWork}

\subsection{Adversarial Attack and Defense on State Observations}
\label{subsecRelatedWork_SO}


Building on a seminal work by \cite{goodfellow2014explaining}, 
there has been a surge of research activity in the field of supervised learning, 
particularly focusing on various adversarial attacks and corresponding defense methods \citep{kurakin2016adversarial,papernot2016limitations,papernot2017practical,carlini2017adversarial,ilyas2018black}. 
In the context of RL, 
\cite{huang2017adversarial} demonstrated that similar challenges could arise from small perturbations, 
such as the Fast Gradient Sign Method (FGSM). 
This study led to the early proposal of various attacks on observations and corresponding robust methods \citep{kos2017delving,behzadan2017whatever,mandlekar2017adversarially,pattanaik2017robust}.

Recently, a line of research has focused on maintaining the consistency (smoothness) of the agent's policy to acquire robustness against observation perturbations \citep{zhang2020robust,shen2020deep,oikarinen2021robust,sun2024tactics}. 
\cite{zhang2020robust} first defined the adversarial observation problem as \textit{State Adversarial Markov Decision Processes} (SA-MDPs). 
They proved and demonstrated that the loss in performance due to adversarial perturbations could be bounded by the consistency of the policy. 
However, in this study, 
they did not show practical methods to create the adversary that could estimate the long-term reward assumed in the SA-MDPs. 
Due to this gap, the proposed robust methods were not robust enough against stronger attacks \citep{zhang2021robust}.

To address this issue, 
\cite{zhang2021robust} proposed that the optimal (worst-case) adversary for the policy could be learned as a DRL agent. 
The policy, learned alternatively with such an adversary, can become robust against strong attacks (ATLA).
Building on the ATLA framework, 
\cite{sun2021strongest} suggested dividing the adversary into two parts: 
searching for the mislead direction in the action space for the policy and calculating the actual perturbations in the state space through numerical approximation (PA-AD). 
This makes the adversary capable of handling high-dimensional state problems (such as Atari), 
which are difficult to address in the ATLA framework. 
These frameworks provide practical methods for on-policy algorithms (PPO, A2C), 
but applications for off-policy algorithms were not shown. 
This is because the adversarial attacker learns from trajectories generated by the (current) fixed policy, 
but the policy is constantly updated during training.

\cite{liang2022efficient} introduced an additional worst-case action-value function to improve the policy's robustness by utilizing a mixture with the original value function, 
referred to as \textit{Worst-case-aware Robust} RL (WocaR-RL). 
This approach computes the worst-case action-value through convex relaxation and heuristic gradient iterations, 
thereby omitting additional environment steps, 
unlike ATLA-based methods (ATLA, PA-ATLA). 
WocaR-RL demonstrated effectiveness in high-dimensional discrete action domains using off-policy algorithms (e.g., DQN). 
However, applications for off-policy algorithms in continuous action domains were not shown and remain unknown.

As a separate line of research, there are studies that address the ratio and temporal strategies of attacks as a multi-agent problem \citep{lin2017tactics,gleave2019adversarial,sun2020stealthy,liu2023rethinking,franzmeyer2024illusory}. 
However, these studies focus on the objectives of time-step efficiency or the stealthiness of attacks. 
This motivation differs from our approach, which limits the attacker based on constraints from an assumed distribution.

\section{Preliminaries and Background}
\label{secPreliminaries}

\paragraph{Notations}
We describe the environment using a Markov Decision Process (MDP) characterized by parameters $\langle\mathcal{S,A,F},r,\gamma,\mathcal{S}_0\rangle$, where $\mathcal{S}$ is the state space, 
$\mathcal{A}$ is the action space, 
$\mathcal{F}:\mathcal{S}\times\mathcal{A}\rightarrow\Delta(\mathcal{S})$ defines the environment's transition probabilities, where $\Delta(\mathcal{S})$ denotes the set of probability distributions over the state space $\mathcal{S}$, 
$r$ is the reward function, $\gamma$ is the discount factor, 
and $\mathcal{S}_0$ is the set of initial states.
In RL framework, 
the objective is to find a policy $\pi(a|s):\mathcal{S}\rightarrow\Delta(\mathcal{A})$ 
that maximizes the cumulative discounted reward along the trajectory, 
$\mathbb{\max}_{\pi}\mathbb{E}_{\mathcal{S}_0,\pi,\mathcal{F}}\left[\sum_{t=0}^{\infty}\gamma^{t} r(s_t, a_t)\right]$. 
To reduce the variance in episodic trajectory estimates, 
RL maintains an action-value function $Q^{\pi}(s_t,a_t)$ and/or a state-value function $V^\pi(s_t)$.

\subsection{Max-Entropy Off-Policy Actor Critic Algorithm}
\label{subsecPreliOff}
In this study, we utilize the Soft Actor Critic (SAC) \citep{haarnoja2018softa,haarnoja2018softb} as our base algorithm.
SAC is one of the most popular off-policy RL methods due to its theoretically sound foundations, 
sample-efficient, excellent performance, and simplicity.
SAC assume a modified reward function:
\begin{equation}
\label{eq_sac_reward}
\hat{r}(s_t,a_t) \triangleq r(s_t,a_t) 
+ \mathbb{E}_{s_{t+1}\sim\mathcal{F}}
\lbrack \alpha_{ent} \mathcal{H}(\pi(\cdot|s_{t+1})) \rbrack,
\end{equation}
where $\mathcal{H}(\pi(\cdot|s_{t+1}))$ represents the entropy term of the policy $\pi$ for the state at time-step $t+1$, 
and $\alpha_{ent}$ is an entropy coefficient to balance obtaining the original reward and encouraging exploration of actions.

Then, policy evaluation and improvement for $\pi$ and $Q^{\pi}$ are done as:
\footnotesize
\begin{equation}
\label{eq_sac_bellman}
Q^{\pi}(s_t,a_t) = r(s_t,a_t)
+\gamma {\mathbb{E}}_{s_{t+1}\sim \mathcal{F}}
\left[
  {\mathbb{E}}_{a_{t+1}\sim\pi}
  \lbrack
  Q^{\pi}(s_{t+1},a_{t+1}) - \alpha_{ent} \log \pi(a_{t+1}|s_{t+1})
  \rbrack
\right],
\end{equation}
\normalsize
\footnotesize
\begin{equation}
\label{eq_sac_policy_loss1}
L(\pi) =
{\mathbb{E}}_{s_t \sim D(\cdot)}
\left[
D_{KL}(\pi(\cdot|s_t) \parallel   \frac{\exp(Q^{\pi}(s_t,\cdot)/\alpha_{ent})}{
\int_{a_t} \exp(Q^{\pi}(s_t,a_t)/\alpha_{ent}) da_t
})
\right],
\end{equation}
\normalsize
where $D_{KL}$ denotes Kullback-Leibler Divergence, and $D(\cdot)$ represents batch data from the replay buffer.
By ignoring the constant term, we derive the final loss for the policy:
\footnotesize
\begin{equation}
\label{eq_sac_policy_loss2}
L(\pi) =
{\mathbb{E}}_{s_t \sim D(\cdot)}
\left[
{\mathbb{E}}_{a_t \sim \pi}
\left[
\alpha_{ent}\log{\pi(a_t|s_t)} - Q^{\pi}(s_t,a_t)
\right]
\right].
\end{equation}
\normalsize

\subsection{Reinforcement Learning under Adversarial Attack on State Observation}
\label{subsecPreliBasis2}

In scenarios with noisy observations, 
we consider adversarial perturbations $\nu(\tilde{s}_t|s_t) \in \mathcal{N}$, 
where $\mathcal{N}:\mathcal{S}\rightarrow\Delta(\mathcal{S})$ represents all possible perturbation functions 
mapping true state $s_t$ to the set of probability over the state space $\mathcal{S}$. 
The perturbation occurs at each time step $t$, 
misleading the agent's policy to output action $\tilde{a}_t$, 
while the environment dynamics transition to state $s_{t+1}$ based on $s_t$ and $\tilde{a}_t$. 
Crucially, \textbf{only the policy is deceived by the perturbation $\nu(\tilde{s}_t|s_t)$}, 
altering its action choice from $a_t \sim \pi(\cdot|s_t)$ to $\tilde{a}_t \sim \pi(\cdot|\tilde{s}_t)$. 
Previous research \citep{pattanaik2017robust,zhang2020robust,zhang2021robust,sun2021strongest,oikarinen2021robust,liang2022efficient} limits adversarial strength using an $L_p$-norm constraint (typically $p=\infty$), 
defining a restricted perturbation subset $\mathcal{B}_{\epsilon_p}(\nu;s_t) \subset \mathcal{N}$.
\begin{equation}
\label{eq_lp_norm_set}
\mathcal{B}_{\epsilon_p}(\nu;s_t) := \{\nu_{\epsilon_p} \in \mathcal{N}: \nu_{\epsilon_p}(\tilde{s}_t|s_{t})=0 \text{ if } |\tilde{s}_t - s_t|_{p} > \epsilon, \text{ other } \nu_{\epsilon_p}(\tilde{s}_t|s_{t})\geq0\}.
\end{equation}
If the expected action-value function is learned for the policy $\pi$ and the policy is once fixed,
then we can regard the optimal attacker problem for the policy as:
\begin{equation}
\label{eq_lp_norm_attack}
\nu^{\star}_{\pi}(\tilde{s_t}|s_t) = \argmin_{
  \nu \in \mathcal{B}_{\epsilon_{p}}(\nu;s_t)
  }{
  \mathop{\mathbb{E}}_{
    \tilde{s}_t \sim\nu(\cdot|s_t)
  }
  \lbrack
  \mathop{\mathbb{E}}_{
    \tilde{a}_t \sim\pi(\cdot|\tilde{s}_t)
    }
  \lbrack
  Q^{ \tilde{\pi}}(s_t,\tilde{a}_t)}
  \rbrack
  \rbrack,
\end{equation}
where $Q^{\tilde{\pi}}(s,a):=\mathbb{E}_{\mathcal{F}, \tilde{a}_t \sim \pi\circ\nu}\left[ \sum_{t=0}^{\infty}\gamma^t r(s_t,\tilde{a}_t) | s_0 =s, a_0 = a \right]$ 
is the action-value function learned under the corresponding perturbation $\nu$. 
We will reconsider this restriction $\mathcal{B}_{\epsilon_p}(\nu;s_t)$ as the soft constrained problem in the next section.

\section{Methodology}
\label{secMethodology}

\subsection{Soft Constrained Representation of Adversarial Attack on State Observation}
\label{subsecRepSofA}

To accommodate more flexible perturbation scenarios, we revisit Eq. \eqref{eq_lp_norm_attack}, 
assuming prior knowledge of the perturbation distribution $p(\tilde{s}_t|s_t) \in \mathcal{N}$ in our target environment.
We add a mild assumption that the adversarial attacker $\nu(\tilde{s}_t|s_t)$ aligns with the prior distribution
\footnote[1]{This assumption is considered mild as it encompasses conventional $L_{p}$-norm constraints when $p(\tilde{s}_t|s_t)$ is defined as a uniform distribution within the $L_{p}$-norm ball.},
$\forall s_t,\tilde{s}_t \in \mathcal{S},\text{ if } \nu(\tilde{s}_t|s_t) > 0 \Rightarrow p(\tilde{s}_t|s_t) > 0$.
This framework allows us to define the soft constrained optimal adversary:
\begin{dfn}[Soft Constrained Optimal Adversary on State Observation]
\label{dfn_soft_constrained_adversary}
\begin{equation}
\label{eq_soft_attack_def}
\nu^{\star}_{\pi}(\tilde{s}_t|s_t) = \argmin_{
    \nu \in \mathcal{N}
  }
  {
  \mathbb{E}_{
    \tilde{s}_t \sim \nu
  }
  \left[
  \mathop{\mathbb{E}}_{
    \tilde{a}_t \sim \pi(\cdot|\tilde{s}_t)
  }
  \left[
  Q^{\tilde{\pi}}(s_t,\tilde{a}_t)
  \right]
  \right]
  + \alpha_{attk} D_{f}(\nu(\cdot|s_t)\parallel p(\cdot|s_t))
}.
\end{equation}
\end{dfn}
Here, $D_{f}(\nu\parallel p)$ represents the general $f$-divergence between the perturbation distribution $\nu(\tilde{s}_t|s_t)$ and the prior distribution $p(\tilde{s}_t|s_t)$.
$\alpha_{attk}$ is the coefficient term used to balance the worst action value and the constraints on the attacker distribution imposed by the prior knowledge distribution $p(\tilde{s}_t|s_t)$.

As discussed in \cite{sun2021strongest}, 
using $Q^{\tilde{\pi}}(s_t, \tilde{a}_t)$ and $Q^{\pi}(s_t, \tilde{a}_t)$ in Eq. \eqref{eq_soft_attack_def} differ from a strict perspective. 
$Q^{\tilde{\pi}}(s_t, \tilde{a}_t)$ can estimate the sequential effect of perturbation $\nu$, 
while $Q^{\pi}(s_t, \tilde{a}_t)$ only estimates the one-step influence from $\nu$. 
Then, using $Q^{\tilde{\pi}}(s_t, \tilde{a}_t)$ results in a stronger attacker by correctly estimating the long-term effects. 
However, from the attacker’s perspective, 
it is difficult to specify whether DRL methods account for robust frameworks or not. 
Therefore, we widely use $Q^{\pi}(s_t, \tilde{a}_t)$, even though it does not account for perturbations, for practical applicability.

Although various attackers can be characterized by specifying different $f$-divergences, 
in this study, 
we specifically propose two typical attacks, each detailed in subsequent subsections.

\subsubsection{Soft Worst Attack (SofA) Sampling Method for the KL-divergence Constraint}
\label{subsubsec_softa_attacker}

When we set the $f$-divergence to KL-divergence, 
the optimal attacker for a fixed policy $\pi$ and the corresponding action-value function $Q^{\pi}$ 
can be derived by the Fenchel-Legendre transform 
(detailed in Appendix \ref{subsecApp_softa_deriviation}) as follows:
\begin{equation}
\label{eq_optimal_softa}
\nu^{\star soft}_{\pi}(\tilde{s_t}|s_t) =
  \frac{
  p(\tilde{s}_t|s_t)
  \exp{(
    \mathop{\mathbb{E}}_{
      \tilde{a}_t \sim\pi(\cdot|\tilde{s}_t)
      }
      \lbrack
      -Q^{\pi}(s_t,\tilde{a}_t)/\alpha_{attk}
      \rbrack
    )}
  }{
  \int_{\tilde{s}_t}
  p(\tilde{s}_t|s_t)
  \exp{(
    \mathop{\mathbb{E}}_{
      \tilde{a}_t \sim\pi(\cdot|\tilde{s}_t)
      }
      \lbrack
      -Q^{\pi}(s_t,\tilde{a}_t)/\alpha_{attk}
      \rbrack
    )} d\tilde{s}_t
  }.
\end{equation}
When dealing with continuous state and action spaces, direct access to this distribution is not possible.
We can approximate this distribution using Markov Chain Monte Carlo (MCMC) or variational inference method; 
however, these methods require multiple accesses to the policy $\pi$ and the action-value function $Q^{\pi}$ to obtain even a single sample at each time-step $t$.
To address this, we propose approximating a limited number ($N$) of samples using the prior knowledge distribution $p(\tilde{s}_t|s_t)$, and then adjusting the probability with importance weights:
\footnotesize
\begin{equation}
\begin{aligned}
\label{eq_softa_sampling}
\tilde{s}_{ti}
&\sim p(\tilde{s}_t|s_t)
,\text{ }i = {1,2,\dots ,N}, \\
\nu^{\star soft}_{\pi}(\tilde{s}_t|s_t)
\simeq
\nu^{\star soft}_{\pi}(\tilde{s}_{ti}|s_t)
&\propto
\frac{1
}{
 p(\tilde{s}_{ti}|s_t)
}
\frac{
 p(\tilde{s}_{ti}|s_t)
 \exp(
  \mathop{\mathbb{E}}_{\tilde{a}_{ti}\sim\pi(\cdot|\tilde{s}_{ti})}
  \lbrack
    -Q^{\pi}(s_t,\tilde{a}_{ti})/\alpha_{attk}
  \rbrack
  )
}{
\mathop{\mathbb{\sum}}_{i=1}^{N} p(\tilde{s}_{ti} | s_{t})
  \exp(
    \mathop{\mathbb{E}}_{\tilde{a}_{ti}\sim\pi(\cdot|\tilde{s}_{ti})}
    \lbrack
      -Q^{\pi}(s_t,\tilde{a}_{ti})/\alpha_{attk}
    \rbrack
    )
}, \\
&\propto
\exp(
  \mathbb{E}_{\tilde{a}_{ti}\sim\pi(\cdot|\tilde{s}_{ti})}
  \lbrack
    -Q^{\pi}(s_t,\tilde{a}_{ti})/\alpha_{attk}
  \rbrack
  )
.
\end{aligned}
\end{equation}
\normalsize
This approximation is biased and of high variance, especially when the prior distribution $p(\tilde{s}_t|s_t)$ significantly deviates from the soft optimal adversary $\nu_{\pi}^{\star soft}(\tilde{s}_t|s_t)$, 
or when the number of samples is insufficient to adequately represent $\nu_{\pi}^{\star soft}(\tilde{s}_t|s_t)$. 
Despite these potential issues, the approximated distribution proves highly useful when interpreted as:
\textbf{testing the noise $p(\tilde{s}_t|s_t)$ with $N$ parallel samples at a time, then selecting a sample where the policy $\pi$ performs suboptimally, with a soft weighted probability determined by the temperature parameter $\alpha_{attk}$}.
By selecting the number of samples $N$ and adjusting the weakness parameter $\alpha_{attk}$, 
we can simulate adversarial scenarios that realistically consider the frequency of pre-assumed noise, 
without being constrained to $L_{p}$-norm noise. 
We refer to this attacker as the Soft (worst) Attack (SofA) and explore its use for training and evaluation in Sections \ref{subsubsec_softa_rl_kl} and \ref{subsubsecEvalSofA}.

\subsubsection{Epsilon Worst Approximation Attack (EpsA) for the $\alpha$-divergence Constraint}
\label{subsubsec_epsa_attacker}

Considering the case when the $f$-divergence in Eq.\eqref{eq_soft_attack_def} is specified as the $\alpha$-divergence, 
we can explore broader categories that encompass the SofA case (Section\ref{subsubsec_softa_attacker}). 
These $\alpha$-divergence constraint problems tend to have more mode-seeking solutions as $\alpha$ decreases~\citep{belousov2017f, xu2023offline}. 
Specifically, by setting $\alpha \ll 0$ and restricting the attacker's distribution selection to a uniform distribution within the $L_{\infty}$-norm range with an attack scale $\epsilon$, 
denoted as $\mathcal{U}(\tilde{s} \mid s-\epsilon, s+\epsilon)$, 
we can approximate the distribution by using the mode probability, 
denoted as $\kappa_{worst}$, as follows:
\begin{equation}
\label{eq_epsilon_worst_attack}
\nu^{\star}_{\epsilon}(\tilde{s}|s) \simeq
\begin{cases}
\kappa_{worst}
+\frac{1-\kappa_{worst}}{|\mathcal{S}_{\epsilon}|}
, & \text{if }\tilde{s}=\argmin_{\tilde{s}' \in \mathcal{B}_{\epsilon}} \mathbb{E}_{\tilde{a} \sim \pi(\cdot|\tilde{s}')}\left[ Q^{\pi}(s,\tilde{a}) \right], \\
\frac{1-\kappa_{worst}}{|\mathcal{S}_{\epsilon}|}
, & \text{otherwise},
\end{cases}
\end{equation}
where $|\mathcal{S}_{\epsilon}|$ represents the measure of the state space within the $\epsilon$-bounded domain. 
The details of this approximation are discussed in Appendix \ref{subsecApp_epsa_derivation_approximation}.
Eq. \eqref{eq_epsilon_worst_attack} can be approximated by combining the uniform distribution with 
a numerical gradient approach, similar to the \textit{Critic} attack \citep{pattanaik2017robust,zhang2020robust}.
We refer to this strategy as the Epsilon (worst) Attack (EpsA) and utilize it for training and evaluation in Sections \ref{subsubsec_epsa_attacker} and \ref{subsubsecEvalEpsA}.

\subsection{Robust off-policy Reinforcement Learning via Soft Constrained Adversary}
\label{subsec_soft_rl}
As the previous discussion, 
we introduced the two typical adversaries as the solution of the soft constrained dual problems.
By defining the soft (epsilon) worst-attack action-value function for policy $\pi$ as 
$Q^{\tilde{\pi}}(s,a):=\mathbb{E}_{\mathcal{F}, \tilde{a}_t \sim \pi\circ\nu}\left[ \sum_{t=0}^{\infty}\gamma^t r(s_t,\tilde{a}_t) | s_0 =s, a_0 = a \right]$, 
we can propose two robust off-policy RL algorithms which assume the corresponding adversaries in the appropriate MDP manners.
We should note that this framework is sample-efficient because 
it can work not only with \text{off-policy} algorithms but also \text{does not require additional interaction with the environment for the adversary}.

\subsubsection{Soft Worst Max-Entropy Reinforcement Learning (SofA-SAC)}
\label{subsubsec_softa_rl_kl}
We assume the policy is degraded by the adversary, 
and therefore introduce an additional modification to the reward function in Eq. \eqref{eq_sac_reward} as follows:
\begin{equation}
\label{eq_soft_worst_sac_reward}
\hat{\hat{r}}(s_t,a_t) \triangleq
r(s_t,a_t) + 
\mathbb{E}_{s_{t+1}\sim\mathcal{F}}\lbrack 
\alpha_{ent} 
\mathcal{H}(\pi \textcolor{black}{\circ \nu^{\star soft}_{\pi_{ent}}}(\cdot|s_{t+1})) 
\rbrack.
\end{equation}
For simplicity, we define $V^{\tilde{\pi}}(s,\tilde{s}):=\mathbb{E}_{\tilde{a} \sim \pi}\left[ Q(s,\tilde{a}) - \alpha_{ent} \log \pi(\tilde{a}|\tilde{s}) \right]$, 
then we get the following soft worst attack for the max-entropy version:
\begin{equation}
\label{eq_optimal_softa_max_ent}
\nu^{\star soft}_{\pi_{ent}}(\tilde{s_t}|s_t) =
  \frac{
    p(\tilde{s}_t|s_t)
    \exp( - V^{\tilde{\pi}}(s_t, \tilde{s}_t)/\alpha_{attk} )
  }
  {
  \int_{\tilde{s}_t}
  p(\tilde{s}_t|s_t)
  \exp \left( 
    - V^{\tilde{\pi}}(s_t, \tilde{s}_t)/\alpha_{attk} 
    \right) d\tilde{s}_t
  }.
\end{equation}
Under such adversary, 
we can define the corresponding Bellman operator with a contraction property: 
\begin{footnotesize}
\begin{equation}
\label{eq_soft_worst_kl_bellman}
\begin{aligned}
({\underline{\mathcal{T}}_{soft}^{\pi}}Q(s_t,a_t) 
\triangleq r(s_t,a_t)
= r(s_t,a_t)
+ \gamma \mathbb{E}_{s_{t+1}\sim \mathcal{F} }
\left[
-\alpha_{attk} \log \left(
  \mathbb{E}_{\tilde{s}_{t+1}\sim p }
  \left[
  \exp(
    - V^{\tilde{\pi}}(s_{t+1}, \tilde{s}_{t+1})/\alpha_{attk} 
    )
  \right]
  \right)
\right].
\end{aligned}
\end{equation}
\end{footnotesize}
For the policy improvement, we consider the KL minimization problem under the adversary as:
\begin{equation}
\label{eq_soft_worst_sac_policy_loss1}
\begin{aligned}
L(\pi) &=
{\mathbb{E}}_{s_t \sim D(\cdot)}
\left[
D_{KL}(\pi \textcolor{black}{\circ \nu^{\star soft}_{\pi_{ent, fixed}} (\cdot|s_t)} \parallel   \frac{\exp(Q^{\pi}(s_t,\cdot)/\alpha_{ent})}{
\int_{\tilde{a}_t}\exp(Q^{\pi}(s_t,\tilde{a}_t)/\alpha_{ent}) d\tilde{a}_t
})
\right]
\\
&\propto
{\mathbb{E}}_{s_t \sim D(\cdot)}
\left[
\textcolor{black}{{\mathbb{E}}_{\tilde{s}_t \sim \nu^{\star soft} _{\pi_{ent, fixed}} }}
\left[
{\mathbb{E}}_{\tilde{a}_t \sim \pi }
\left[
\alpha_{ent}\log{\pi(\tilde{a}_t|\tilde{s}_t)} - Q^{\pi}(s_t,\tilde{a}_t)
\right]
\right] + const.
\right]
\\
&=
{\mathbb{E}}_{s_t \sim D(\cdot)}
\left[
  {\mathbb{E}}_{\tilde{s}_t \sim p }
\left[
\frac{\nu^{\star soft}_{\pi_{ent, fixed}}(\tilde{s}_t|s_t)}{p(\tilde{s}_t|s_t)}
{\mathbb{E}}_{\tilde{a}_t \sim \pi }
\left[
\alpha_{ent}\log{\pi(\tilde{a}_t|\tilde{s}_t)} - Q^{\pi}(s_t,\tilde{a}_t)
\right]
\right]
\right]
,
\end{aligned}
\end{equation}
where $\nu^{\star soft}_{\pi_{ent, fixed}}$ denotes the (soft) optimal adversary that is fixed during policy improvement. 
We call this RL framework as the Soft (worst) Attack SAC (SofA-SAC).
In the practical algorithm, we approximate using $N$ samples for $\tilde{s}_t \sim p(\cdot|s_t)$. 
For a comprehensive derivation, theoretical validity properties (contraction and policy improvement), 
and detailed practical implementation, 
see Appendix \ref{subsecApp_softa_sac_derivation}, \ref{subsecApp_softa_sac_contraction_policy_improvement}, and \ref{subsecApp_softa_practical_implementation}.

\subsubsection{Epsilon Worst Max-Entropy Reinforcement Learning (EpsA-SAC)}
\label{subsubsec_epsa_rl}

As in the previous subsection, 
we assume the policy is mislead by the attack, 
then we consider the max-entropy version of the epsilon worst attack in Eq. \eqref{eq_epsilon_worst_attack} as:
\begin{align}
\label{eq_epsilon_worst_attack_maxent}
\textcolor{black}{\nu^{\star epsilon}_{\pi_{ent}}(\tilde{s_t}|s_t)}
&\simeq
\begin{cases}
\kappa_{worst}
+\frac{1-\kappa_{worst}}{|\mathcal{S}_{\epsilon}|}
, & \text{if }\tilde{s}=\argmin_{\tilde{s}'\in\mathcal{B}_{\epsilon}} \mathbb{E}_{\tilde{a} \sim \pi}\left[ Q^{\pi}(s,\tilde{a}) -\alpha_{ent} \log\pi(\tilde{a}|\tilde{s}') \right] \\
\frac{1-\kappa_{worst}}{|\mathcal{S}_{\epsilon}|}
, & \text{otherwise}
\end{cases}.
\end{align}
Under this perturbation, we define the epsilon-worst Bellman operator as:
\footnotesize
\begin{equation}
\begin{aligned}
\label{eq_epsilon_worst_bellman}
&(\mathcal{\underline{T}}^{\pi}_{epsilon}Q)(s_t,a_t)
\triangleq r(s_t,a_t)
+\gamma {\mathbb{E}}_{s_{t+1}\sim \mathcal{F}}
  \left[
  \textcolor{black}{\mathbb{E}_{\tilde{s}_{t+1}\sim \nu^{\star epsilon}_{\pi_{ent}}}}
  \left[
    V^{\tilde{\pi}}(s_{t+1}, \tilde{s}_{t+1})
  \right]
  \right]\\
= & r
+\gamma {\mathbb{E}}_{s_{t+1}\sim \mathcal{F}}
\left[
    \kappa_{worst}
    \mathbb{E}_{\tilde{s}_{t+1}\sim \nu^{*worst}_{\pi_{ent}}}
  \left[
    V^{\tilde{\pi}}(s_{t+1}, \tilde{s}_{t+1})
  \right]
  +
  \textcolor{black}{
    (1-\kappa_{worst})
    \mathbb{E}_{\tilde{s}_{t+1}\sim \mathcal{U}_{\epsilon}}
    \left[
      V^{\tilde{\pi}}(s_{t+1}, \tilde{s}_{t+1})
    \right]
  }
\right].
\end{aligned}
\end{equation}
\normalsize
For simplicity, we again use $V^{\tilde{\pi}}(s,\tilde{s}):=\mathbb{E}_{\tilde{a} \sim \pi}\left[ Q(s,\tilde{a}) - \alpha_{ent} \log \pi(\tilde{a}|\tilde{s}) \right]$.
By considering the same divergence minimization problem in Eq. \eqref{eq_soft_worst_sac_policy_loss1}, 
we can improve the policy through:
\begin{align}
\label{eq_epsilon_worst_sac_policy_loss1}
\begin{split}
L(\pi) &=
{\mathbb{E}}_{s_t \sim D}
\left[
\textcolor{black}{
  \mathbb{E}}_{
    \tilde{s}_t \sim \nu^{\star epsilon}_{\pi_{ent}}
  }
\left[
\mathbb{E}_{
  \tilde{a}_t \sim \pi
  }
\left[
\alpha_{ent}\log{\pi(\tilde{a}_t|\tilde{s}_t)} - Q^{\pi}(s_t,\tilde{a}_t)
\right]
\right]
\right] \\
&=
{\mathbb{E}}_{s_t \sim D}
\left[
\textcolor{black}{
    {\kappa_{worst}\mathbb{E}}_{
      \tilde{s}_t \sim \nu^{*worst}_{\pi_{ent}}
      }
  }
  \left[
    -V^{\tilde{\pi}}(s_{t}, \tilde{s}_{t})
  \right]
  + \textcolor{black}{
    {(1-\kappa_{worst})\mathbb{E}}_{\tilde{s}_t \sim \mathcal{U}_{\epsilon}}
  }
  \left[
    -V^{\tilde{\pi}}(s_{t}, \tilde{s}_{t})
  \right]
\right].
\end{split}
\end{align}
We refer to this RL framework as the Epsilon (worst) Attack SAC (EpsA-SAC).
We can assert that this Bellman operator also possesses a contraction property under a fixed policy, and once the adversary is fixed, the policy can be improved monotonically. 
For detailed information, please see Appendix \ref{subsecApp_epsa_sac_contraction_policy_improvement}.
In Eq. \eqref{eq_epsilon_worst_attack_maxent}, 
obtaining the analytical worst sample ($\tilde{s}_t \sim \nu^{\star worst}_{\pi_{ent}}(\cdot|s_t)$) 
is infeasible when the environment comprises continuous states and actions.
However, we have found that numerical approximations of the worst samples using the Critic attack \citep{pattanaik2017robust,zhang2020robust} are practically effective. 
We employ the policy mean $\mu(\tilde{s}_t)$ and the action-value $Q^{\pi}(s_t, \tilde{a}_t)$, 
then apply gradient descent iteration to solve $\tilde{s}_{t} \simeq \argmin_{\tilde{s}'_t \in \mathcal{B}_{\epsilon}}{Q^{\pi}(s_t, \mu(\tilde{s}'_t))}$ in subsequent experiments.

\section{Experiments}
\label{secExperiments}

In this section, 
we set up experiments to address the questions posed in the introduction: 
\textbf{(1)} Can we develop a robust \textbf{off-policy} algorithm that accounts for long-term rewards without requiring additional interactions? 
\textbf{(2)} Is it possible to incorporate more arbitrary distributions, based on prior knowledge, beyond the conventional $L_{p}$-norm constrained range, into both adversaries and defenses? 
Responses to 
\textbf{(1)} are addressed in Sections \ref{subsubsecEvalSofA} and \ref{subsubsecEvalEpsA}, 
and those to \textbf{(2)} are presented in Section \ref{subsubsecEvalSofA}. 

\paragraph{Environments and Common settings}
\label{paraEnvironments}
We use four OpenAI Gym MuJoCo environments \citep{todorov2012mujoco}: Hopper, HalfCheetah, Walker2d, and Ant, 
as utilized in most prior works \citep{zhang2020robust,zhang2021robust,oikarinen2021robust,liang2022efficient}.
Detailed settings are documented in \ref{subsecApp_SAC_settings}.

\paragraph{Baselines and Implementations.}
\label{paraBaselines}
We select the Soft Actor Critic (SAC) \citep{haarnoja2018softa,haarnoja2018softb} as our base off-policy method 
and prepare our algorithms alongside two state-of-the-art robust algorithms for comparison, 
both of which potentially operate in an off-policy manner. 
First, the SAC version of SA-MDP \citep{zhang2020robust}, referred to as \textit{SA-SAC}, 
and second, the SAC version of WocaR \citep{liang2022efficient}, 
which we denote as \textit{WocaR-SAC}. 
These comparisons aim to evaluate the performance capabilities of these algorithms under off-policy conditions.
There is no provided code \footnote[2]{
While SA-DDPG is implemented in \cite{zhang2020robust}, 
it appears that DDPG does not perform adequately on the four MuJoCo benchmarks. 
}
for SAC-versions (off-policy) of SA-MDP, WocaR, nor evaluation methods, 
then we incorporate these methods into our implementation, followed by tuning hyper-parameters and settings appropriately.
WocaR-SAC works well in small benchmarks (Pendulum, InvertedPendulum). 
However, we found that during training, as the attack scale increases, 
the policy suddenly degrades in performance due to the worst Q-value dropping too low in the four benchmarks. 
This occurred even when we used tighter bound convex-relaxation methods \citep{zhang2018efficient,zhang2019towards} than the IBP \citep{gowal2018effectiveness} used in the original WocaR implementation. 
More information for implementations and hyper-parameters are provided 
in Appendix \ref{subsecApp_SAC_settings} for the base SAC, 
in Appendix \ref{subsecApp_SofA-SAC_settings} for SofA-SAC, 
in Appendix \ref{subsecApp_SA-SAC_settings} for SA-SAC, 
and Appendix \ref{subsecApp_WocaR-SAC_settings} for WocaR-SAC.

\subsection{Training and Evaluation}
\label{subsecEvaluations}

We have established two evaluation metrics for our analysis. 
The first metric aims to assess the effectiveness against adversaries not constrained by the $L_{p}$-norm, 
as discussed in Sections \ref{subsubsec_softa_attacker} and \ref{subsubsec_softa_rl_kl}. 
The second metric evaluates the resilience of our methods against conventional strong attacks within the $L_{\infty}$-norm ball, 
detailed in Sections \ref{subsubsec_epsa_attacker} and \ref{subsubsec_epsa_rl}.

\subsubsection{Evaluations of Soft Worst Case Scenarios under Gaussian-Based Attacks}
\label{subsubsecEvalSofA}

\begin{figure}[t]
\centering
\includegraphics[keepaspectratio, width=1\linewidth]{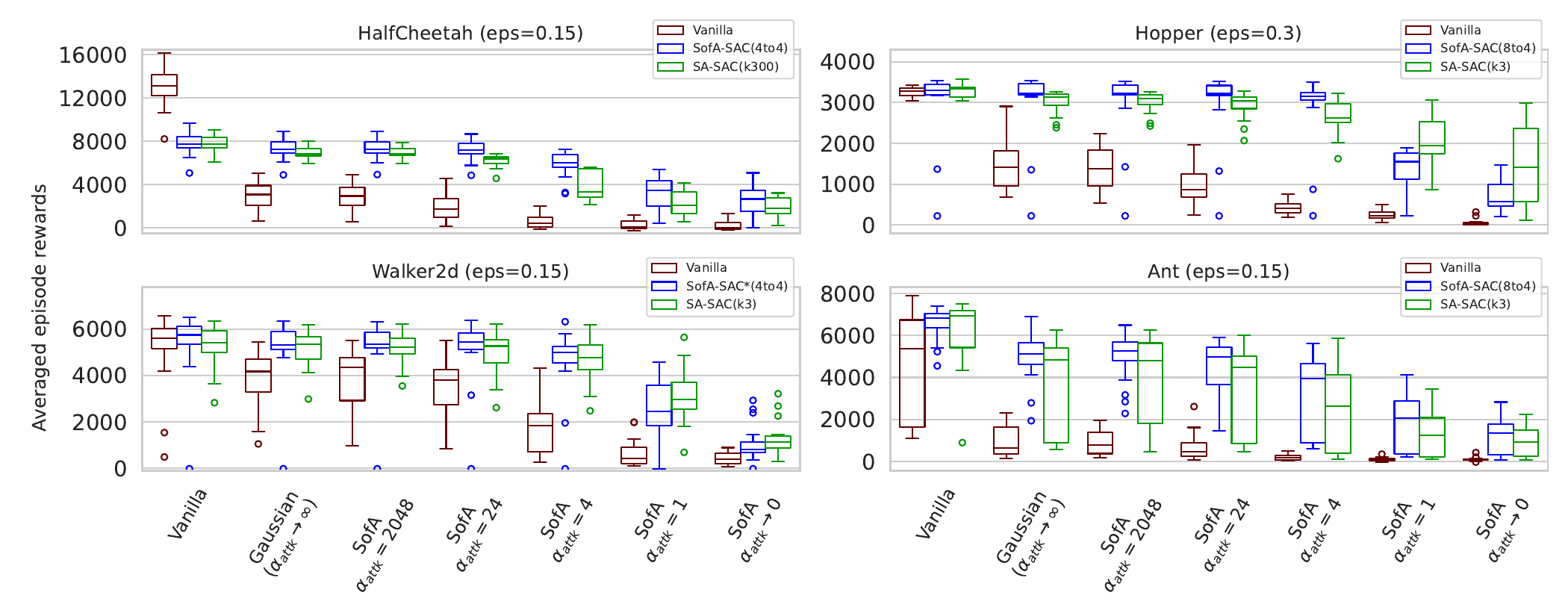}
\vspace{-5truemm}
\caption{
Robustness evaluation results of SofA-SAC and baseline algorithms under the Gaussian based attacks.
Each boxplot depicts the 25\%, 50\%, and 75\% percentile values of the mean returns.
}
\label{fig_main_softa_boxplot}
\end{figure}

\paragraph{Task Setup}

In this study, 
we use a Gaussian distribution as the prior knowledge perturbation $p(\tilde{s}_t|s_t)$, 
setting the standard deviations for the attack scales as $\sigma=0.15, 0.30, 0.15, 0.15$ for HalfCheetah, Hopper, Walker2d, and Ant, respectively. 
We train the standard SAC (\textit{Vanilla-SAC}), 
SA-SAC, and our proposed defense method SofA-SAC using identical training steps. 
Subsequently, we conduct evaluations under various attack settings.

We prepare four attack settings: 
one using the prior knowledge distribution(\textit{Gaussian}) as-is,
another applying our proposed method, denoted as SofA($\alpha_{attk}$), with varying degrees of adversarial preference parameter $\alpha_{attk}$.
Finally, for reference, 
we report results for the \textit{MaxActionDiff}(\textit{MAD}) \citep{zhang2020robust} and the \textit{Critic} \citep{pattanaik2017robust,zhang2020robust}, 
where the standard deviation value serves as the noise constraint range in Appendix \ref{secApp_SofA_additional_results}.
For the sampling approximation, we use $N=64$ both for the evaluation attacker (SofA) and the proposal DRL methods (SotfA-SAC).
For SA-SAC and SofA-SAC training, 
we appropriately tune the coefficient terms that achieve robustness without compromising task performance (see Appendix \ref{subsecApp_SofA_additional_results_ablation_hyperparams}).
For SofA-SAC, we maintain good performance and robustness across all tasks by $\alpha_{attk}=4.0$ at the end of trainings,
therefore, we consistently use this value. 
During tuning of the consistency parameter for SA-SAC, 
we find there is a trade-off between robustness and performance without attacks (see Appendix \ref{subsecApp_SoftA_additional_results_softa_coefficient}), especially in HalfCheetah, 
as reported in \cite{liang2022efficient}. 
To compare under equal conditions, 
we tune the coefficient term of SA-SAC to achieve the same performance without attacks as SofA-SAC. 
Details of these hyperparameter settings are documented in Appendix \ref{subsecApp_SofA-SAC_settings} and \ref{subsecApp_SA-SAC_settings}.

\paragraph{Result}

Fig. \ref{fig_main_softa_boxplot} shows results of the robustness evaluation scores for Vanilla-SAC, our proposal method (SofA-SAC), and SA-SAC.
As we estimated in Section \ref{subsubsec_softa_attacker},
task scores drop as the attacker worst preference parameter ($\alpha_{attk}$) approaches 0, here $\alpha_{attk}\rightarrow0$ means worst sample pick-up.
Though Vanilla-SAC drop its scores drastically, SofA-SAC and SA-SAC keep the performance even as the attacker gets stronger.
Among all tasks, SofA-SAC keeps superior or competitive performance in the range ot $\alpha_{attk} = [4.0,2048.0]$ to SA-SAC,
while inferior in the range of [0.0, 1.0] for Hopper. 
In this study, SofA-SAC incorporates the attacked observation with the temperature parameter, 
$\alpha_{attk}=4$ into the optimization problem during training, 
then stronger attack than this assumption may occur performance deterioration.
Remarkably, both SofA-SAC and SA-SAC not only enhance robustness to observation noise 
but also consistently achieve high scores across multiple seeds. 
Therefore, we recommend these methods to the reader, 
except for tasks requiring conservative behavior where a trade-off between performance and robustness is necessary like in the HalfCheetah task.

\subsubsection{Evaluations of Strong Attackers under Conventional $L_{\infty}$-norm Constraints}
\label{subsubsecEvalEpsA}

\begin{figure}[t]
\centering
\includegraphics[keepaspectratio, width=1\linewidth]{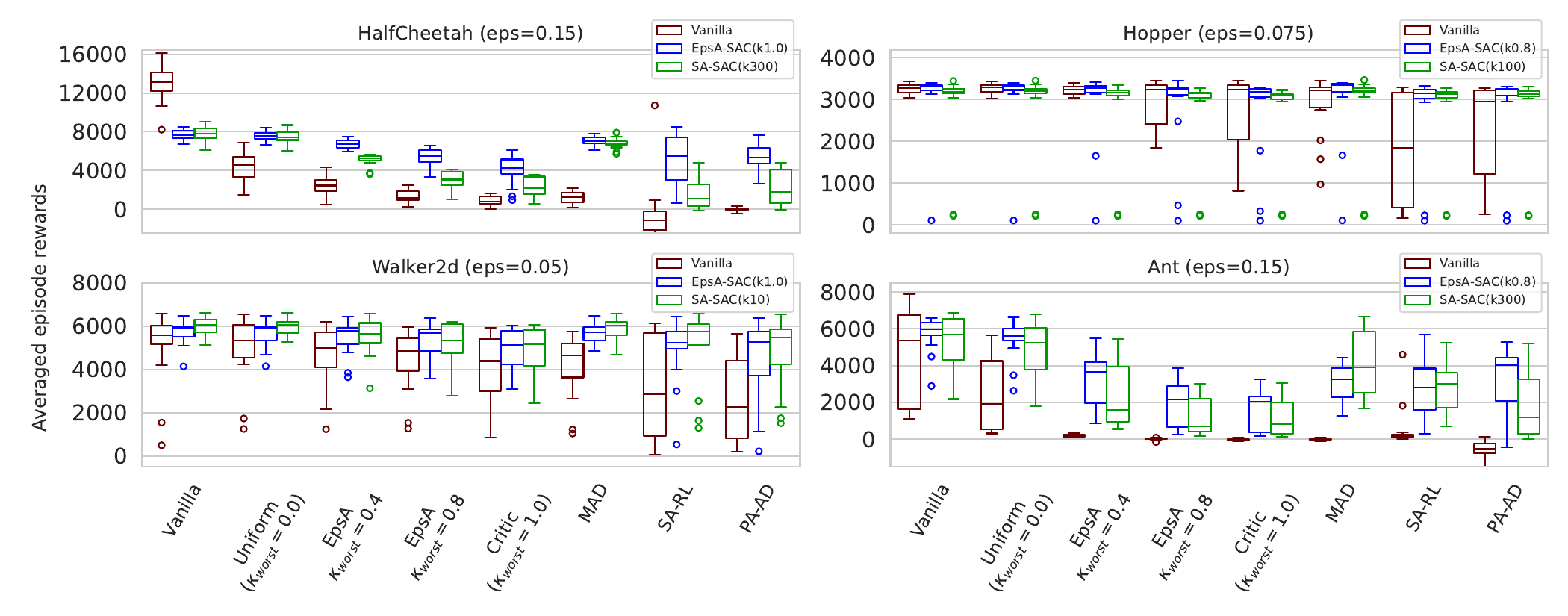}
\vspace{-5truemm}
\caption{
Robustness evaluation results of EpsA-SAC and baseline algorithms under the $L_{\infty}$-norm attacks.
Each boxplot depicts the 25\%, 50\%, and 75\% percentile values of the mean returns.
}
\label{fig_main_epsa_boxplot}
\end{figure}

\paragraph{Task Setup}

For the robustness evaluation, we incorporate strong attackers as used in the most recent studies \citep{zhang2020robust, zhang2021robust, sun2021strongest, oikarinen2021robust, liang2022efficient}. 
In addition to the conventional Random (Uniform), 
MAD, and Critic attacks, 
we include powerful attacks known as \textit{SA-RL} \citep{zhang2021robust} 
and \textit{PA-AD} \citep{sun2021strongest} in our main evaluations. 
Furthermore, we introduce our proposed attack, 
EpsA ($\kappa_{worst}$), to examine robustness trends across different methods.
We adopt the same attack scales as those commonly used in previous studies: 
$\epsilon=0.15, 0.075, 0.05, 0.15$ for HalfCheetah, Hopper, Walker2d, and Ant, respectively.

For EpsA-SAC training, we find that increasing the worst ratio ${\kappa_{worst}}$ from 0.0 to 0.8 or 1.0 during training works well, 
then we choose the one that is better in each task. 
As in Section \ref{subsubsecEvalSofA}, 
there is a trade-off between robustness and performance without attacks in some tasks (see Appendix \ref{fig_appendix_epsa_coefficient_boxplot}). 
Then, we tune the coefficient term to achieve the same performance without attacks as EpsA-SAC does. 
Details of these hyperparameter settings are documented in Appendix \ref{subsecApp_EpsA-SAC_settings} and \ref{subsecApp_SA-SAC_settings}.

\paragraph{Result}

Fig. \ref{fig_main_epsa_boxplot} shows the evaluation results under the strong $L_{\infty}$-norm attackers.
Especially in the more complex tasks (HalfCheetah, Ant) 
and strong attack evaluations (EpsA, Critic, SA-RL, PA-AD) that estimate long-term horizon, 
EpsA-SAC demonstrates superior or competitive performance compared to SA-SAC. 
In tasks with smaller perturbations (Walker), 
EpsA-SAC's performance is competitive but slightly inferior to SA-SAC. 
We hypothesize that in tasks sensitive to small perturbations, 
simply maintaining consistent behaviors under these perturbations is more stable and preferable than comprehensive methods that account for MDPs under perturbations.

We should note that our results are based on the SAC implementation, 
not PPO, 
so the Q-function is trained according to the original method. 
Our DRL framework incorporates the adversarial effect into the Q-function, 
making the evaluation metrics that use such Q-functions much stronger. 
However, it is still superior to SA-SAC for the most tasks.

\subsection{Ablation Studies}
\label{subsecAblations}

Due to limited space, 
we focus on the tasks where the influence of the hyperparameters and the effect of the adversaries during training are most apparent.
Additional results, including other tasks and discussions, are documented in Appendices \ref{secApp_SofA_additional_results} and \ref{secApp_Epsilon_additional_results}.
\paragraph{SofA-SAC's trade-off between performance and robustness}

\begin{figure}[t]
  \begin{minipage}[c]{0.45\hsize}
    \centering
    \includegraphics[keepaspectratio, width=\linewidth]{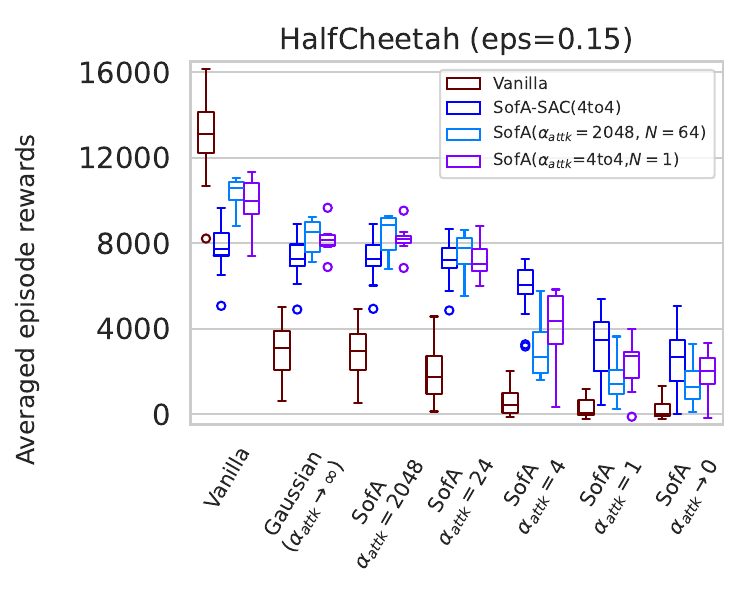}
    \vspace{-5truemm}
    \caption{
      Ablation results for SofA-SAC's hyperparameter. 
      We change sample number and worst preference parameter, 
      from $N=64$ to $1$ and from $\alpha_{attk}=4$ to $2048$.
    }
    \label{fig_abla_half_softa_boxplot}
  \end{minipage}
  \hspace{0.05\columnwidth}
  \begin{minipage}[c]{0.45\hsize}
    \centering
    \includegraphics[keepaspectratio, width=\linewidth]{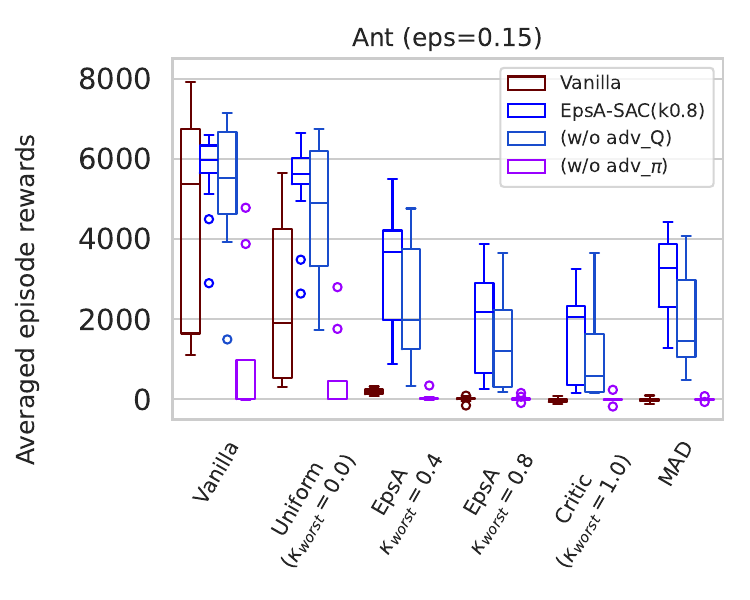}
    \vspace{-5truemm}
    \caption{
      Ablation results for EpsA-SAC's training strategy. 
      We omit adversarial perturbation during policy improvement and Q updating in training.
    }
    \label{fig:spaceship}
    \label{fig_abla_ant_epsa_boxplot}
  \end{minipage}
\end{figure}
Some readers may concern difference between SofA-SAC and SAC learning with just perturbed observation. 
Fig. \ref{fig_abla_half_softa_boxplot} shows the confirmation result in HalfCheetah, 
by changing hyperparameter $\alpha_{attk}$ and sample number $N$. 
Learning with high temperature parameter, $\alpha_{attk}=2048$, 
corresponding near averaged Gaussian perturbation, 
and N = 1 means using just sample from Gaussian distribution. 
Those both two variations result in relatively high performance without attacks than the reference, 
while less robustness under the stronger attack ($\alpha_{attk}\leq 24.0$). 
Therefore, we can say our SofA-SAC appropriately
incorporates conservative behaviors that we assume at the training phase.

\paragraph{
  Importance of adversaries during policy improvement and action value update
  }
To analyze the effective factors of our learning framework, we conduct two variant trainings of EpsA-SAC. 
The one is lack of adversary during policy improvement, 
and the other is lack of adversary during action value update.
Fig. \ref{fig_abla_ant_epsa_boxplot} shows these ablation results.
Without adversaries during Q-value updates, 
the results exhibit high variance and less robust performance, 
while the performance deteriorates when adversaries are absent during policy improvement.
In the former case, the Q-function cannot adequately consider the pessimistic scenarios around the state $s_{t}$, 
and in the latter case, the policy fails to improve performance under attack, leading to low Q-values.

\section{Conclusion and Future Work}
\label{secConclusion}

We propose a new perspective on robust RL by introducing f-divergence constrained optimal adversaries with prior distributions. 
Our methods align naturally with theoretical expectations and demonstrate robust performance, 
particularly against strong attacks that comprehensively consider MDPs. 
This framework offers a more flexible approach to designing robustness that closely aligns with real-world challenges and enhances the performance of off-policy algorithms.

For future work, we identify three main directions: developing a stronger algorithm that integrates functional smoothness; creating more efficient metrics for EpsA-SAC computation, as current methods rely heavily on gradient iterations for both policy evaluation and policy improvement; and extending this framework to other research domains, such as domain randomization.


\section{Acknowledgements}
\label{secAcknowledgements}

We would like to express our gratitude to Satoshi Yamamori and Sotetsu Koyamada for their valuable advice on this research. 
We also extend our appreciation to all members of the Computer Science Domain at Honda R\&D Co., 
especially to Ken Iinuma, Akira Kanahara, and Kenji Goto, 
for their cooperation and understanding in supporting this research.



\small

\medskip

\bibliographystyle{plainnat}
\bibliography{neurips_2024arr}

\begin{thebibliography}{57}
\providecommand{\natexlab}[1]{#1}
\providecommand{\url}[1]{\texttt{#1}}
\expandafter\ifx\csname urlstyle\endcsname\relax
  \providecommand{\doi}[1]{doi: #1}\else
  \providecommand{\doi}{doi: \begingroup \urlstyle{rm}\Url}\fi

\bibitem[Behzadan and Munir(2017)]{behzadan2017whatever}
Vahid Behzadan and Arslan Munir.
\newblock Whatever does not kill deep reinforcement learning, makes it
  stronger.
\newblock \emph{arXiv preprint arXiv:1712.09344}, 2017.

\bibitem[Belousov and Peters(2017)]{belousov2017f}
Boris Belousov and Jan Peters.
\newblock f-divergence constrained policy improvement.
\newblock \emph{arXiv preprint arXiv:1801.00056}, 2017.

\bibitem[Belousov and Peters(2019)]{belousov2019entropic}
Boris Belousov and Jan Peters.
\newblock Entropic regularization of markov decision processes.
\newblock \emph{Entropy}, 21\penalty0 (7):\penalty0 674, 2019.

\bibitem[Boularias et~al.(2011)Boularias, Kober, and
  Peters]{boularias2011relative}
Abdeslam Boularias, Jens Kober, and Jan Peters.
\newblock Relative entropy inverse reinforcement learning.
\newblock In \emph{Proceedings of the fourteenth international conference on
  artificial intelligence and statistics}, pages 182--189. JMLR Workshop and
  Conference Proceedings, 2011.

\bibitem[Boyd and Vandenberghe(2004)]{boyd2004convex}
Stephen~P Boyd and Lieven Vandenberghe.
\newblock \emph{Convex optimization}.
\newblock Cambridge university press, 2004.

\bibitem[Carlini and Wagner(2017)]{carlini2017adversarial}
Nicholas Carlini and David Wagner.
\newblock Adversarial examples are not easily detected: Bypassing ten detection
  methods.
\newblock In \emph{Proceedings of the 10th ACM Workshop on Artificial
  Intelligence and Security}, pages 3--14, 2017.

\bibitem[Fox et~al.(2015)Fox, Pakman, and Tishby]{fox2015taming}
Roy Fox, Ari Pakman, and Naftali Tishby.
\newblock Taming the noise in reinforcement learning via soft updates.
\newblock \emph{arXiv preprint arXiv:1512.08562}, 2015.

\bibitem[Franzmeyer et~al.(2024)Franzmeyer, McAleer, Henriques, Foerster, Torr,
  Bibi, and de~Witt]{franzmeyer2024illusory}
Tim Franzmeyer, Stephen~Marcus McAleer, Joao~F. Henriques, Jakob~Nicolaus
  Foerster, Philip Torr, Adel Bibi, and Christian~Schroeder de~Witt.
\newblock Illusory attacks: Information-theoretic detectability matters in
  adversarial attacks.
\newblock In \emph{The Twelfth International Conference on Learning
  Representations}, 2024.
\newblock URL \url{https://openreview.net/forum?id=F5dhGCdyYh}.

\bibitem[Gleave et~al.(2019)Gleave, Dennis, Wild, Kant, Levine, and
  Russell]{gleave2019adversarial}
Adam Gleave, Michael Dennis, Cody Wild, Neel Kant, Sergey Levine, and Stuart
  Russell.
\newblock Adversarial policies: Attacking deep reinforcement learning.
\newblock \emph{arXiv preprint arXiv:1905.10615}, 2019.

\bibitem[Goodfellow et~al.(2014)Goodfellow, Shlens, and
  Szegedy]{goodfellow2014explaining}
Ian Goodfellow, Jonathon Shlens, and Christian Szegedy.
\newblock Explaining and harnessing adversarial examples.
\newblock \emph{JSME International Journal, Series 1: Solid Mechanics, Strength
  of Materials}, 33\penalty0 (4):\penalty0 468--473, 12 2014.

\bibitem[Gowal et~al.(2018)Gowal, Dvijotham, Stanforth, Bunel, Qin, Uesato,
  Arandjelovic, Mann, and Kohli]{gowal2018effectiveness}
Sven Gowal, Krishnamurthy Dvijotham, Robert Stanforth, Rudy Bunel, Chongli Qin,
  Jonathan Uesato, Relja Arandjelovic, Timothy Mann, and Pushmeet Kohli.
\newblock On the effectiveness of interval bound propagation for training
  verifiably robust models.
\newblock \emph{arXiv preprint arXiv:1810.12715}, 2018.

\bibitem[Haarnoja et~al.(2017)Haarnoja, Tang, Abbeel, and
  Levine]{haarnoja2017reinforcement}
Tuomas Haarnoja, Haoran Tang, Pieter Abbeel, and Sergey Levine.
\newblock Reinforcement learning with deep energy-based policies.
\newblock In \emph{International conference on machine learning}, pages
  1352--1361. PMLR, 2017.

\bibitem[Haarnoja et~al.(2018{\natexlab{a}})Haarnoja, Zhou, Abbeel, and
  Levine]{haarnoja2018softa}
Tuomas Haarnoja, Aurick Zhou, Pieter Abbeel, and Sergey Levine.
\newblock Soft actor-critic: Off-policy maximum entropy deep reinforcement
  learning with a stochastic actor.
\newblock In \emph{International conference on machine learning}, pages
  1861--1870. PMLR, 2018{\natexlab{a}}.

\bibitem[Haarnoja et~al.(2018{\natexlab{b}})Haarnoja, Zhou, Hartikainen,
  Tucker, Ha, Tan, Kumar, Zhu, Gupta, Abbeel, et~al.]{haarnoja2018softb}
Tuomas Haarnoja, Aurick Zhou, Kristian Hartikainen, George Tucker, Sehoon Ha,
  Jie Tan, Vikash Kumar, Henry Zhu, Abhishek Gupta, Pieter Abbeel, et~al.
\newblock Soft actor-critic algorithms and applications.
\newblock \emph{arXiv preprint arXiv:1812.05905}, 2018{\natexlab{b}}.

\bibitem[He et~al.(2016)He, Zhang, Ren, and Sun]{he2016deep}
Kaiming He, Xiangyu Zhang, Shaoqing Ren, and Jian Sun.
\newblock Deep residual learning for image recognition.
\newblock In \emph{Proceedings of the IEEE conference on computer vision and
  pattern recognition}, pages 770--778, 2016.

\bibitem[Huang et~al.(2017)Huang, Papernot, Goodfellow, Duan, and
  Abbeel]{huang2017adversarial}
Sandy Huang, Nicolas Papernot, Ian Goodfellow, Yan Duan, and Pieter Abbeel.
\newblock Adversarial attacks on neural network policies.
\newblock \emph{arXiv preprint arXiv:1702.02284}, 2017.

\bibitem[Ilyas et~al.(2018)Ilyas, Engstrom, Athalye, and Lin]{ilyas2018black}
Andrew Ilyas, Logan Engstrom, Anish Athalye, and Jessy Lin.
\newblock Black-box adversarial attacks with limited queries and information.
\newblock \emph{35th International Conference on Machine Learning (ICML)}, 5,
  2018.

\bibitem[Kim et~al.(2021)Kim, Seo, Lee, Jeon, Hwang, Yang, and
  Kim]{kim2021demodice}
Geon-Hyeong Kim, Seokin Seo, Jongmin Lee, Wonseok Jeon, HyeongJoo Hwang,
  Hongseok Yang, and Kee-Eung Kim.
\newblock Demodice: Offline imitation learning with supplementary imperfect
  demonstrations.
\newblock In \emph{International Conference on Learning Representations}, 2021.

\bibitem[Kim et~al.(2022)Kim, Lee, Jang, Yang, and Kim]{kim2022lobsdice}
Geon-Hyeong Kim, Jongmin Lee, Youngsoo Jang, Hongseok Yang, and Kee-Eung Kim.
\newblock Lobsdice: Offline learning from observation via stationary
  distribution correction estimation.
\newblock \emph{Advances in Neural Information Processing Systems},
  35:\penalty0 8252--8264, 2022.

\bibitem[Kingma and Ba(2014)]{kingma2014adam}
Diederik~P Kingma and Jimmy Ba.
\newblock Adam: A method for stochastic optimization.
\newblock \emph{arXiv preprint arXiv:1412.6980}, 2014.

\bibitem[Kiran et~al.(2021)Kiran, Sobh, Talpaert, Mannion, Al~Sallab, Yogamani,
  and P{\'e}rez]{kiran2021deep}
B~Ravi Kiran, Ibrahim Sobh, Victor Talpaert, Patrick Mannion, Ahmad~A
  Al~Sallab, Senthil Yogamani, and Patrick P{\'e}rez.
\newblock Deep reinforcement learning for autonomous driving: A survey.
\newblock \emph{IEEE Transactions on Intelligent Transportation Systems},
  23\penalty0 (6):\penalty0 4909--4926, 2021.

\bibitem[Kober et~al.(2013)Kober, Bagnell, and Peters]{kober2013reinforcement}
Jens Kober, J~Andrew Bagnell, and Jan Peters.
\newblock Reinforcement learning in robotics: A survey.
\newblock \emph{The International Journal of Robotics Research}, 32\penalty0
  (11):\penalty0 1238--1274, 2013.

\bibitem[Kos and Song(2017)]{kos2017delving}
Jernej Kos and Dawn Song.
\newblock Delving into adversarial attacks on deep policies.
\newblock \emph{arXiv preprint arXiv:1705.06452}, 2017.

\bibitem[Kostrikov et~al.(2019)Kostrikov, Nachum, and
  Tompson]{kostrikov2019imitation}
Ilya Kostrikov, Ofir Nachum, and Jonathan Tompson.
\newblock Imitation learning via off-policy distribution matching.
\newblock \emph{arXiv preprint arXiv:1912.05032}, 2019.

\bibitem[Krizhevsky et~al.(2012)Krizhevsky, Sutskever, and
  Hinton]{krizhevsky2012imagenet}
Alex Krizhevsky, Ilya Sutskever, and Geoffrey~E Hinton.
\newblock Imagenet classification with deep convolutional neural networks.
\newblock \emph{Advances in neural information processing systems}, 25, 2012.

\bibitem[Kurakin et~al.(2016)Kurakin, Goodfellow, and
  Bengio]{kurakin2016adversarial}
Alexey Kurakin, Ian Goodfellow, and Samy Bengio.
\newblock Adversarial examples in the physical world.
\newblock \emph{arXiv preprint arXiv:1607.02533}, 2016.

\bibitem[Levine et~al.(2016)Levine, Finn, Darrell, and Abbeel]{levine2016end}
Sergey Levine, Chelsea Finn, Trevor Darrell, and Pieter Abbeel.
\newblock End-to-end training of deep visuomotor policies.
\newblock \emph{The Journal of Machine Learning Research}, 17\penalty0
  (1):\penalty0 1334--1373, 2016.

\bibitem[Liang et~al.(2022)Liang, Sun, Zheng, and Huang]{liang2022efficient}
Yongyuan Liang, Yanchao Sun, Ruijie Zheng, and Furong Huang.
\newblock Efficient adversarial training without attacking: Worst-case-aware
  robust reinforcement learning.
\newblock \emph{Advances in Neural Information Processing Systems},
  35:\penalty0 22547--22561, 2022.

\bibitem[Lin et~al.(2017)Lin, Hong, Liao, Shih, Liu, and Sun]{lin2017tactics}
Yen-Chen Lin, Zhang-Wei Hong, Yuan-Hong Liao, Meng-Li Shih, Ming-Yu Liu, and
  Min Sun.
\newblock Tactics of adversarial attack on deep reinforcement learning agents.
\newblock \emph{arXiv preprint arXiv:1703.06748}, 2017.

\bibitem[Liu et~al.(2024)Liu, Chakraborty, Sun, and Huang]{liu2023rethinking}
Xiangyu Liu, Souradip Chakraborty, Yanchao Sun, and Furong Huang.
\newblock Rethinking adversarial policies: A generalized attack formulation and
  provable defense in rl.
\newblock 2024.

\bibitem[Ma et~al.(2022)Ma, Shen, Jayaraman, and Bastani]{ma2022versatile}
Yecheng Ma, Andrew Shen, Dinesh Jayaraman, and Osbert Bastani.
\newblock Versatile offline imitation from observations and examples via
  regularized state-occupancy matching.
\newblock In \emph{International Conference on Machine Learning}, pages
  14639--14663. PMLR, 2022.

\bibitem[Mandlekar et~al.(2017)Mandlekar, Zhu, Garg, Fei-Fei, and
  Savarese]{mandlekar2017adversarially}
Ajay Mandlekar, Yuke Zhu, Animesh Garg, Li~Fei-Fei, and Silvio Savarese.
\newblock Adversarially robust policy learning: Active construction of
  physically-plausible perturbations.
\newblock In \emph{2017 IEEE/RSJ International Conference on Intelligent Robots
  and Systems (IROS)}, pages 3932--3939. IEEE, 2017.

\bibitem[Mnih et~al.(2015)Mnih, Kavukcuoglu, Silver, Rusu, Veness, Bellemare,
  Graves, Riedmiller, Fidjeland, Ostrovski, et~al.]{mnih2015human}
Volodymyr Mnih, Koray Kavukcuoglu, David Silver, Andrei~A Rusu, Joel Veness,
  Marc~G Bellemare, Alex Graves, Martin Riedmiller, Andreas~K Fidjeland, Georg
  Ostrovski, et~al.
\newblock Human-level control through deep reinforcement learning.
\newblock \emph{Nature}, 518\penalty0 (7540):\penalty0 529--533, 2015.

\bibitem[Nachum and Dai(2020)]{nachum2020reinforcement}
Ofir Nachum and Bo~Dai.
\newblock Reinforcement learning via fenchel-rockafellar duality.
\newblock \emph{arXiv preprint arXiv:2001.01866}, 2020.

\bibitem[Oikarinen et~al.(2021)Oikarinen, Zhang, Megretski, Daniel, and
  Weng]{oikarinen2021robust}
Tuomas Oikarinen, Wang Zhang, Alexandre Megretski, Luca Daniel, and Tsui-Wei
  Weng.
\newblock Robust deep reinforcement learning through adversarial loss.
\newblock \emph{Advances in Neural Information Processing Systems},
  34:\penalty0 26156--26167, 2021.

\bibitem[Papernot et~al.(2016)Papernot, McDaniel, Jha, Fredrikson, Celik, and
  Swami]{papernot2016limitations}
Nicolas Papernot, Patrick McDaniel, Somesh Jha, Matt Fredrikson, Berkay-Z
  Celik, and Ananthram Swami.
\newblock The limitations of deep learning in adversarial settings.
\newblock In \emph{2016 IEEE European symposium on security and privacy
  (EuroS\&P)}, pages 372--387. IEEE, 2016.

\bibitem[Papernot et~al.(2017)Papernot, McDaniel, Goodfellow, Jha, Celik, and
  Swami]{papernot2017practical}
Nicolas Papernot, Patrick McDaniel, Ian Goodfellow, Somesh Jha, Berkay-Z Celik,
  and Ananthram Swami.
\newblock Practical black-box attacks against machine learning.
\newblock In \emph{Proceedings of the 2017 ACM on Asia conference on computer
  and communications security}, pages 506--519, 2017.

\bibitem[Pattanaik et~al.(2018)Pattanaik, Tang, Liu, Bommannan, and
  Chowdhary]{pattanaik2017robust}
Anay Pattanaik, Zhenyi Tang, Shuijing Liu, Gautham Bommannan, and Girish
  Chowdhary.
\newblock Robust deep reinforcement learning with adversarial attacks.
\newblock \emph{Proceedings of the International Joint Conference on Autonomous
  Agents and Multiagent Systems, AAMAS}, 3, 2018.

\bibitem[Peters et~al.(2010)Peters, Mulling, and Altun]{peters2010relative}
Jan Peters, Katharina Mulling, and Yasemin Altun.
\newblock Relative entropy policy search.
\newblock In \emph{Proceedings of the AAAI Conference on Artificial
  Intelligence}, volume~24, pages 1607--1612, 2010.

\bibitem[Raffin et~al.(2021)Raffin, Hill, Gleave, Kanervisto, Ernestus, and
  Dormann]{stable-baselines3}
Antonin Raffin, Ashley Hill, Adam Gleave, Anssi Kanervisto, Maximilian
  Ernestus, and Noah Dormann.
\newblock Stable-baselines3: Reliable reinforcement learning implementations.
\newblock \emph{Journal of Machine Learning Research}, 22\penalty0
  (268):\penalty0 1--8, 2021.
\newblock URL \url{http://jmlr.org/papers/v22/20-1364.html}.

\bibitem[Schaul et~al.(2015)Schaul, Quan, Antonoglou, and
  Silver]{schaul2015prioritized}
Tom Schaul, John Quan, Ioannis Antonoglou, and David Silver.
\newblock Prioritized experience replay.
\newblock \emph{arXiv preprint arXiv:1511.05952}, 2015.

\bibitem[Schulman et~al.(2015)Schulman, Levine, Abbeel, Jordan, and
  Moritz]{schulman2015trust}
John Schulman, Sergey Levine, Pieter Abbeel, Michael Jordan, and Philipp
  Moritz.
\newblock Trust region policy optimization.
\newblock In \emph{International conference on machine learning}, pages
  1889--1897. PMLR, 2015.

\bibitem[Schulman et~al.(2017)Schulman, Wolski, Dhariwal, Radford, and
  Klimov]{schulman2017proximal}
John Schulman, Filip Wolski, Prafulla Dhariwal, Alec Radford, and Oleg Klimov.
\newblock Proximal policy optimization algorithms.
\newblock \emph{arXiv preprint arXiv:1707.06347}, 2017.

\bibitem[Shen et~al.(2020)Shen, Li, Jiang, Wang, and Zhao]{shen2020deep}
Qianli Shen, Yan Li, Haoming Jiang, Zhaoran Wang, and Tuo Zhao.
\newblock Deep reinforcement learning with robust and smooth policy.
\newblock In \emph{International Conference on Machine Learning}, pages
  8707--8718. PMLR, 2020.

\bibitem[Silver et~al.(2017)Silver, Schrittwieser, Simonyan, Antonoglou, Huang,
  Guez, Hubert, Baker, Lai, Bolton, et~al.]{silver2017mastering}
David Silver, Julian Schrittwieser, Karen Simonyan, Ioannis Antonoglou, Aja
  Huang, Arthur Guez, Thomas Hubert, Lucas Baker, Matthew Lai, Adrian Bolton,
  et~al.
\newblock Mastering the game of go without human knowledge.
\newblock \emph{Nature}, 550\penalty0 (7676):\penalty0 354--359, 2017.

\bibitem[Simonyan and Zisserman(2014)]{simonyan2014vgg}
Karen Simonyan and Andrew Zisserman.
\newblock Very deep convolutional networks for large-scale image recognition.
\newblock \emph{arXiv preprint arXiv:1409.1556}, 2014.

\bibitem[Sun et~al.(2024)Sun, Gao, and Weng]{sun2024tactics}
Chung-En Sun, Sicun Gao, and Tsui-Wei Weng.
\newblock Tactics of robust deep reinforcement learning with randomized
  smoothing, 2024.
\newblock URL \url{https://openreview.net/forum?id=sRop0N5NYV}.

\bibitem[Sun et~al.(2020)Sun, Zhang, Xie, Ma, Zheng, Chen, and
  Liu]{sun2020stealthy}
Jianwen Sun, Tianwei Zhang, Xiaofei Xie, Lei Ma, Yan Zheng, Kangjie Chen, and
  Yang Liu.
\newblock Stealthy and efficient adversarial attacks against deep reinforcement
  learning.
\newblock In \emph{Proceedings of the AAAI Conference on Artificial
  Intelligence}, volume~34, pages 5883--5891, 2020.

\bibitem[Sun et~al.(2021)Sun, Zheng, Liang, and Huang]{sun2021strongest}
Yanchao Sun, Ruijie Zheng, Yongyuan Liang, and Furong Huang.
\newblock Who is the strongest enemy? towards optimal and efficient evasion
  attacks in deep rl.
\newblock \emph{arXiv preprint arXiv:2106.05087}, 2021.

\bibitem[Sutton and Barto(1998)]{sutton1998introduction}
Richard~S. Sutton and Andrew~G. Barto.
\newblock \emph{Introduction to reinforcement learning}.
\newblock MIT Press, 1998.
\newblock URL
  \url{http://webdocs.cs.ualberta.ca/~sutton/book/ebook/the-book.html,http://webdocs.cs.ualberta.ca/~sutton/book/the-book.html}.

\bibitem[Todorov et~al.(2012)Todorov, Erez, and Tassa]{todorov2012mujoco}
Emanuel Todorov, Tom Erez, and Yuval Tassa.
\newblock Mujoco: A physics engine for model-based control.
\newblock In \emph{2012 IEEE/RSJ International Conference on Intelligent Robots
  and Systems}, pages 5026--5033. IEEE, 2012.
\newblock \doi{10.1109/IROS.2012.6386109}.

\bibitem[Xu et~al.(2023)Xu, Jiang, Li, Yang, Wang, Chan, and
  Zhan]{xu2023offline}
Haoran Xu, Li~Jiang, Jianxiong Li, Zhuoran Yang, Zhaoran Wang, Victor Wai~Kin
  Chan, and Xianyuan Zhan.
\newblock Offline rl with no ood actions: In-sample learning via implicit value
  regularization.
\newblock \emph{arXiv preprint arXiv:2303.15810}, 2023.

\bibitem[Xu et~al.(2020)Xu, Shi, Zhang, Wang, Chang, Huang, Kailkhura, Lin, and
  Hsieh]{xu2020automatic}
Kaidi Xu, Zhouxing Shi, Huan Zhang, Yihan Wang, Kai-Wei Chang, Minlie Huang,
  Bhavya Kailkhura, Xue Lin, and Cho-Jui Hsieh.
\newblock Automatic perturbation analysis for scalable certified robustness and
  beyond.
\newblock \emph{Advances in Neural Information Processing Systems}, 33, 2020.

\bibitem[Zhang et~al.(2018)Zhang, Weng, Chen, Hsieh, and
  Daniel]{zhang2018efficient}
Huan Zhang, Tsui-Wei Weng, Pin-Yu Chen, Cho-Jui Hsieh, and Luca Daniel.
\newblock Efficient neural network robustness certification with general
  activation functions.
\newblock \emph{Advances in neural information processing systems}, 31, 2018.

\bibitem[Zhang et~al.(2019)Zhang, Chen, Xiao, Gowal, Stanforth, Li, Boning, and
  Hsieh]{zhang2019towards}
Huan Zhang, Hongge Chen, Chaowei Xiao, Sven Gowal, Robert Stanforth, Bo~Li,
  Duane Boning, and Cho-Jui Hsieh.
\newblock Towards stable and efficient training of verifiably robust neural
  networks.
\newblock \emph{arXiv preprint arXiv:1906.06316}, 2019.

\bibitem[Zhang et~al.(2020)Zhang, Chen, Xiao, Li, Liu, Boning, and
  Hsieh]{zhang2020robust}
Huan Zhang, Hongge Chen, Chaowei Xiao, Bo~Li, Mingyan Liu, Duane Boning, and
  Cho-Jui Hsieh.
\newblock Robust deep reinforcement learning against adversarial perturbations
  on state observations.
\newblock \emph{Advances in Neural Information Processing Systems},
  33:\penalty0 21024--21037, 2020.

\bibitem[Zhang et~al.(2021)Zhang, Chen, Boning, and Hsieh]{zhang2021robust}
Huan Zhang, Hongge Chen, Duane Boning, and Cho-Jui Hsieh.
\newblock Robust reinforcement learning on state observations with learned
  optimal adversary.
\newblock \emph{arXiv preprint arXiv:2101.08452}, 2021.

\end{thebibliography}

\clearpage

\appendix

\section*{APPENDIX}
\label{secALLappendix}


We describe additional related works, 
provide more detailed theoretical derivations, 
and offer additional evaluation results to help the reader gain a deeper understanding of our work. 
In the following, we temporarily set aside strict notation and represent expressions like "$\int_{a\in\mathcal{A}} \cdot , da$" as "$\sum_{a\in\mathcal{A}} \cdot$" when discussing continuous action space.

\section{Additional Related Works}
\subsection{f-Divergence Constrained Methods}
\label{subsecRelatedWork_FDivergence}
The use of optimization methods constrained by f-divergence has spanned various contexts from historical applications to recent advancements.
Historically, these methods were utilized for relative entropy maximization in inverse RL \citep{boularias2011relative}.
In offline RL and imitation learning (IL), they primarily serve to constrain divergence from the probability density distribution of state-action pairs within the dataset \citep{xu2023offline,kostrikov2019imitation,nachum2020reinforcement,kim2021demodice,kim2022lobsdice,ma2022versatile}.
Moreover, methods that limit update intervals by distribution constraints (old policy, prior) during policy updates \citep{peters2010relative,fox2015taming,schulman2015trust,schulman2017proximal,belousov2017f,belousov2019entropic} are closely related to our proposed soft-constrained adversary approach.

\section{Details of Soft Worst Attack (SofA) and SofA-SAC}
\label{secApp_softa_theory}

\subsection{Derivation of Soft Worst Attack (SofA)}
\label{subsecApp_softa_deriviation}

We derive Eq. \eqref{eq_optimal_softa} from Eq. \eqref{eq_soft_attack_def}.
At first, by flipping sign and converting $\argmin$ into $\argmax$ as:
\begin{equation}
\begin{aligned}
\label{eq_app_softa_derivation_worst_fdivergence}
\nu^{\star soft}_{\pi}(\tilde{s_t}|s_t) &=
  \argmin_{
      \textcolor{black}{
        \nu \in \mathcal{N}
        }
    }
  \mathop{\mathbb{E}}_{
    \tilde{s}_t \sim\nu(\cdot|s_t)
  }
  \left[
    \mathop{\mathbb{E}}_{
      \tilde{a}_t \sim\pi(\cdot|\tilde{s}_t)
      }
    \left[
      Q^{\pi}(s_t,\tilde{a}_t)
    \right]
  \right]
  \textcolor{black}{
  + \alpha_{attk} D_{f}
  \left(
    \nu(\cdot|s_t)\parallel p(\cdot|s_t)
    \right)
  }
  \\
  &=
  \textcolor{black}{
    - \argmax_{
        \textcolor{black}{\nu \in \mathcal{N}}
    }
  }
  \mathop{\mathbb{E}}_{
    \tilde{s}_t \sim\nu(\cdot|s_t)
  }
  \left[
  \mathop{\mathbb{E}}_{
    \tilde{a}_t \sim\pi(\cdot|\tilde{s}_t)
    }
    \left[
      (\textcolor{black}{-} Q^{\pi}(s_t,\tilde{a}_t))
    \right]
  \right]
  \textcolor{black}{ 
    \textcolor{black}{-} \alpha_{attk} D_{f}
    \left(
      \nu(\cdot|s_t)\parallel p(\cdot|s_t)
    \right)
    }
  .
\end{aligned}
\end{equation}
Here, we only consider the KL-divergence case, then we derive:
\begin{equation}
\label{eq_app_softa_derivation_worst_kldivergence}
\nu^{\star soft}_{\pi}(\tilde{s_t}|s_t) = 
  \textcolor{black}{
    - \argmax_{
        \textcolor{black}{\nu \in \mathcal{N}}
    }
  }
  \mathop{\mathbb{E}}_{
    \tilde{s}_t \sim\nu(\cdot|s_t)
  }
  \left[
  \mathop{\mathbb{E}}_{
    \tilde{a}_t \sim\pi(\cdot|\tilde{s}_t)
    }
    \left[
      (\textcolor{black}{-} Q^{\pi}(s_t,\tilde{a}_t))
    \right]
  \right]
  \textcolor{black}{ \textcolor{black}{-} \alpha_{attk} D_{KL}(\nu(\cdot|s_t)\parallel p(\cdot|s_t))}
  .
\end{equation}
This type of optimization problem is often utilized in various context \citep{boyd2004convex,peters2010relative,boularias2011relative,belousov2017f},
and there are two kinds of derivations, 
solving the Lagrangian with constraint or 
using the result of conjugate function of KL-divergence in the more general $f$($\alpha$)-divergence context.
In this section, we describe the former type of derivation.
This problem is described as follows:
\begin{equation}
\label{eq_app_softa_lagrangian_kl_coonstrained_softa_problem}
\begin{gathered}
\underset{
    \nu(\tilde{s}|s)
  }{
    \text{maximize}
    } \quad 
\sum_{\tilde{s}\in\mathcal{S}}
  \underset{
      \nu_{\tilde{s}}
  }{
    \underbrace{
      \nu(\tilde{s}|s) 
    }
  }
\underset{
    q_{\tilde{s}}
  }{
    \underbrace{
      \sum_{\tilde{a}\in\mathcal{A}} \pi(\tilde{a}|\tilde{s}) (-Q^{\pi}(s,\tilde{a})) 
    }
  }
- \alpha_{attk} \sum_{\tilde{s}\in\mathcal{S}} \nu(\tilde{s}|s)  (\log\nu(\tilde{s}|s) 
- \log 
\underset{
    p_{\tilde{s}}
}{
  \underbrace{
    p(\tilde{s}|s)
  }
}
)
\\
= \sum_{\tilde{s}\in\mathcal{S}} \nu_{\tilde{s}} q_{\tilde{s}} 
- \alpha_{attk} \sum_{\tilde{s}\in\mathcal{S}} \nu_{\tilde{s}} (\log \nu_{\tilde{s}} - \log p_{\tilde{s}}) := g(\boldsymbol{\nu})
\\
\text{subject to} \quad  
\forall\tilde{s}\in\mathcal{S}, \quad \nu_{\tilde{s}} \geq 0, \\
\sum_{\tilde{s}\in\mathcal{S}} \nu_{\tilde{s}} = 1.
\end{gathered}
\end{equation}
The Lagrangian function for this problem is given by \footnote[3]{
  We abbreviate the inequality constraint because it can be vanished in Eq. \eqref{eq_app_softa_stationary_condition} } :
\begin{equation}
\label{eq_app_softa_lagrangian}
L(\boldsymbol{\nu}, \boldsymbol{\lambda}) 
= g(\boldsymbol{\nu}) + \lambda (1 - \sum_{\tilde{s}\in\mathcal{S}} \nu_{\tilde{s}}),
\end{equation}
where $\lambda$ is the Lagrange multiplier associated with the equality constraint.
By rolling out the Karush-Kuhn-Tucker (KKT)'s stationary condition:
\begin{equation}
\begin{gathered}
\label{eq_app_softa_stationary_condition}
\nabla_{\boldsymbol{\nu}} g(\boldsymbol{\nu}) 
+ \lambda \nabla_{\boldsymbol{\nu}} g(\boldsymbol{\nu}) = \mathbf{0}  \\
\rightarrow 
\forall \tilde{s}\in\mathcal{S},\quad q_{\tilde{s}} - \alpha_{attk} ( \log \nu_{\tilde{s}} - \log p_{\tilde{s}} + 1) - \lambda = 0 \\
\rightarrow 
\log \nu_{\tilde{s}} =  \log p_{\tilde{s}} - 1  + \frac{ q_{\tilde{s}} - \lambda }{\alpha_{attk}} \\
\rightarrow 
\nu_{\tilde{s}} =  p_{\tilde{s}} \exp ( q_{\tilde{s}} / \alpha_{attk}) \exp (- \lambda / \alpha_{attk} - 1).
\end{gathered}
\end{equation}
By giving this result back into the equality constraint at Eq. \eqref{eq_app_softa_lagrangian_kl_coonstrained_softa_problem}, 
we can specify the Lagrange multiplier $\lambda$ and optimal value for $\nu_{\tilde{s}}$:
\begin{equation}
\begin{aligned}
\label{eq_app_softa_attacker_solution}
\nu^{\star}_{\tilde{s}} &=  \nu^{\star}(\tilde{s}|s) =  
\frac{
  p_{\tilde{s}} \exp ( q_{\tilde{s}} / \alpha_{attk})
  }{
  \sum_{\tilde{s}\in\mathcal{S}} p_{\tilde{s}} \exp ( q_{\tilde{s}} / \alpha_{attk})
  } \\
&=
\frac{
  p(\tilde{s}|s) \exp \{ \sum_{\tilde{a}\in\mathcal{A}}\pi(\tilde{a}|\tilde{s}) (-Q^{\pi}(s,\tilde{a}) ) / \alpha_{attk} \}
  }{
  \sum_{\tilde{s}\in\mathcal{S}} 
  p(\tilde{s}|s) \exp \{ \sum_{\tilde{a}\in\mathcal{A}}\pi(\tilde{a}|\tilde{s}) (-Q^{\pi}(s,\tilde{a}) ) / \alpha_{attk} \}
  }.
\end{aligned}
\end{equation}
This is the same equation as Eq. \eqref{eq_optimal_softa} and by giving $\nu^{\star}$ back into $g(\boldsymbol{\nu})$, 
the optimal value is:
\begin{equation}
\label{eq_app_softa_optimal_softa_value}
g(\boldsymbol{\nu^{\star}}) = \alpha_{attk} \log \sum_{\tilde{s}\in\mathcal{S}} p(\tilde{s}|s) 
\exp \{
  \sum_{\tilde{a}\in\mathcal{A}} \pi(\tilde{a}|\tilde{s}) (-Q^{\pi}(s,\tilde{a})/\alpha_{attk})
  \}.
\end{equation}

\subsection{Procedure and Characteristics of Practical Soft Worst Attack (SofA)}
\label{subsecApp_softa_procedure}

\begin{algorithm}[tb]
\caption{Soft Worst Attack (SofA) Sampling Method}
\label{alg:softa-sampling}
\begin{algorithmic}[1] 
 \renewcommand{\algorithmicrequire}{\textbf{Input:}}
 \renewcommand{\algorithmicensure}{\textbf{Output:}}
\Require{state $s_t$, policy $\pi(a|s)$, action-value function $Q^{\pi}(s,a)$, temperature parameter $\alpha_{attk}$, number of samples $N$, prior distribution function $p(\tilde{s}|s)$}
\Ensure{Soft worst sample $\tilde{s}_t$}
\Function{SofA}{$s_t, \pi, Q^{\pi}, \alpha_{attk}, N, p$}
  \State Sample $N$ times from the prior distribution $p(\tilde{s}|s)$:
    \Statex \quad\quad\quad $\tilde{s}_{ti} \sim p(\tilde{s}|s_t)$ for $i = 1, 2, \ldots, N$
  \State Estimate the action values for the perturbed states $\tilde{s}_{ti}$ and policy $\pi(a|s)$:
    \Statex \quad\quad\quad $\tilde{a}_{ti} \sim \pi(\cdot|\tilde{s}_{ti})$, $\tilde{Q}_{ti} = Q^{\pi}(s_t, \tilde{a}_{ti})$
  \State Calculate the probability of the Soft Worst Attack as in Eq. \eqref{eq_softa_sampling}:
    \Statex \quad\quad\quad $\nu(\tilde{s}_{ti}|s_t) = \frac{\exp(-\tilde{Q}_{ti}/\alpha_{attk})}{\sum_{j=1}^{N} \exp(-\tilde{Q}_{tj}/\alpha_{attk})}$
  \State Select one sample $\tilde{s}_t$ from $\tilde{s}_{ti}$ based on the calculated probability weights $\nu(\tilde{s}_{ti}|s_t)$:
    \Statex \quad\quad\quad (if $\alpha_{attk} \rightarrow 0$, select $i = \argmin_{i'} \tilde{Q}_{ti'}$)
  \State \Return $\tilde{s}_t$
\EndFunction
\end{algorithmic}
\end{algorithm}

To make clear vision of the practical soft worst attack, proposed in Section \ref{subsubsec_softa_attacker}, 
we depict Algorithm\ref{alg:softa-sampling} as a pseudo code.
When $N\rightarrow\infty$ and $\alpha_{attk}\rightarrow0$, as in the line 5, 
$\tilde{s}_t\rightarrow \argmin_{s'\in supp(p)}{\mathbb{E}_{\tilde{a}_t\sim\pi(\cdot|s')}\lbrack Q(s_t, \tilde{a}_t)}\rbrack$.
Therefore, this procedure can be seemed as a sampling based attack, 
while \textit{Critic} attack \citep{pattanaik2017robust,zhang2020robust} is the gradient iteration based attack.
Compared to the Critic attack, this attack,
\begin{enumerate}[(1)]
  \item is gradient iteration free (not use back-propagation, forward two times), and can calculate $N$ samples in parallel at once
  \item can consider more flexible shape of adversary even when $\text{dom}(\nu(\tilde{s}|s))$ is not continuous, not differentiable
  \item can incorporate realistic negative possibilities by choosing $N$ and $\alpha_{attk}$
\end{enumerate}
In cases where the state space is as large as an image, 
it may not be possible to sample the worst-case scenarios, 
making this method unsuitable for situations where cyber attacks are anticipated. 
However, \textbf{in realistic engineering and development contexts, 
where noise is often assumed to follow a normal distribution, 
we believe that the characteristic (2) and (3) are considered to be very important.}

\subsection{Derivation of Soft Worst Attack SAC (SofA-SAC)}
\label{subsecApp_softa_sac_derivation}

In this subsection, we describe derivation of SofA-SAC in Section \ref{subsubsec_softa_rl_kl}.
We assume that the policy only knows perturbed states, 
then the input of its max-entropy objective $\mathcal{H}$ is also perturbed as in Eq. \eqref{eq_soft_worst_sac_reward}.
To think one step rollout, we can build modified Bellman equation associated with the Bellman operator as:
\begin{equation}
\label{eq_app_softa_bellman_derivation}
(\mathcal{T}^{\pi}_{\nu}Q^{}_{})(s_t, a_t) \triangleq r(s_t, a_t) + 
\gamma
\mathbb{E}_{s_{t+1} \sim\mathcal{F}} 
  \left[
  \textcolor{black}{
  \mathbb{E}_{\tilde{s}_{t+1} \sim \nu} 
    \left[
    \mathbb{E}_{\tilde{a}_{t+1} \sim \pi}
      \left[
        Q^{}_{}(s_{t+1}, \tilde{a}_{t+1})
      \right]
      + \alpha_{ant} \mathcal{H}(\pi(\cdot|\tilde{s}_{i+1}))
    \right]
    }
  \right].
\end{equation}
Now we think the policy $\pi$ is fixed and the soft constrained optimal adversary:
\footnotesize
\begin{equation}
\begin{aligned}
\label{eq_app_softa_worst_bellman_derivation}
(\mathcal{\underline{T}}^{\pi}_{soft}Q^{}_{})(s_t, a_t)
&\triangleq
r(s_t, a_t)
+ \gamma
\mathbb{E}_{s_{t+1} \sim\mathcal{F}}
  \left[  \right. \\
  &
  \left.
  \min_{\nu \in \mathcal{N} } 
  \mathbb{E}_{\tilde{s}_{t+1} \sim \nu} 
    \left[
    \underline{
      \mathbb{E}_{\tilde{a}_{t+1} \sim \pi}
        \left[
          Q^{}_{}(s_{t+1}, \tilde{a}_{t+1})
        \right]
        + \alpha_{ant} \mathcal{H}(\pi(\cdot|\tilde{s}_{i+1}))
      }
      \right]
      + \alpha_{attk} D_{KL}( \nu \parallel  p)
  \right].
\end{aligned}
\end{equation}
\normalsize
The minimization problem in this equation can be solved similarly to Eq. \eqref{eq_app_softa_derivation_worst_kldivergence} in Appendix \ref{subsecApp_softa_deriviation}, by setting:
\begin{equation}
q_{\tilde{s}} \leftarrow \sum_{\tilde{a}\in\mathcal{A}} \pi(\tilde{a}|\tilde{s}) \left( - \left( Q^{}(s, \tilde{a}) - \alpha_{ent} \log \pi(\tilde{a}|\tilde{s}) \right) \right)
\end{equation}
Substituting this term into Eq. \eqref{eq_app_softa_attacker_solution} and Eq. \eqref{eq_app_softa_optimal_softa_value}, we derive Eq. \eqref{eq_optimal_softa_max_ent} and Eq. \eqref{eq_soft_worst_kl_bellman}.

For policy improvement, 
we consider the analytical solution to the maximization problem highlighted in the underlined part of Eq. \eqref{eq_app_softa_worst_bellman_derivation}:
\begin{equation}
\begin{aligned}
\label{eq_app_softa_policy_optimization_target}
\pi^{\star}(a|\tilde{s};s) &= \argmax_{\pi'}
      \mathbb{E}_{\tilde{a} \sim \pi'(\cdot|\tilde{s})}
        \left[
          Q^{\pi}_{}(s_{}, \tilde{a}_{})
        \right]
        + \alpha_{ent} \mathcal{H}(\pi'(\cdot|\tilde{s}_{})) \\
        &= 
        \frac{
          \exp \left( Q^{\pi}(s,\tilde{a})/\alpha_{ent} \right)
        }{
          \sum_{\tilde{a}'\in \mathcal{A}} \exp \left( Q^{\pi}(s,\tilde{a}')/\alpha_{ent} \right)
        }
\end{aligned}
\end{equation}
This maximization problem is identical to the original Soft Q-learning \citep{haarnoja2017reinforcement} 
and SAC papers \citep{haarnoja2018softa,haarnoja2018softb}, 
so we do not derive it here.
Then, we think policy update with the perturbation,
\begin{equation}
\begin{aligned}
\label{eq_app_softa_policy_optimization_update}
\mathbb{E}_{\tilde{s}\sim\nu(\cdot|s)}
\lbrack
  \pi^{}_{new}(\tilde{a}|\tilde{s})
\rbrack
=
(\pi_{new}\circ\nu)(\tilde{a}|\tilde{s})
\leftarrow
\pi^{\star} (\tilde{a} | \tilde{s};s) = 
  \frac{
    \exp \{ Q^{\pi}(s,\tilde{a})/\alpha_{ent} \}
  }{
    Z
  }
\end{aligned}
\end{equation}
Similar to the original SAC, 
this update is calculated by minimizing the KL-divergence between the left and right terms. 
Assuming the soft worst attacker $\nu^{\star}_{\pi_{ent}}$ for $\nu$, 
we derive Eq. \eqref{eq_soft_worst_sac_policy_loss1}.

\subsection{Contraction and Policy Improvement Properties of SofA-SAC}
\label{subsecApp_softa_sac_contraction_policy_improvement}

\paragraph{Contraction Property}

To demonstrate the reliability of our proposed SofA-SAC algorithm, we discuss its contraction property in this subsection. 
We revisit the definition provided in Eq. \eqref{eq_soft_worst_kl_bellman}:
\begin{equation}
\label{eq_app_softa_recall_bellman_def}
\begin{aligned}
\underline{\mathcal{T}}^{\pi}_{soft}Q(s_t, a_t) \triangleq 
&r(s_t, a_t) - 
\gamma \mathbb{E}_{s_{t+1}\sim\mathcal{F}} 
\left[ \alpha_{attk} \log \left( \mathbb{E}_{\tilde{s}_{t+1}\sim p} 
\left[ \right. \right. \right.\\
&\left. \left. \left. \exp \left( \mathbb{E}_{\tilde{a}_{t+1}\sim \pi} \left[ \frac{-( Q^{}(s_{t+1}, \tilde{a}_{t+1}) - \alpha_{ent} \log \pi(\tilde{a}_{t+1}|\tilde{s}_{t+1}) )}{\alpha_{attk}} \right] \right)
\right]
\right)
\right].
\end{aligned}
\end{equation}

\begin{thm}
The Soft Worst Bellman Operator $\underline{\mathcal{T}}^{\pi}_{soft}$ acts as a contraction operator for a fixed policy.
\label{thm:app_softa_bellman_contraction}
\end{thm}
\textit{Proof.} 

For simplicity in notation, we define:
\begin{equation}
f(Q(t)) \triangleq 
- \alpha_{attk} \log \sum_{\tilde{s}_t \in \mathcal{S}} p(\tilde{s}_t|s_t) 
\exp
\left(
  -\sum_{\tilde{a}_t\in\mathcal{A}} 
\pi(\tilde{a}_t|\tilde{s}_t) 
\left(
  Q(s_t,\tilde{a}_t)-\alpha_{ent}\log\pi(\tilde{a}_t|\tilde{s}_t)
\right)
/\alpha_{attk} 
\right)
\end{equation}
We assume there are two different action-value functions, $Q_1(s_t,a_t)$ and $Q_2(s_t,a_t)$,
and as the same metric in G-learning \citep{fox2015taming} and Soft Q-learning \citep{haarnoja2017reinforcement}, 
we define $\epsilon = ||Q_{1}(s_t,a_t) - Q_{2}(s_t,a_t)||_{s_t,a_t}$, here $||\dots||_{s_t,a_t}$ denotes the max norm over $s_t,a_t$.
Since $f(Q)$ is a monotonically increasing function for Q, then we can say:
\footnotesize
\begin{equation}
\label{eq_app_softa_contraction1}
\begin{aligned}
\forall s_{t},& f(Q_{1}(t)) \leq f(Q_{2}(t) + \epsilon)
\\
&= 
-\alpha_{attk} \log \sum_{\tilde{s}_t \in \mathcal{S}}p(\tilde{s}_t|s_t) \exp
\left( -\sum_{\tilde{a}_t \in \mathcal{A}} \pi(\tilde{a}_t|\tilde{s}_t)
\left(Q_{2}(s_t,\tilde{a}_t) + \epsilon -\alpha_{ent}\log\pi(\tilde{a}_t|\tilde{s}_t)
\right)/\alpha_{attk} 
\right)
\\
&= 
- \alpha_{attk} \log \sum_{\tilde{s}_t \in \mathcal{S}}p(\tilde{s}_t|s_t) \exp
\left(
  -\epsilon/\alpha_{attk} -\sum_{\tilde{a}_t \in \mathcal{A}} \pi(\tilde{a}_t|\tilde{s}_t)
  \left(Q_{2}(s_t,\tilde{a}_t) -\alpha_{ent}\log\pi(\tilde{a}_t|\tilde{s}_t)
  \right)/\alpha_{attk} 
\right)
\\
&= 
\epsilon - \alpha_{attk} \log \sum_{\tilde{s}_t \in \mathcal{S}}p(\tilde{s}_t|s_t) 
\exp
\left(
  -\sum_{\tilde{a}_t \in \mathcal{A}} \pi(\tilde{a}_t|\tilde{s}_t)
  \left(
    Q_{2}(s_t,\tilde{a}_t) -\alpha_{ent}\log\pi(\tilde{a}_t|\tilde{s}_t)
  \right)/\alpha_{attk}
\right)
\\
&= 
|| Q_1(s_t, a_t)-Q_2(s_t, a_t) ||_{s_t,a_t} + f(Q_{2}(t))
\\
&\leftrightarrow
f(Q_{1}(t)) - f(Q_{2}(t))
\leq ||Q_1(s_t, a_t) -Q_2(s_t, a_t) ||_{s_t,a_t}.
\end{aligned}
\end{equation}
\normalsize
In the same way:
\begin{equation}
\label{eq_app_softa_contraction2}
\begin{aligned}
&f(Q_{1}(t)) \geq f(Q_{2}(t) - \epsilon) = 
f(Q_{2}(t)) - ||Q_1(s_t,a_t)-Q_2(s_t,a_t)||_{s_t,a_t}
\\
&\leftrightarrow
f(Q_1(t)) - f(Q_2(t)) \geq -||Q_1(s_t,a_t)-Q_2(s_t,a_t)||_{s_t,a_t}.
\end{aligned}
\end{equation}
Therefore, from the both inequalities, we derive:
\begin{equation}
\label{eq_app_softa_contraction3}
||f(Q_1(t)) - f(Q_2(t))||_{s_t} \leq ||Q_1(s_t,a_t)-Q_2(s_t,a_t)||_{s_t,a_t}.
\end{equation}

Then, we consider difference between the two action-value functions after our Bellman operator:
\begin{equation}
\label{eq_app_softa_contraction4}
\begin{aligned}
|| \underline{\mathcal{T}}_{soft}^{\pi}Q_{1}(s_t,a_t)
- \underline{\mathcal{T}}_{soft}^{\pi} Q_{2}(s_t,a_t) ||_{s_t,a_t}
&= 
||
\gamma \mathbb{E}_{{s}_{t+1}\sim \mathcal{F}} 
\lbrack
f(Q_{1}(t+1)) -f(Q_{2}(t+1))
\rbrack
||_{s_t,a_t} \\
&\leq
\gamma
||
f(Q_{1}(t+1)) -f(Q_{2}(t+1))
||_{s_{t+1}} \\
&\underset{(\ref{eq_app_softa_contraction3})}{\leq}
\gamma
||
Q_{1}(s_{t+1},a_{t+1}) - Q_{2}(s_{t+1}, a_{t+1})
||_{s_{t+1},a_{t+1}}.
\end{aligned}
\end{equation}
Therefore, we can say $\underline{\mathcal{T}}_{soft}^{\pi}$ is a contraction operator,
because we perform this operation an infinite number of times, 
$Q_{1}$ and $Q_{2}$ converge to a fixed point. \qed

\paragraph{Policy Improvement Property (with a fixed adversary)}

If we once fix the attacker $\nu$, 
we can discuss the policy improvement theorem \citep{sutton1998introduction} as the same manner in \cite{haarnoja2017reinforcement,haarnoja2018softa,haarnoja2018softb}.
For simplicity, we omit the entropy coefficient $\alpha_{ent}$ in this paragraph.
\begin{thm}[Policy Improvement Theorem with a Fixed Adversary]
\label{thm:app_softa_policy_improvement}

Given a fixed adversary $\nu$ and a policy $\pi$, define a new policy $\hat{\pi}$ such that for all states $s$,
\begin{equation}
\label{eq_app_softa_policy_improvement_new_policy_def}
\hat{\pi} \circ \nu(\cdot|s) \propto \exp(Q^{\pi}_{\nu}(s, \cdot)).
\end{equation}
Assuming that $Q^{\pi}_{\nu}$ and $\sum_{a\in\mathcal{A}} \exp(Q^{\pi}_{\nu}(s,a))$ are bounded for all $s$, 
it follows that $Q^{\hat{\pi}}_{\nu}(s,a) \geq Q^{\pi}_{\nu}(s,a)$ 
for all actions $a$ and state $s$.
\end{thm}

\textit{Proof.}

If we extract a new policy $\hat{\pi}$ by using $\pi$ and the corresponding action-value function $Q^{\pi}_{\nu}$,
a following equation holds:
\begin{equation}
\label{eq_app_softa_policy_improvement_inequality}
\begin{aligned}
\mathbb{E}_{\tilde{s}\sim\nu}
\lbrack 
  \mathcal{H}(\pi(\cdot|\tilde{s})) + \mathbb{E}_{\tilde{a}\sim\pi}
  \lbrack 
    Q^{\pi}_{\nu}(s,\tilde{a})
  \rbrack 
\rbrack 
\leq
\mathbb{E}_{\tilde{s}\sim\nu}
\lbrack 
  \mathcal{H}(\textcolor{black}{\hat{\pi}}(\cdot|\tilde{s})) + \mathbb{E}_{\tilde{a}\sim\textcolor{black}{\hat{\pi}}}
  \lbrack 
    Q^{\pi}_{\nu}(s,\tilde{a})
  \rbrack 
\rbrack  \\
\leftrightarrow
  \mathcal{H}(\pi\circ\nu(\cdot|{s})) + \mathbb{E}_{\tilde{a}\sim\pi\circ\nu}
  \lbrack 
    Q^{\pi}_{\nu}(s,\tilde{a})
  \rbrack 
\leq
  \mathcal{H}(\textcolor{black}{\hat{\pi}}\circ\nu(\cdot|{s})) 
  + \mathbb{E}_{\tilde{a}\sim\textcolor{black}{\hat{\pi}}\circ\nu}
  \lbrack 
    Q^{\pi}_{\nu}(s,\tilde{a})
  \rbrack. 
\end{aligned}
\end{equation}
This is proven by: 
\begin{equation}
\label{eq_app_softa_policy_improvement_kl_old_new}
\begin{aligned}
  D_{KL}(\pi\circ\nu(\cdot|s) &\parallel  \textcolor{black}{\hat{\pi}} \circ \nu(\cdot|s) )
  = 
  \sum_{\tilde{a}\in\mathcal{A}} \pi\circ\nu(\tilde{a}|s) 
  \left(
    \log {\pi\circ\nu(\tilde{a}|s)}
    -\log {
      \textcolor{black}{\hat{\pi}} \circ \nu(\tilde{a}|s)
    }
  \right)
  \\
  &=
  -\mathcal{H}(\pi\circ\nu( \cdot | {s} ))
  -\sum_{\tilde{a} \in \mathcal{A}} \pi\circ\nu(\tilde{a}|s) 
  \left(
    \log {
      \textcolor{black}{\hat{\pi}} \circ \nu(\tilde{a}|s)
    }
  \right)
  \\
  &=
  -\mathcal{H}(\pi\circ\nu(\cdot|{s}))
  -\sum_{\tilde{a}\in\mathcal{A}} \pi\circ\nu(\tilde{a}|s)
  \left(
    Q^{\pi}_{\nu}(s,\tilde{a})
      - \log \sum_{\tilde{a}' \in \mathcal{A}} \exp \{Q^{\pi}_{\nu}(s, \tilde{a}') \}
  \right)
  \\
  &=
  -\mathcal{H}(\pi\circ\nu(\cdot|{s}))
  -\mathbb{E}_{\tilde{a}\sim \pi\circ\nu(\cdot|s)} 
  \left[
    Q^{\pi}_{\nu}(s,\tilde{a})
  \right]
  + \log \sum_{\tilde{a}'\in\mathcal{A}} \exp 
  \left(
    Q^{\pi}_{\nu}(s, \tilde{a}')
  \right)
  \\
  &=
  -\lbrack \text{LHS of \eqref{eq_app_softa_policy_improvement_inequality}} \rbrack
  +\lbrack \text{RHS of \eqref{eq_app_softa_policy_improvement_inequality}} \rbrack
  \geq 0.
\end{aligned}
\end{equation}
In the third equality, we used the result defined in Theorem\ref{thm:app_softa_policy_improvement}:
\begin{equation}
\label{eq_app_softa_policy_improve_def_target}
\begin{aligned}
  \textcolor{black}{\hat{\pi}}\circ\nu(\tilde{a}|s) 
  = \frac{
    \exp Q^{\pi}_{\nu}(s, \tilde{a})
  }{
    \sum_{\tilde{a}'\in\mathcal{A}} \exp Q^{\pi}_{\nu}(s, \tilde{a}')
  }.
\end{aligned}
\end{equation}
By using this result, we can confirm:
\begin{equation}
\begin{aligned}
\label{eq_app_softa_policy_improve_rollout_Q}
  Q^{\pi}_{\nu}(s_t, \tilde{a}_t)
  &= 
    r(s_t, \tilde{a}_t) +
    \gamma
    \mathbb{E}_{s_{t+1}\sim\mathcal{F}}
    \lbrack
      \mathbb{E}_{\tilde{s}_{t+1}\sim\mathcal{\nu}}
      \lbrack
        \mathbb{E}_{\tilde{a}_{t+1}\sim\pi}
        \lbrack
          Q^{\pi}_{\nu}(s_{t+1}, \tilde{a}_{t+1})
        \rbrack
          + \mathcal{H}(\pi(\cdot|\tilde{s}_{t+1}))
      \rbrack
    \rbrack \\
  &\leq
    r(s_t, \tilde{a}_t) +
    \gamma
    \mathbb{E}_{s_{t+1}\sim\mathcal{F}}
    \lbrack
      \mathbb{E}_{\tilde{s}_{t+1}\sim\mathcal{\nu}}
      \lbrack
        \mathbb{E}_{\tilde{a}_{t+1}\sim\textcolor{black}{\hat{\pi}}}
        \lbrack
          \underset{\text{rollout}}{
          \underline{
            Q^{\pi}_{\nu}(s_{t+1}, \tilde{a}_{t+1})
            }}
        \rbrack
          + \mathcal{H}(\textcolor{black}{\hat{\pi}}(\cdot|\tilde{s}_{t+1}))
      \rbrack
    \rbrack \\
  &\quad\vdots \\
  &\leq Q^{\textcolor{black}{\hat{\pi}}}_{\nu}(s_t,\tilde{a}_t).
\end{aligned}
\end{equation}
Then, under a fixed adversary $\nu$, we can improve our policy $\pi\circ\nu$ by targeting Eq. \eqref{eq_app_softa_policy_improve_def_target}.
\qed

As the discussion in \cite{zhang2020robust}, 
we cannot prove optimality for the resulting policy due to the adversary's movement.
However, we can train our algorithm (SofA-SAC) stably by pausing the adversary during the policy improvement, using the stop-gradient function as described in Eq. \eqref{eq_soft_worst_sac_policy_loss1}. 
This approach ensures stability in the training process.

\subsection{Practical Implementation of SofA-SAC}
\label{subsecApp_softa_practical_implementation}

Algorithm \ref{alg:softa-sac-implementation} presents a pseudocode representing the practical implementation of the SofA-SAC algorithm described in Section \ref{subsubsec_softa_rl_kl}. 
In lines 5 and 6, $done_t$ is referred to as a binary flag indicating whether the task has ended at the end of step $t$. 
We basically follow learning process as in the original SAC \citep{haarnoja2018softb}, 
use two critic networks with target networks and one policy network without a target network. 
We find that, though SofA-SAC also estimate a target value for the critic in the pessimistic way, 
taking min value of the two critic outputs for the critic targets and the actor objectives is needed for stable learning. 

Only in Walker2d, we observe that as the training proceeds, Q-values diverged  in some seeds (about 2-3 seeds in 8 seeds), 
then we put two measures to stabilize only for Walker2d (line11, line16 in Algorithm \ref{alg:softa-sac-implementation}) .
This may be because Walker2d is a sensitive task and sometimes experiences drops in scores even when using Vanilla-SAC. 
We assume once some unintended large values from a DNN are used for updates, it cannot recover from the fallen states.

\begin{algorithm}[tb]
\caption{SofA-SAC Training}
\label{alg:softa-sac-implementation}
\begin{algorithmic}[1]
\State Initialize critic $Q_{\theta_{1,2}}(s,a)$ and actor $\pi_{\phi}(s)$
\State Initialize target networks $Q_{\theta_{1,2}'}(s,a)$ by setting $\theta_{1,2}' \leftarrow \theta_{1,2}$

\State Initialize a replay buffer $\mathcal{R} \leftarrow \emptyset$ and entropy coefficient $\alpha_{ent} \leftarrow 1.0$
\For{$t = 1$ to $T$}
  \State Execute action $a_t \sim \pi_{\phi}(s_t)$, observe $(s_t, a_t, r_t, s_{t+1}, \textit{done}_t)$, and store in $\mathcal{R}$
  \State Sample a mini-batch of $M$ transitions $(s_t^i, a_t^i, r_t^i, s_{t+1}^i, \textit{done}_t^i) \sim \mathcal{R}$
  \Statex \quad \textbf{- Update Critic:}
  \State Sample $N$ perturbed states $\tilde{s}_{t+1}^{ij} \sim p(\cdot | s_{t+1}^i)$ for each $i$
  \State Sample perturbed actions $\tilde{a}_{t+1}^{ij} \sim \pi(\cdot | \tilde{s}_{t+1}^{ij})$ for each $ij$
  \State Estimate target values for each sample:
  \Statex \qquad $v^{ij}_{1,2}=
    Q_{\theta'_{1,2}}(s_{t+1}^{i},\tilde{a}^{ij}_{t+1}) 
    - \alpha_{ent} \log \pi_{\phi}(\tilde{a}^{ij}_{t+1}|\tilde{s}^{ij}_{t+1}) 
    $
  \State Compute targets by averaging over $N$ using log-sum-exp for stability:
  \Statex \qquad
  $y^i_{1,2} = r_t^i + (1-\textit{done}^i_t) \gamma \left(-\alpha_{attk} \log \frac{1}{N} \sum_{j=1}^N \exp\left(\frac{-v_{1,2}^{ij}}{\alpha_{attk}}\right)\right)$

  \State \qquad (- optional for Walker2d, 
    clip $y^{i}_{1,2}$ by percentile values over $N$ samples)

  \State Update $\theta_{1,2}$ by minimizing HuberLoss:
  \Statex \qquad
    $L(\theta_{1,2}) = \frac{1}{M} \sum_{i=1}^{M} L_\text{Hu}\left(Q_{\theta_{1,2}}(s^{i}_t, a^{i}_t), \min\left(y_{1}^{i},y_{2}^{i}\right) \right)$

  \Statex \quad \textbf{- Update Actor:}
  \State Sample $N$ perturbed states $\tilde{s}_{t}^{ij} \sim p(\cdot | s_{t}^i)$ for each $i$
  \State For each sample, sample actions for loss and importance weight calculation: 
  \Statex \qquad
    $\tilde{a}^{ij}_{t}, \tilde{a}^{ij'}_{t} \sim \pi_{\phi}(\cdot | \tilde{s}^{ij}_t)$

  \State Calculate loss and importance weight:
  \Statex \qquad
  $L^{ij} = \alpha_{ent} \log\pi_{\phi}(\tilde{a}^{ij}_{t}|\tilde{s}^{ij}_t ) - \min_{\theta_{1,2}} Q_{\theta_{1,2}}^{}(s^{i}_t, \tilde{a}^{ij}_t)$
  \Statex \qquad
  $w^{ij} = \text{softmax}_{j'}\left( 
    \left( 
      \alpha_{ent} \log\pi_{\phi}(\tilde{a}^{ij'}_{t}|\tilde{s}^{ij}_t ) - \text{mean}_{\theta_{1,2}} Q_{\theta_{1,2}}^{}(s^{i}_t, \tilde{a}^{ij'}_t) 
    \right)
  /\alpha_{attk} 
  \right)$
  \State \qquad (- optional for Walker2d, 
    ignore importance weight: $w^{ij}\leftarrow1$)

  \State Update $\phi$ using policy loss and importance weight:
  \Statex \qquad \qquad 
  $L(\phi_{}) = \frac{1}{MN} \sum_{i=1}^{M}\sum_{j=1}^{N} sg(w^{ij}) \bigodot L^{ij}$

  \Statex \quad \textbf{- Update Entropy Coefficient:}
  \State Re-use the entropy values during policy improvement and update $\alpha_{ent}$:
  \Statex \qquad
  $\mathcal{H}_{current}^{i} = -\frac{1}{N} \sum_{j=1}^{N} \log\pi_{\phi}(\tilde{a}^{ij}_{t}|\tilde{s}^{ij}_t )$
  \Statex \qquad
  $L(\alpha_{ent}) = -\frac{1}{M} \sum_{i=1}^{M} \alpha_{ent} (
    \mathcal{H}_{target} 
    -\mathcal{H}^{i}_{current} 
    )$
  \Statex \quad \textbf{- Post Processing:}
  \State Soft update the target networks: $\theta_{1,2}' \leftarrow (1-\tau) \theta_{1,2}' + \tau \theta_{1,2}$
  \State (-optional for Hopper and Ant, update the temperature $\alpha_{attk}$ according to the scheduling)
\EndFor
\end{algorithmic}
\end{algorithm}

\clearpage

\section{Details of Epsilon Worst Attack (EpsA) and EpsA-SAC}
\label{secApp_Epsilon_theory}

\subsection{Derivation (Approximation) of Epsilon Worst Attack (EpsA)}
\label{subsecApp_epsa_derivation_approximation}

If we do not limit ourselves to only the KL-divergence case but consider a more general $\alpha$-divergence in Definition \ref{dfn_soft_constrained_adversary}, 
we can obtain a broader perspective on the worst-case attack. 
This approach is inspired by \cite{belousov2017f,belousov2019entropic}, 
which provides more detail about the derivation and related perspectives on policy improvement.

An $f$-divergence is a measurements between two distribution and defined as:
\begin{equation}
\label{eq_app_epsa_fdivergence_def}
\begin{aligned}
D_{f}( \nu \parallel  p) \triangleq \sum_{\tilde{s}\in \mathcal{S}} p(\tilde{s}|s) f(\frac{\nu(\tilde{s}|s)}{p(\tilde{s}|s)}) \quad (s \text{ is given}).
\end{aligned}
\end{equation}
Here, $f(\cdot):\Omega\rightarrow\mathbb{R}$ is a convex function with the properties, $\text{Range}(f) = (0, \infty)$ and $f(x')=0 \leftrightarrow x'=1$.
An $\alpha$-divergence is a sub-family of the $f$-divergence that is defined as:
\begin{equation}
\label{eq_app_epsa_alphadivergence_def}
\begin{aligned}
f_{\alpha}(x) \triangleq \frac{
(x^{\alpha}-1)-\alpha(x-1)
}{
\alpha (\alpha - 1)
}
\end{aligned}
\end{equation}
The $\alpha$-divergence includes many popular divergence: 
for example,
the case, $\alpha\rightarrow0$, results in Reverse-KL-divergence, 
$\alpha\rightarrow1$ derive KL-divergence, 
and $\alpha=2$ occurs (Pearson's) $\chi^2$-divergence.
To solve dual problems, the convex conjugate function $f^*(y)$ for $f(x)$ is known to be useful. 
This is defined as:
\begin{equation}
\label{eq_app_epsa_conjugate_def}
\begin{aligned}
f^*(y) = \text{sup}_{x\in \text{dom}(f)} \{ \langle y,x \rangle -f(x) \},
\end{aligned}
\end{equation}
where $\langle \cdot, \cdot \rangle$ denotes the inner dot product.
If $f$ and $f^{*}$ are differentiable, 
by taking the derivative of Eq. \eqref{eq_app_epsa_conjugate_def}, 
$\nabla_{y}f^{*}(y) = x^{\star}$ and $\nabla_{x} \{\langle y,x^{}\rangle -f(x^{})\} |_{x=x^{\star}}=\mathbf{0}$, 
we obtain the property $(f^{*})'=(f')^{-1}$.

For the $\alpha$-divergence case, 
its conjugate function and the derivative function of the conjugate are:
\begin{equation}
\label{eq_app_epsa_conjugate_alpha}
\begin{aligned}
f_{\alpha}^*(y) = \frac{1}{\alpha}(1+(\alpha-1)y)^{\frac{\alpha}{\alpha -1}} -\frac{1}{\alpha},
\end{aligned}
\end{equation}
\begin{equation}
\label{eq_app_epsa_conjugate_derivative_alpha}
\begin{aligned}
(f_{\alpha}^*)'(y) = (1+(\alpha-1)y)^{\frac{1}{\alpha-1}}, \text{ for } (1-\alpha)y < 1.
\end{aligned}
\end{equation}
To consider the case described in Eq. \eqref{eq_app_softa_derivation_worst_fdivergence},
we can rewrite the problem as:
\begin{equation}
\label{eq_app_epsa_lagrangian_alpha_problem}
\begin{gathered}
\underset{
    \nu(\tilde{s}|s)
  }{
    \text{maximize}
    } \quad 
\sum_{\tilde{s} \in \mathcal{S}}
  \underset{
      \nu_{\tilde{s}}
  }{
    \underbrace{
      \nu(\tilde{s}|s) 
    }
  }
\underset{
    q_{\tilde{s}}
  }{
    \underbrace{
      \sum_{\tilde{a}\in\mathcal{A}} \pi(\tilde{a}|\tilde{s}) (-Q^{\pi}(s,\tilde{a})) 
    }
  }
- \alpha_{attk} \sum_{\tilde{s} \in \mathcal{S} } 
    p(\tilde{s}|s)
    f_{\alpha}(
      \underset{
        \frac{\nu_{\tilde{s}}}{p_{\tilde{s}}}
      }{
        \underbrace{
          \frac{\nu(\tilde{s}|s)}{p(\tilde{s}|s)}
        }
      }
      )
\\
= \sum_{\tilde{s} \in \mathcal{S}} \nu_{\tilde{s}} q_{\tilde{s}} 
- \alpha_{attk} \sum_{\tilde{s}\in \mathcal{S}} 
    p_{\tilde{s}}
    f_{\alpha}(\frac{\nu_{\tilde{s}}}{p_{\tilde{s}}})
  := g(\boldsymbol{\nu})
\\
\text{subject to} \quad  
\forall\tilde{s}\in\mathcal{S}, \quad \nu_{\tilde{s}} \geq 0, \\
\sum_{\tilde{s}\in\mathcal{S}} \nu_{\tilde{s}} = 1.
\end{gathered}
\end{equation}
The Lagrangian for this problem is given by:
\begin{equation}
\label{eq_app_epsa_lagrangian1}
\begin{gathered}
L(\boldsymbol{\nu}, \boldsymbol{\lambda}, \boldsymbol{\kappa}) 
= 
g(\boldsymbol{\nu}) 
+ \lambda (1 - \sum_{\tilde{s}\in\mathcal{S}} \nu_{\tilde{s}}) 
+ \sum_{\tilde{s}\in\mathcal{S}} \kappa_{\tilde{s}} \nu_{\tilde{s}},
\\
\text{with the KKT's condition:} \quad  
\forall\tilde{s}\in\mathcal{S},\\ 
\nabla_{\boldsymbol{\nu}} g(\boldsymbol{\nu}) - \lambda + \kappa_{\tilde{s}} = 0, \\
\nu_{\tilde{s}} \geq 0, \\
\kappa_{{\tilde{s}}}\nu_{\tilde{s}} = 0, \\
\kappa_{{\tilde{s}}} \geq 0.
\end{gathered}
\end{equation}
Here, $\kappa_{\tilde{s}} := \kappa(\tilde{s}|s)$ represents the complementary slackness for the inequality constraint. 
From the stationarity condition, $\forall \tilde{s}$:
\begin{equation}
\label{eq_app_epsa_stationarity_rollout}
\begin{gathered}
q_{\tilde{s}} 
- \alpha_{attk} f_{\alpha}'(\frac{\nu_{\tilde{s}}}{p_{\tilde{s}}})
- \lambda + \kappa_{\tilde{s}} = 0 \\
\rightarrow
f_{\alpha}'(\frac{\nu_{\tilde{s}}}{p_{\tilde{s}}})
=\frac{q_{\tilde{s}} - \lambda + \kappa_{\tilde{s}}
}{\alpha_{attk}}
\rightarrow
\nu_{\tilde{s}}
=
p_{\tilde{s}}
(f_{\alpha}')^{-1}
\left(
\frac{q_{\tilde{s}} - \lambda + \kappa_{\tilde{s}}
}{\alpha_{attk}}
\right).
\end{gathered}
\end{equation}
Then, using the property, where $(f^*)' = (f')^{-1}$,
we derive the optimal $\nu_{\tilde{s}}$ as:
\begin{equation}
\label{eq_app_epsa_optimal_alpha_nu}
\begin{gathered}
\nu^{\star}(\tilde{s}|s) 
= p(\tilde{s}|s) 
(f_{\alpha}^*)'
\left(
  \frac{
    \sum_{\tilde{a}\in\mathcal{A}} \pi(\tilde{a}|\tilde{s}) 
    \left(
      -Q^{\pi}(s,\tilde{a})
    \right) 
    - \lambda^{\star} + \kappa(\tilde{s}|s)
    }{\alpha_{attk}}
\right).
\end{gathered}
\end{equation}
This solution can be regarded as how to asign probability mass on the prior distribution $p(\tilde{s}|s)$ 
and it depends on $\pi, Q^{\pi}, \alpha_{attk}$, and $\alpha$.
If we assume only $\alpha<1$ case, 
from Eq. \eqref{eq_app_epsa_conjugate_derivative_alpha}, 
the term of $(f^{*}_{\alpha})'$ must hold $>0$, then $p(\tilde{s}|s)>0 \leftrightarrow \nu^{\star}(\tilde{s}|s)>0$ holds.
From the complementary condition in Eq. \eqref{eq_app_epsa_lagrangian1}, in this case, the slackness parameter $\kappa(s|{\tilde{s}})$ must be 0.
And from the constraint in Eq. \eqref{eq_app_epsa_conjugate_derivative_alpha}, 
we get the condition for $\lambda$:
\begin{equation}
\label{eq_app_epsa_condition_lambda}
\begin{gathered}
\forall \tilde{s}\in\mathcal{S}, \lambda^{} > \sum_{\tilde{a}\in\mathcal{A}}\pi(\tilde{a}|\tilde{s})(-Q^{\pi}(s,\tilde{a})) - \alpha_{attk}\frac{1}{1-\alpha}.
\end{gathered}
\end{equation}
This inequality must hold for all $\tilde{s}$, thus in the case of maximum of the right term.
Then, we define:
\begin{equation}
\label{eq_app_epsa_lambda_star_approx}
\begin{gathered}
\lambda^{\star} := 
\text{max}_{\tilde{s}} \sum_{\tilde{a}\in\mathcal{A}}\pi(\tilde{a}|\tilde{s})
  (-Q^{\pi}(s,\tilde{a})) 
  - \alpha_{attk}\frac{1}{1-\alpha} + \xi_\alpha,
\end{gathered}
\end{equation}
where $\xi_\alpha$ is a residual that satisfies $\xi_\alpha > 0$. 
Considering the case if $\alpha\rightarrow-\infty$,
the optimal value of constraint term approach the bound, $\lambda^{\star}\rightarrow \max_{\tilde{s} \in supp(p)}{\sum_{\tilde{a}\in\mathcal{A}}\pi(\tilde{a}|\tilde{s})(-Q^{\pi}(s,\tilde{a}))}$.
Therefore, $\xi_\alpha$ approaches to zero as $\alpha \rightarrow -\infty$. 
Substituting this back into Eq. \eqref{eq_app_epsa_optimal_alpha_nu} and setting 
$\xi(Q; \tilde{s}) := -\min_{\tilde{s}\in\text{supp}(p)} \sum_{\tilde{a}\in\mathcal{A}} \pi(\tilde{a}|\tilde{s}) Q^{\pi}(s, \tilde{a}) + \sum_{\tilde{a}\in\mathcal{A}} \pi(\tilde{a}|\tilde{s}) Q^{\pi}(s, \tilde{a})$, 
we can consider the cases where $\alpha \ll 0$ as follows:
\begin{equation}
\label{eq_app_epsa_nu_solution_approx}
\begin{aligned}
\nu^{\star}(\tilde{s}|s) 
&= p(\tilde{s}|s) 
(f_{\alpha}^*)'
  \left(
    \frac{
      -\xi(Q;\tilde{s}) + \alpha_{attk}\frac{1}{1-\alpha} -\xi_\alpha
      }{\alpha_{attk}}
  \right)
\\
&= p(\tilde{s}|s) 
\left(
  \frac{1-\alpha}{\alpha_{attk}} (\xi_\alpha + \xi(Q^{\pi};\tilde{s}))
\right)^{\frac{1}{\alpha-1}}\\
&= 
p(\tilde{s}|s) 
\left(
  \frac{\alpha_{attk}}{1-\alpha} \frac{1}{
    \xi_\alpha + \xi(Q^{\pi};\tilde{s})
    }
\right)^{\frac{1}{1-\alpha}}. 
\end{aligned}
\end{equation}
From this equation, we can determine that $\nu^{\star}(\tilde{s}|s)$ exhibits a strong peak at $\xi(Q^{\pi};\tilde{s})=0$, which corresponds to $\tilde{s}^{\star} := \argmin_{\tilde{s}\in \text{supp}(p)} \left(\sum_{\tilde{a}\in\mathcal{A}} \pi(\tilde{a}|\tilde{s}) Q^{\pi}(s,\tilde{a})\right)$. 
It also shows a mild probability mass for the other perturbation states 
due to the term of exponential, $0<\frac{1}{1-\alpha}\ll1$. 

Now, we assume the prior $p(\tilde{s}|s)$ is the uniform distribution over $L_{\infty}$-norm constrained range.
Then, we approximate the peak of the probability by 
a constant multiple of Dirac's delta function as $\kappa_{worst} \delta (\tilde{s}^{\star})$ 
and distribute the remaining probability equally as $1 - \kappa_{worst}$.
We represent this approximation as:
\begin{equation}
\label{eq_app_epsa_nu_final_approx_epsilon}
\nu^{\star}(\tilde{s}|s) \simeq
\begin{cases}
\kappa_{worst}
+\frac{1-\kappa_{worst}}{|\mathcal{S}_\epsilon|}
, \quad \text{if }\tilde{s}=\argmin_{\tilde{s}'\in \mathcal{B}_{\epsilon}}\sum_{\tilde{a}\in\mathcal{A}}\pi(\tilde{a}|\tilde{s}')Q^{\pi}(s,\tilde{a})  \\
\frac{1-\kappa_{worst}}{|\mathcal{S}_\epsilon|}, \quad \text{others}
\end{cases}
.
\end{equation}

\subsection{Procedure of the Epsilon Worst Attack (EpsA)}
\label{subsecApp_epsa_procedure}

\begin{algorithm}[tb]
\caption{Epsilon Worst Attack (EpsA) Sampling Method}
\label{alg:epsa-sampling}
\begin{algorithmic}[1] 
\renewcommand{\algorithmicrequire}{\textbf{Input:}}
\renewcommand{\algorithmicensure}{\textbf{Output:}}
\Require{state $s_t$, policy $\pi(a|s)$, action-value function $Q^{\pi}(s,a)$, worst rate parameter $\kappa_{worst}$, 
attack scale $\epsilon$}
\Ensure{Epsilon worst sample $\tilde{s}_t$}
\Function{EpsA}{$s_t, \pi, Q^{\pi}, \kappa_{worst}, \epsilon$}
  \State Sample random value $b$ from uniform distribution [0.0, 1.0)
  \State \textbf{if the case } $\mathbf{b < \kappa_{worst}}$:
    \State \qquad Calculate the worst-case state as the same as the Critic attack:
    \Statex \qquad\qquad
      $
        \tilde{s}^{\star new}_{t}
        \leftarrow 
        proj(\tilde{s}^{\star old}_{t} 
        - \eta \frac{\partial \overline{Q}_{\theta}(s^{}_{t},\mu(\tilde{s}^{\star old}_{t}))}{\partial \tilde{s}^{\star old}_{t}}), 
      $
    \Statex \qquad\qquad
        $
        (\eta \text{ is step size, } \mu \text{ is a mean output of the policy } \pi)
        $
  \State \textbf{else}:
    \State \qquad Sample perturbed state from the uniform distribution in the $L_{\infty}$-normed $\epsilon$ range:
    \Statex \qquad\qquad
      $
      \tilde{s}_t \sim \mathcal{U}(\cdot | s_{t}-\epsilon, s_{t}+\epsilon)
      $
  \State \Return $\tilde{s}_t$
\EndFunction
\end{algorithmic}
\end{algorithm}

We detail Algorithm \ref{alg:epsa-sampling} to clarify the procedure for utilizing the Epsilon Worst Attack (EpsA). 
Initially, we sample a value $b$ from the uniform distribution [0.0, 1.0). 
If $b < \kappa_{worst}$, we apply the Critic attack; conversely, 
if $b \geq \kappa_{worst}$, 
we apply a prior perturbation from the uniform distribution within the $L_{\infty}$-norm range. 
Although this attack is not conceptually new, essentially functioning as an $\epsilon$-greedy strategy for the adversary, it facilitates the estimation of intermediate states between the worst-case scenario and the prior perturbation in a natural form.
This method facilitates understanding scenarios where the worst-case occurs probabilistically and aids in integrating these adversarial elements into robust learning frameworks such as EpsA-SAC.


\subsection{Contraction and Policy Improvement Properties of EpsA-SAC}
\label{subsecApp_epsa_sac_contraction_policy_improvement}

\paragraph{Contraction Property}

To confirm the certainty of our proposal EpsA-SAC algorithm, 
we show the contraction property in this subsection.
We recall the definition:
\footnotesize
\begin{equation}
\label{eq_app_epsa_recall_bellman_def}
\begin{aligned}
\underline{\mathcal{T}}^{\pi}_{epsilon}Q(s_t, a_t) \triangleq 
r(s_t, a_t) & + 
\gamma  \mathbb{E}_{s_{t+1}\sim\mathcal{F}}
\lbrack \\
  &
  \kappa_{worst}
  \underset{V^{\pi}_{worst}(s_{t+1})}{
  \underbrace{
    \mathbb{E}_{\tilde{s}^{\star}_{t+1}\sim \nu^{\star worst}_{\pi_{ept}}}
    \lbrack
      \mathbb{E}_{\tilde{a}^{\star}_{t+1}\sim \pi}
      \lbrack
      Q(s_{t+1}, \tilde{a}^{\star}_{t+1}) - \alpha_{ent} \log\pi(\tilde{a}^{\star}_{t+1}|\tilde{s}^{\star}_{t+1})
      \rbrack
    \rbrack
  }}
  \\
  &
  +
  (1-\kappa_{worst})
  \underset{V^{\pi}_{p}(s_{t+1})}{
  \underbrace{
    \mathbb{E}_{\tilde{s}^{}_{t+1}\sim p}
    \lbrack
      \mathbb{E}_{\tilde{a}^{}_{t+1}\sim \pi}
      \lbrack
      Q(s_{t+1}, \tilde{a}^{}_{t+1}) - \alpha_{ent} \log\pi(\tilde{a}^{}_{t+1}|\tilde{s}^{}_{t+1})
      \rbrack
    \rbrack
  }}
  \\
&\rbrack
\end{aligned}
\end{equation}
\normalsize

\begin{thm}{
The Epsilon Worst Bellman Operator $\underline{\mathcal{T}}^{\pi}_{epsilon}$ is a contraction operator for a fixed policy.
  }
\label{thm:app_epsa_bellman_contraction}
\end{thm}
\textit{Proof.}  

We should remind the fact from the definition of the epsilon worst-case:
\begin{equation}
\label{eq_app_epsa_recall_worst_def}
\begin{aligned}
V^{\pi}_{worst}(s_{t+1})
=
\min_{\tilde{s}^{\star}_{t+1} \in \mathcal{B}_{\epsilon_{p}}(s_{t+1}) }
    \mathbb{E}_{\tilde{a}^{\star}_{t+1}\sim \pi}
    \lbrack
    Q(s_{t+1}, \tilde{a}^{\star}_{t+1}) - \alpha_{ent} \log\pi(\tilde{a}^{\star}_{t+1}|\tilde{s}^{\star}_{t+1})
    \rbrack.
\end{aligned}
\end{equation}
Then, we consider difference between two action-value functions, $Q_1, Q_2$, after using the Epsilon Worst Bellman operator:
\begin{equation}
\label{eq_app_epsa_contraction_rollout1}
\begin{aligned}
  &||
    \underline{\mathcal{T}}^{\pi}_{epsilon}Q_1(s_t,a_t)  -
    \underline{\mathcal{T}}^{\pi}_{epsilon}Q_2(s_t,a_t)
  ||_{s_t,a_t}\\
&=
  ||
    \gamma \mathbb{E}_{s_{t+1}\sim \mathcal{F}}
    \lbrack
        \kappa V^{\pi}_{1,worst} + (1-\kappa) V^{\pi}_{1,p}
        -\kappa V^{\pi}_{2,worst} - (1-\kappa) V^{\pi}_{2,p}
    \rbrack
  ||_{s_t,a_t}\\
&\leq
  \gamma \kappa
  ||
    \mathbb{E}_{s_{t+1}\sim \mathcal{F}}
    \lbrack
        V^{\pi}_{1,worst}
        -V^{\pi}_{2,worst}
    \rbrack
  ||_{s_t,a_t}
  +
  \gamma (1-\kappa)
  ||
    \mathbb{E}_{s_{t+1}\sim \mathcal{F}}
    \lbrack
        V^{\pi}_{1,p}
        -V^{\pi}_{2,p}
    \rbrack
  ||_{s_t,a_t} \\
&\underset{(1)}{\leq}
  \gamma \kappa
  ||
    \mathbb{E}_{s_{t+1}\sim \mathcal{F}}
    \lbrack
        V^{\pi}_{1,worst}
        -V^{\pi}_{2,worst}
    \rbrack
  ||_{s_t,a_t}
  +
  \gamma (1-\kappa)
  ||
        Q_{1} - Q_{2}
  ||_{s_{t+1},a_{t+1}} \\
&\underset{(2)}{\leq}
  \gamma \kappa
  ||
        Q_{1} - Q_{2}
  ||_{s_{t+1},a_{t+1}}
  +
  \gamma (1-\kappa)
  ||
        Q_{1} - Q_{2}
  ||_{s_{t+1},a_{t+1}} \\
&=
  \gamma
  ||
        Q_{1}(s_{t+1}, a_{t+1}) - Q_{2}(s_{t+1}, a_{t+1})
  ||_{s_{t+1},a_{t+1}}.
\end{aligned}
\end{equation}
For the inequality (1), 
we cancel entropy terms (under the same policy and perturbation) and use:
\begin{equation}
\label{eq_app_epsa_minor_inequality}
\mathbb{E}_{\tilde{s}_{t+1}\sim p}
\left[
\mathbb{E}_{\tilde{a}_{t+1}\sim \pi}
\left[
Q_1(s_{t+1},\tilde{a}_{t+1}) - Q_2(s_{t+1},\tilde{a}_{t+1}) 
\right]
\right]
\leq ||Q_1(s_{t+1},\tilde{a}_{t+1}) - Q_2(s_{t+1},\tilde{a}_{t+1}) ||_{a_{t+1}}.
\end{equation}
For the inequality (2), similarly to the WocaR case \citep{liang2022efficient}, we utilized:
\begin{equation}
\label{eq_app_epsa_minmin_max}
|\min_{\tilde{s}\in\mathcal{B}} A - \min_{\tilde{s}\in\mathcal{B}} B|
\leq
\max_{\tilde{s}\in\mathcal{B}} |A-B|
\end{equation}
and canceling entropy terms due to the same fixed policy.

Therefore, we can say $\underline{\mathcal{T}}_{epsilon}^{\pi}$ is a contraction operator,
because we perform this operation an infinite number of times, 
$Q_{1}$ and $Q_{2}$ converge to a fixed point. \qed

\paragraph{Policy Improvement Property (with a fixed adversary)}

We can also say EpsA-SAC can improve the policy under a fixed adversary.
Derivation is perfectly same as for SofA-SAC in Appendix \ref{subsecApp_softa_sac_contraction_policy_improvement}, 
then we omit the description here.
We should note that 
in Eq. \eqref{eq_epsilon_worst_sac_policy_loss1}, 
we approximate $\tilde{s}_t\sim\nu^{\star worst}_{\pi_{ent}}(\cdot|s_t)$ 
by the numerical method, Projected Gradient Descent (PGD), 
then naturally we fix the adversary during updating the policy.

\subsection{Practical Implementation of EpsA-SAC}
\label{subsecApp_epsa_practical_implementation}

Algorithm \ref{alg:epsa-sac-implementation} shows a pseudocode that represents 
practical implementation of the EpsA-SAC algorithm in Section \ref{subsubsec_epsa_rl}.
As for the same reason as in SofA-SA, 
we use two critic networks with target networks and one policy network without a target network.

Compared to SofA-SAC, 
without any special measurements, 
EpsA-SAC can train agents stably among all the four tasks 
by gradually increasing the worst-case weight $\kappa_{worst}$ from 0.0 to the final value 
(most case we found \{0.8, 1.0\} is preferable).

\begin{algorithm}[tb]
\caption{EpsA-SAC Training}
\label{alg:epsa-sac-implementation}
\begin{algorithmic}[1] 

\State Initialize critic $Q_{\theta_{1,2}}(s,a)$ and actor $\pi_{\phi}(s)$
\State Initialize target networks $Q_{\theta_{1,2}'}(s,a)$ by setting $\theta_{1,2}' \leftarrow \theta_{1,2}$

\State Initialize a replay buffer $\mathcal{R} \leftarrow \emptyset$ and entropy coefficient $\alpha_{ent} \leftarrow 1.0$
\For{$t = 1$ to $T$}
  \State Execute action $a_t \sim \pi_{\phi}(s_t)$, observe $(s_t, a_t, r_t, s_{t+1}, \textit{done}_t)$, and store in $\mathcal{R}$
  \State Sample a mini-batch of $M$ transitions $(s_t^i, a_t^i, r_t^i, s_{t+1}^i, \textit{done}_t^i) \sim \mathcal{R}$
  \Statex \quad \textbf{- Update Critic:}

  \Statex \quad\quad \textbf{{- For random-case adversaries:}}
    \State \quad 
    Sample perturbed states from the uniform distribution:
    \Statex \qquad \qquad 
      $\tilde{s}^{i}_{t+1}\sim \mathcal{U}(\cdot|s^{i}_{t+1}-\epsilon, s^{i}_{t+1}+\epsilon)$
    \State \quad 
    Estimate random-case target values for each sample:
    \Statex \qquad \qquad 
      $\tilde{a}^{i}_{t+1} \sim \pi_{\phi}(\cdot|\tilde{s}^{i}_{t+1})$, \quad
      $
      v^{ir}_{1,2}=
      Q_{\theta'_{1,2}}(s_{t+1}^{i},\tilde{a}^{i}_{t+1}) 
      - \alpha_{ent} \log \pi_{\phi}(\tilde{a}^{i}_{t+1}|\tilde{s}^{i}_{t+1}) 
      $
    \Statex \qquad \qquad 
    $y^{ir}= r^{i}_t + (1-\textit{done}^i_t) \gamma \min_{\{1,2\}} v^{ir}_{1,2}$

  \Statex \quad\quad \textbf{{- For worst-case adversaries:}}
  \State \quad 
  Compute worst-case states $\tilde{s}^{i\star}_{t+1}$ by using PGD in the $L_{\infty}$-norm state space (5 steps):
  \Statex \qquad \qquad 
  $\tilde{s}^{i\star new}_{t+1} 
  \leftarrow 
  proj(\tilde{s}^{i\star old}_{t+1} - \eta \frac{\partial \overline{Q}_{\theta_{1,2}'}(s^{i}_{t+1},\mu(\tilde{s}^{i\star old}_{t+1}))}{\partial \tilde{s}^{i\star old}_{t+1}})$ 
  , $\eta$: step size, $\mu$: mean action
  \State \quad 
  Estimate worst-case target values for each sample:
  \Statex \qquad \qquad 
  $\tilde{a}^{i\star}_{t+1} \sim \pi_{\phi}(\cdot|\tilde{s}^{i\star}_{t+1})$, \quad
  $
    v^{i\star}_{1,2}=
    Q_{\theta'_{1,2}}(s_{t+1}^{i},\tilde{a}^{i\star}_{t+1}) 
    - \alpha_{ent} \log \pi_{\phi}(\tilde{a}^{i\star}_{t+1}|\tilde{s}^{i\star}_{t+1}) 
  $
  \Statex \qquad \qquad 
  $y^{i\star}= r^{i}_t + (1-\textit{done}^i_t) \gamma \min_{\{1,2\}} v^{i\star}_{1,2}$

  \State
  Update $\theta_{1,2}$ by minimizing Huber-Loss:
  \Statex \qquad
  $L(\theta_{1,2}) =
  \frac{1}{M} \sum_{i=1}^{M}
  \left(
  \kappa_{worst} L_{\text{Hu}}(Q_{\theta_{1,2}}(s^{i}_t,a^{i}_t), y^{i\star}) + (1-\kappa_{worst}) L_{\text{Hu}}(Q_{\theta_{1,2}}(s^{i}_t,a^{i}_t), y^{ir})
  \right)
  $

  \Statex \quad \textbf{- Update Actor:}
  \Statex \quad\quad \textbf{{- For random-case adversaries:}}
  \State \quad
  Sample perturbed states from the uniform distribution:
  \Statex \qquad \qquad 
    $\tilde{s}^{i}_{t}\sim \mathcal{U}(\cdot|s^{i}_{t}-\epsilon, s^{i}_{t}+\epsilon)$
  \State \quad
  Calculate loss for the policy:
  \Statex \qquad \qquad 
  $\tilde{a}^{ir}_{t} \sim \pi_{\phi}(\cdot|\tilde{s}^{ir}_t)$,
  \Statex \qquad \qquad 
  $L^{ir} = \alpha_{ent} \log\pi_{\phi}(\tilde{a}^{ir}_{t}|\tilde{s}^{ir}_t ) - \min_{\theta_1,\theta_2} Q_{\theta_{1,2}}^{}(s^{i}_t, \tilde{a}^{ir}_t)$

  \Statex \quad\quad \textbf{{- For worst-case adversaries:}}
  \State \quad 
  Compute worst-case states $\tilde{s}^{i\star}_{t}$ by using PGD in the $L_{\infty}$-norm state space (5 steps):
  \Statex \qquad \qquad 
  $\tilde{s}^{i\star new}_{t} 
  \leftarrow 
  proj(\tilde{s}^{i\star old}_{t} - \eta \frac{\partial \overline{Q}_{\theta_{1,2}}(s^{i}_{t},\mu(\tilde{s}^{i\star old}_{t}))}{\partial \tilde{s}^{i\star old}_{t}})$ 
  , $\eta$: step size, $\mu$: action mean
  \State \quad 
  Calculate loss for the policy:
  \Statex \qquad \qquad 
    $\tilde{a}^{i\star }_{t} \sim \pi_{\phi}(\cdot|\tilde{s}^{i\star}_t)$,
  \Statex \qquad \qquad 
  $L^{i\star} = \alpha_{ent} \log\pi_{\phi}(\tilde{a}^{i\star}_{t}|\tilde{s}^{i\star}_t ) - \min_{\theta_1,\theta_2} Q_{\theta_{1,2}}^{}(s^{i}_t, \tilde{a}^{i\star}_t)$

  \State 
  Update $\phi$ by minimizing mixed policy loss:
  \Statex \qquad \qquad 
  $
  L(\phi_{}) = \frac{1}{M} \sum_{i=1}^{M} \left( \kappa_{worst} L^{i\star} + (1-\kappa_{worst}) L^{ir} \right)
  $

  \Statex \quad \textbf{- Update Entropy Coefficient:}
  \State Re-use the entropy values during policy improvement and update $\alpha_{ent}$:
  \Statex \qquad
    $
    \mathcal{H}_{current}^{i} = 
    -\kappa_{worst} \log\pi_{\phi}(\tilde{a}^{i\star}_{t}|\tilde{s}^{i\star}_t )
    -(1 - \kappa_{worst}) \log\pi_{\phi}(\tilde{a}^{ir}_{t}|\tilde{s}^{ir}_t )
    $
  \Statex \qquad
    $
    L(
      \alpha_{ent}) = -\frac{1}{M} \sum_{i=1}^{M} \alpha_{ent} (
      \mathcal{H}_{target} 
      -\mathcal{H}^{i}_{current} 
      )
    $

  \Statex \quad \textbf{- Post Processing:}
  \State Soft update the target networks: $\theta_{1,2}' \leftarrow (1-\tau) \theta_{1,2}' + \tau \theta_{1,2}$
  \State Update the worst-case weight parameter $\kappa_{worst}$ according to the schedule

\EndFor
\end{algorithmic}
\end{algorithm}

\section{Additional Results and Experiments for SofA-SAC}
\label{secApp_SofA_additional_results}

In this section, we present the results of additional experiments to detail the characteristics of SofA-SAC and validate the observed tendencies.

\subsection{Learning Curves and Evaluation Score Tables for SofA-SAC and Baseline Algorithms}
\label{subsecApp_SofA_additional_results_main_learning_curve}

\begin{figure}
\centering
\includegraphics[keepaspectratio, width=1\linewidth]{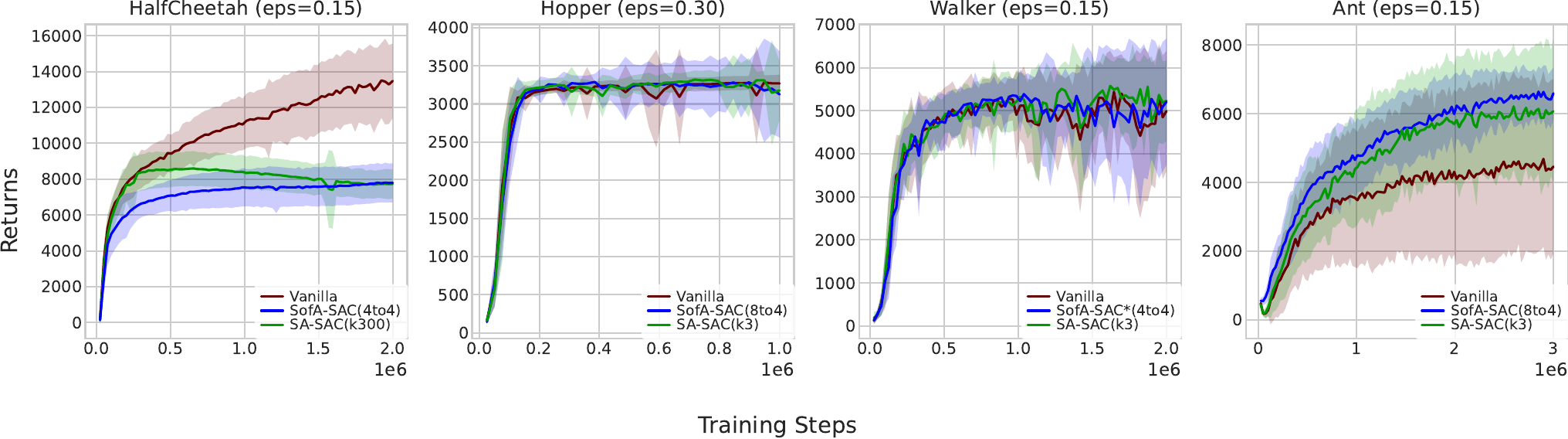}
\vspace{-5truemm}
\caption{
Learning curves for SofA-SAC and baseline algorithms on four MuJoCo control tasks. 
The solid lines represent the average evaluation scores, and the shaded areas indicate the standard deviation.
}
\label{fig_appendix_softa_main_learning_curve}
\end{figure}

\begin{table}
\caption{
  Median values of average episode rewards across 16 different seed models of original SAC, SA-SAC, and SofA-SAC for four MuJoCo control tasks.
  Bold scores indicate the highest evaluation scores among the three methods.
  }
\label{tab:softa_main_result_table}
\centering
\resizebox{\columnwidth}{!}{%
\begin{tabular}{c|c|ccccccc}
\toprule
\multirow{2}{*}{
\textbf{Environment}
}& 
\multirow{2}{*}{
\textbf{Method}
}
& 
\multirow{2}{*}{
\textbf{Vanilla}
}& \textbf{Gaussian} & \textbf{SofA} & \textbf{SofA} & \textbf{SofA} & \textbf{SofA} & \textbf{SofA} \\
 & & & ($\alpha_{\text{attk}} \rightarrow \infty$) & $\alpha_{\text{attk}}=2048$ & $\alpha_{\text{attk}}=24$ & $\alpha_{\text{attk}}=4$ & $\alpha_{\text{attk}}=1$ & $\alpha_{\text{attk}} \rightarrow 0$ \\
\midrule
\multirow{3}{*}{
  \begin{tabular}{c}
  \textbf{HalfCheetah}\\
  $\epsilon=0.15$
  \end{tabular}
  }
&Vanilla 
& \textbf{13110.6} & 3075.6 & 2934.9 & 1746.7 & 422.7 & 37.9 & -8.4 \\
&SA-SAC 
& 7769.3 & 6810.6 & 6832.1 & 6369.4 & 3320.8 & 2096.4 & 1804.1 \\
&\text{SofA-SAC (Ours)}
& 7720.5 & \textbf{7268.5} & \textbf{7278.2} & \textbf{7212.1} & \textbf{6025.6} & \textbf{3443.8} & \textbf{2659.9} \\

\midrule
\multirow{3}{*}{
  \begin{tabular}{c}
  \textbf{Hopper}\\
  $\epsilon=0.30$
  \end{tabular}
  }
 &Vanilla 
& 3278.2 & 1408.0 & 1378.7 & 868.6 & 414.2 & 224.6 & 20.3 \\
 &SA-SAC 
& \textbf{3334.3} & 3137.9 & 3102.3 & 3039.0 & 2625.8 & \textbf{1937.8} & \textbf{1415.9} \\
 &\text{SofA-SAC (Ours)}
& 3293.8 & \textbf{3221.3} & \textbf{3221.4} & \textbf{3225.3} & \textbf{3159.6} & 1551.2 & 567.8 \\

\midrule
\multirow{3}{*}{
  \begin{tabular}{c}
  \textbf{Walker2d}\\
  $\epsilon=0.15$
  \end{tabular}
  }

 &Vanilla 
& 5587.6 & 4172.7 & 4336.6 & 3801.7 & 1848.3 & 438.8 & 384.2 \\
 &SA-SAC 
& 5416.0 & \textbf{5357.6} & 5205.3 & 5264.2 & 4752.7 & \textbf{2965.6} & \textbf{1149.9} \\
 &\text{SofA-SAC* (Ours)}
& \textbf{5744.5} & 5324.6 & \textbf{5347.4} & \textbf{5452.4} & \textbf{5004.6} & 2457.6 & 811.1 \\
\midrule
\multirow{3}{*}{
  \begin{tabular}{c}
  \textbf{Ant}\\
  $\epsilon=0.15$
  \end{tabular}
  }
 &Vanilla 
& 5379.1 & 645.4 & 767.8 & 472.3 & 162.3 & 83.8 & 86.7 \\
 &SA-SAC 
& \textbf{6946.8} & 4835.9 & 4800.4 & 4487.1 & 2637.2 & 1229.3 & 934.6 \\
 &\text{SofA-SAC (Ours)}
& 6816.9 & \textbf{5127.3} & \textbf{5273.2} & \textbf{4990.1} & \textbf{3940.2} & \textbf{2076.4} & \textbf{1339.4} \\

\bottomrule
\end{tabular}
}
\end{table}

Fig. \ref{fig_appendix_softa_main_learning_curve} presents 
the learning curves for SofA-SAC and baseline algorithms in Section \ref{subsubsecEvalSofA}.
We would like to remind you that during training, 
SofA-SAC utilized Gaussian-based observation perturbations (virtually), 
and SA-SAC maintained the consistency of the policy output under the $L_{\infty}$-norm. 
However, the actual trainings were executed without any perturbation. 
Although we observe modest scores due to enhanced robustness in HalfCheetah, 
we found that the introduction of SofA-SAC and SA-SAC did not sacrifice performance. 
Furthermore, learnings were more stable than with Vanilla-SAC in the complex task (Ant).

Table \ref{tab:softa_main_result_table} shows the median values of average evaluation scores obtained from sixteen different training seeds.
Notably, the scores of Vanilla-SAC decrease significantly as the temperature parameter $\alpha_{attk}$ is reduced. In contrast, our proposed method maintains performance and outperforms other baseline models, especially when $\alpha_{attk} \geq 4$.



Following the recommendation in \cite{zhang2021robust}, 
which suggests choosing the median seeds from the evaluation where the methods get the weakest scores, 
we attempt to extract median seeds. 
However, we observe that ranking scores under the strongest attack conditions occasionally results in selecting training seeds that perform significantly worse in attack-free evaluations. 
This issue is particularly pronounced with Vanilla-SAC, which sometimes fails to achieve valid scores during strong attack evaluations.
Therefore, we represent all the averaged evaluation scores in the box plots for each method and evaluation, 
this can represent the variance and tendencies each methods have.

\subsection{Ablation Studies on SofA-SAC Procedures Across All Four Tasks}
\label{subsecApp_SofA_additional_results_ablation_procedures}

\begin{figure}
\centering
\includegraphics[keepaspectratio, width=1\linewidth]{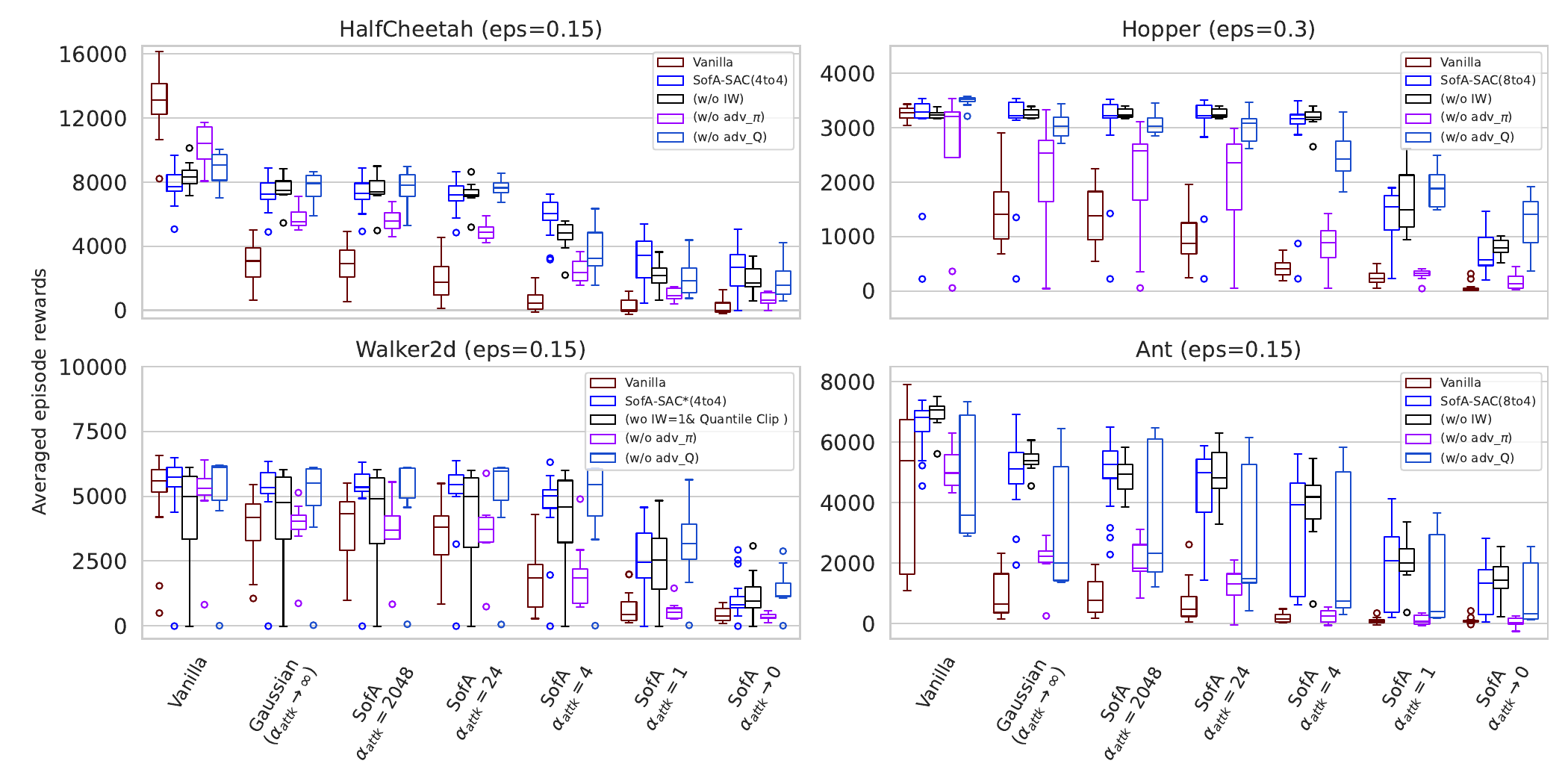}
\vspace{-5truemm}
\caption{
Ablation studies of SofA-SAC were conducted on all four tasks under Gaussian-based soft worst-case attacks. 
We tested versions including: 
neglecting the importance weight during policy improvement (w\slash o IW), 
fully omitting adversaries during policy improvement (w\slash o adv\_$\pi$), 
and omitting adversaries during the updating of Q-values (w\slash o adv\_Q).
}
\label{fig_appendix_softa_ablation_boxplot_softa_procedure}
\end{figure}

Fig. \ref{fig_appendix_softa_ablation_boxplot_softa_procedure} shows the robustness evaluation results for SofA-SAC and its ablation variants. 
We tested three variant methods in addition to Vanilla-SAC and SofA-SAC. 
The first variant does not consider the importance weight in Eq. \ref{eq_soft_worst_sac_policy_loss1} during the policy improvement. 
The second does not involve adversaries during policy improvement, as described in Eq. \ref{eq_soft_worst_sac_policy_loss1}. 
The last variant does not use adversaries during Q updates, as outlined in Eq. \ref{eq_soft_worst_kl_bellman}.

\paragraph{Effect of Importance Weight}
In HalfCheetah, the robustness of the variant that omits the importance weight is clearly inferior to that of SofA-SAC. 
However, in other tasks, it is competitive. 
We assume that the importance weight facilitates precise estimations for pessimistic scenarios under the adversary 
but also introduces a variance element to training.

\paragraph{Effect of Adversary during Policy Improvement}
Across all four tasks, the absence of adversaries during policy improvement results in a significant deterioration of robustness. 
These tendencies are particularly evident in the complex task of Ant and under strong (or large) attacks.

\paragraph{Effect of Adversary during Q-update}
Omitting adversaries during Q-update leads to high variance and slightly inferior scores under attack evaluations. 
We estimate that without adversaries during Q-updates, 
the action values tend to be overly optimistic (under the soft worst-case scenario), 
which causes the policy behavior to become rough and fail to maintain good scores under perturbations.


\subsection{Ablation Studies on SofA-SAC's Sampling Settings Across All Four Tasks}
\label{subsecApp_SofA_additional_results_ablation_hyperparams}

\begin{figure}
\centering
\includegraphics[keepaspectratio, width=1\linewidth]{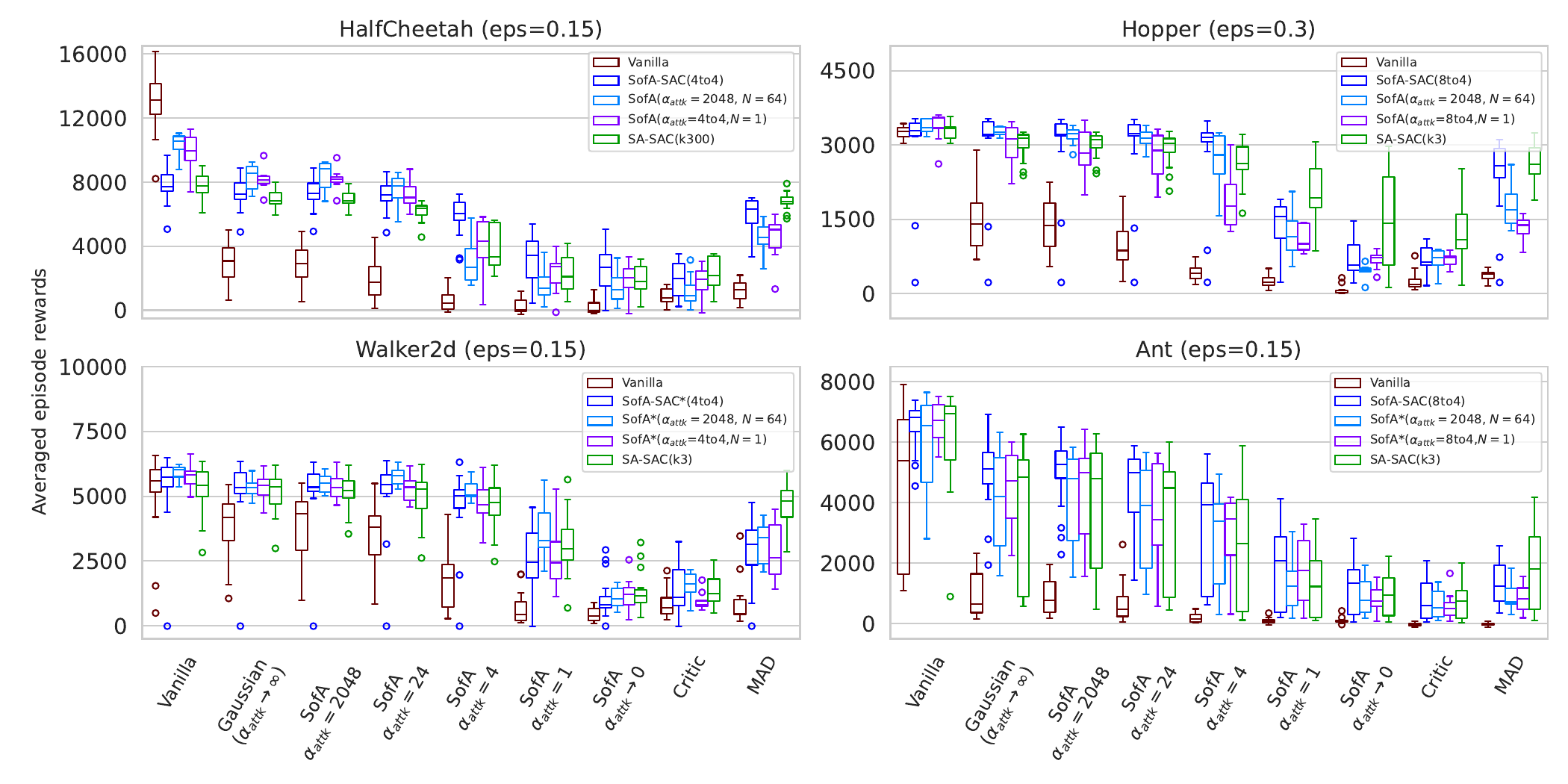}
\vspace{-5truemm}
\caption{
Ablation studies of SofA-SAC were conducted on all four tasks under Gaussian-based soft worst-case attacks.
We tested versions including: one where the coefficient parameter is high (nearing randomness) with $\alpha_{\text{attk}}=2048$,
and another using only one sample to update both the policy and Q-value ($N=1$).
}
\label{fig_appendix_softa_ablation_boxplot_softa_param}
\end{figure}

To validate the differences between adding random perturbations during optimization, 
we examine cases where the hyperparameter is adjusted to extreme values: 
one variant involves increasing the temperature parameter, $\alpha_{attk}$, (which approximates Gaussian perturbation), 
and the other uses only one sample (identical to a single sample from Gaussian perturbation). 
Fig. \ref{fig_appendix_softa_ablation_boxplot_softa_param} displays the robustness evaluation results. 
SofA-SAC with $\alpha_{attk}=4$ maintains higher scores even as attacks intensify. 
However, especially in the case of HalfCheetah, 
a clear trade-off between performance without attacks and robustness under attacked conditions is observed. 
The two variants outperform SofA-SAC ($\alpha_{\text{attk}}=4$) under attack-free conditions but exhibit lower robustness under evaluation conditions with attacks.

\subsection{Sensitivity Analysis of Coefficient Parameters for SofA-SAC and SA-SAC}
\label{subsecApp_SoftA_additional_results_softa_coefficient}

\begin{figure}
\centering
\includegraphics[keepaspectratio, width=1\linewidth]{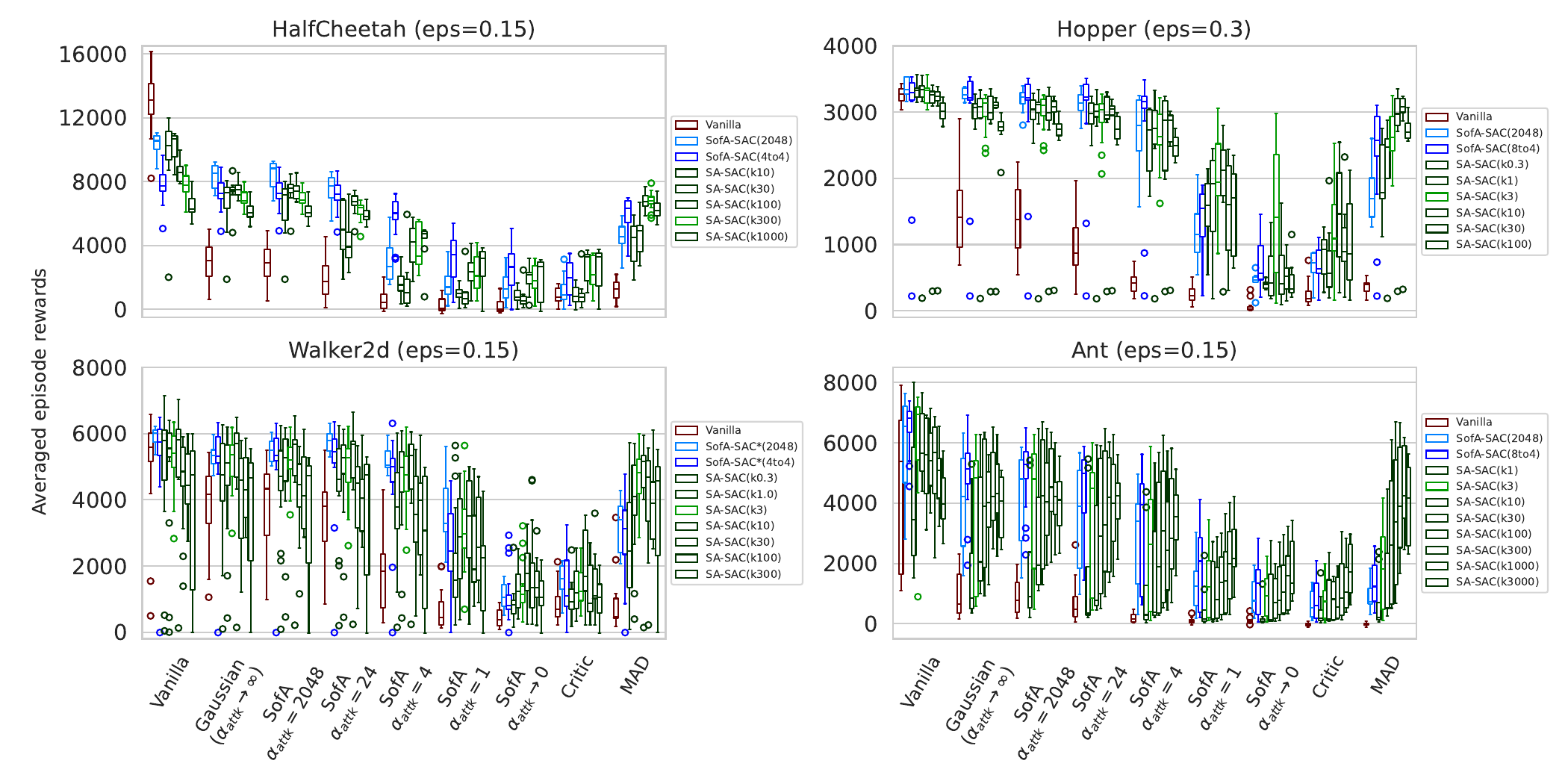}
\vspace{-5truemm}
\caption{
Box plots represent the performance of SofA-SAC and SA-SAC under Gaussian-based attacks across trainings with various coefficient parameters, 
including $\alpha_{attk}$ for SofA-SAC and $\kappa_{reg}$ for SA-SAC. 
We use the same colors for the main settings: Vanilla (brown), SofA-SAC (blue), and SA-SAC (green). 
Variants of SofA-SAC and SA-SAC with different parameters are represented by light blue (for $\alpha_{attk}=2048$) and dark green, respectively.
}
\label{fig_appendix_softa_coefficient_boxplot}
\end{figure}

To elucidate the characteristics of our methods and the tasks evaluated in our experiments, 
we present Fig. \ref{fig_appendix_softa_coefficient_boxplot}, 
which shows the average evaluation scores for SofA-SAC and SA-SAC across various coefficient trainings.

In HalfCheetah, 
the trade-off between performance under evaluation without attacks and robustness against attacks is evident, 
influenced by the coefficient parameters ($\alpha_{attk}$ for SofA-SAC and $\kappa_{reg}$ for SA-SAC). 
Despite this trade-off, when comparing SofA-SAC with SA-SAC at equivalent levels of (Vanilla) evaluation, 
SofA-SAC appears superior for strong attacks that related to MDPs (SofA and Critic attack). 
In the context of the MAD attack, 
SA-SAC consistently achieves higher or competitive scores with strong coefficient parameters 
due to the integration of a consistency term into the policy's objective function to counteract MAD attacks.

In Hopper and Walker, 
variants trained with weaker coefficient parameters than the main results (blue and green) maintain equivalent performance under evaluation without attacks. 
However, stronger attacks reveal the lack of robustness in these variants with weak coefficient parameters. 
Therefore, we can search for parameters that do not degrade performance without attacks but maintain a certain degree of robustness against attacks.

In Ant, weaker coefficient parameters yield slightly inferior evaluation scores, 
and overly strong attacks significantly reduce natural scores. 
We hypothesize that due to Ant's high-dimensional state space, 
a moderate level of pessimism or regularization aids in stabilizing learning. 
However, when regularization is excessively strong, 
it leads to a drop in performance without attacks due to the trade-off between the original task objective and the regularization objectives.

\section{Additional Results and Experiments for EpsA-SAC}
\label{secApp_Epsilon_additional_results}

In this section, we present the results of additional experiments to detail the characteristics of EpsA-SAC and validate the observed tendencies.

\subsection{Learning Curves and Evaluation Score Tables for EpsA-SAC and Baseline Algorithms}
\label{subsecApp_EpsA_additional_results_main_learning_curve}

\begin{figure}
\centering
\includegraphics[keepaspectratio, width=1\linewidth]{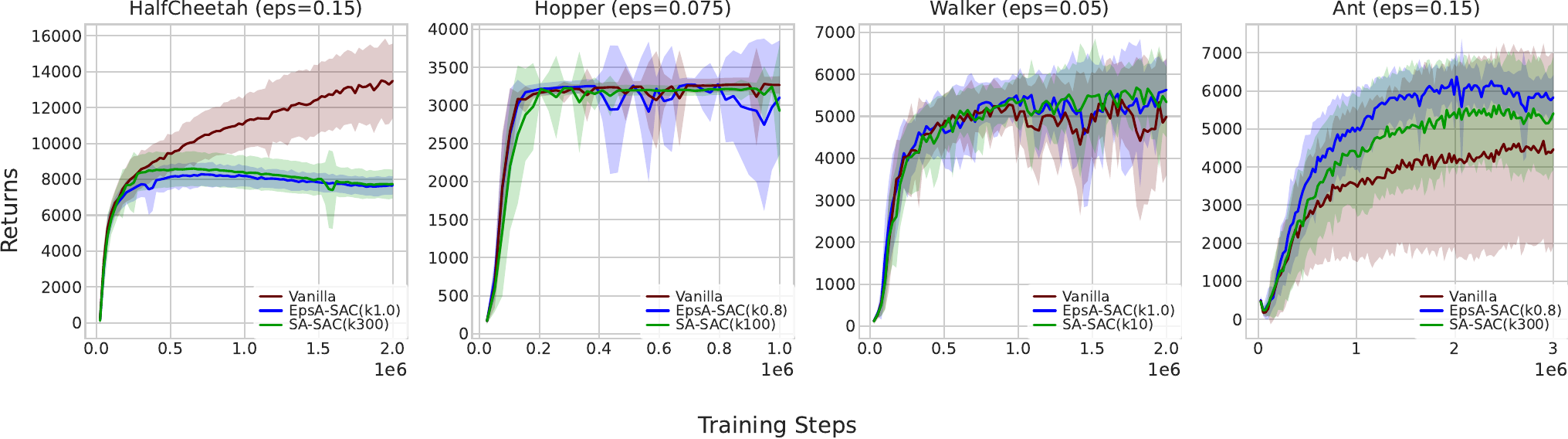}
\vspace{-5truemm}
\caption{
Learning curves for EpsA-SAC and baseline algorithms on four MuJoCo control tasks. 
The solid lines represent the average evaluation scores, and the shaded areas indicate the standard deviation.
}
\label{fig_appendix_epsa_main_learning_curve}
\end{figure}

\begin{table}
\caption{
  Median values of average episode rewards across 16 different seed models of original SAC, SA-SAC, and EpsA-SAC for four MuJoCo control tasks.
  Bold scores indicate the highest evaluation scores among the three methods.
  }
\label{tab:epsa_main_result_table}
\centering
\resizebox{\columnwidth}{!}{%
\begin{tabular}{c|c|ccccccccc}
\toprule
\multirow{2}{*}{\textbf{Environment}} & 
\multirow{2}{*}{\textbf{Method}} & 
\multirow{2}{*}{\begin{tabular}{c}\textbf{Natural} \\ \textbf{Reward}\end{tabular}} & 
\multirow{1}{*}{\textbf{Uniform}} & 
\textbf{EpsA} &
\textbf{EpsA} &
\textbf{Critic} &
\multirow{2}{*}{\textbf{MAD}} & 
\multirow{2}{*}{\textbf{SA-RL}} & 
\multirow{2}{*}{\textbf{PA-AD}} 
\\
&  &  
& ($\kappa_{worst}=0.0$)
& $\kappa_{worst}=0.4$ 
& $\kappa_{worst}=0.8$ 
& ($\kappa_{worst}=1.0$) 
&
& 
& 
& 
\\
\midrule
\multirow{3}{*}{
  \begin{tabular}{c}
  \textbf{HalfCheetah}\\
  $\epsilon=0.15$
  \end{tabular}
  } 
& Vanilla 
& \textbf{13110.6} & 4547.8 & 2435.2 & 1160.3 & 769.5 & 1283.4 & -1139.7 & -32.0 \\
& SA-SAC 
& 7769.3 & 7404.3 & 5240.2 & 3054.9 & 2158.0 & 6794.7 & 1064.9 & 1784.7 \\
 &EpsA-SAC (Ours)
& 7681.9 & \textbf{7572.8} & \textbf{6701.5} & \textbf{5495.0} & \textbf{4279.4} & \textbf{7000.7} & \textbf{5501.4} & \textbf{5327.9} \\

\midrule
\multirow{3}{*}{
  \begin{tabular}{c}
  \textbf{Hopper}\\
  $\epsilon=0.075$
  \end{tabular}
  } 
& Vanilla 
& 3278.2 & 3290.5 & 3239.9 & 3237.9 & 3236.8 & 3225.1 & 1840.1 & 2956.6 \\

&SA-SAC
& 3191.4 & 3195.7 & 3160.0 & 3145.9 & 3096.5 & 3204.9 & 3130.2 & 3141.0 \\

 &EpsA-SAC (Ours)
& \textbf{3310.2} & \textbf{3319.4} & \textbf{3280.0} & \textbf{3249.7} & \textbf{3187.5} & \textbf{3340.4} & \textbf{3149.3} & \textbf{3243.2} \\

\midrule
\multirow{3}{*}{
  \begin{tabular}{c}
  \textbf{Walker2d}\\
  $\epsilon=0.05$
  \end{tabular}
  } 
& Vanilla 
& 5587.6 & 5347.5 & 4976.6 & 4844.4 & 4388.5 & 4648.8 & 2858.9 & 2253.1 \\
&SA-SAC
& \textbf{6056.3} & \textbf{6054.1} & 5654.4 & 5343.9 & \textbf{5169.0} & \textbf{6025.4} & \textbf{5741.4} & \textbf{5474.3} \\
 &EpsA-SAC (Ours)
& 5928.8 & 5904.7 & \textbf{5769.1} & \textbf{5692.1} & 5134.8 & 5717.6 & 5224.6 & 5282.8 \\

\midrule
\multirow{3}{*}{
  \begin{tabular}{c}
  \textbf{Ant}\\
  $\epsilon=0.15$
  \end{tabular}
  } 
& Vanilla 
& 5379.1 & 1907.7 & 177.0 & 15.8 & -15.1 & -19.0 & 168.6 & -542.0 \\
&SA-SAC
& 5691.6 & 5258.3 & 1593.8 & 710.0 & 839.8 & \textbf{3901.3} & \textbf{3026.8} & 1173.1 \\
 &EpsA-SAC (Ours)
& \textbf{5957.3} & \textbf{5610.4} & \textbf{3666.8} & \textbf{2166.4} & \textbf{2052.1} & 3269.2 & 2821.2 & \textbf{4033.7} \\

\bottomrule
\end{tabular}
}
\end{table}

Fig. \ref{fig_appendix_epsa_main_learning_curve} presents the learning curves for EpsA-SAC and baseline algorithms in Section \ref{subsubsecEvalEpsA}.
We would like to remind you that during training, 
EpsA-SAC utilized $L_{\infty}$-norm observation perturbations, 
the uniform distribution and the Critic attack, 
and SA-SAC maintained the consistency of the policy output under the $L_{\infty}$-norm. 
However, the actual trainings were executed without any perturbation. 
As the same as in Appendix \ref{subsecApp_SofA_additional_results_main_learning_curve}, 
we observe modest scores due to enhanced robustness in HalfCheetah, however, 
we found that the introduction of EpsA-SAC and SA-SAC did not sacrifice performance. 
Furthermore, learnings were more stable than with Vanilla-SAC in the complex task (Ant).

\subsection{Ablation Studies on EpsA-SAC Procedures Across All Four Tasks}
\label{subsecApp_EpsA_additional_results_ablation_procedures}

\begin{figure}
\centering
\includegraphics[keepaspectratio, width=1\linewidth]{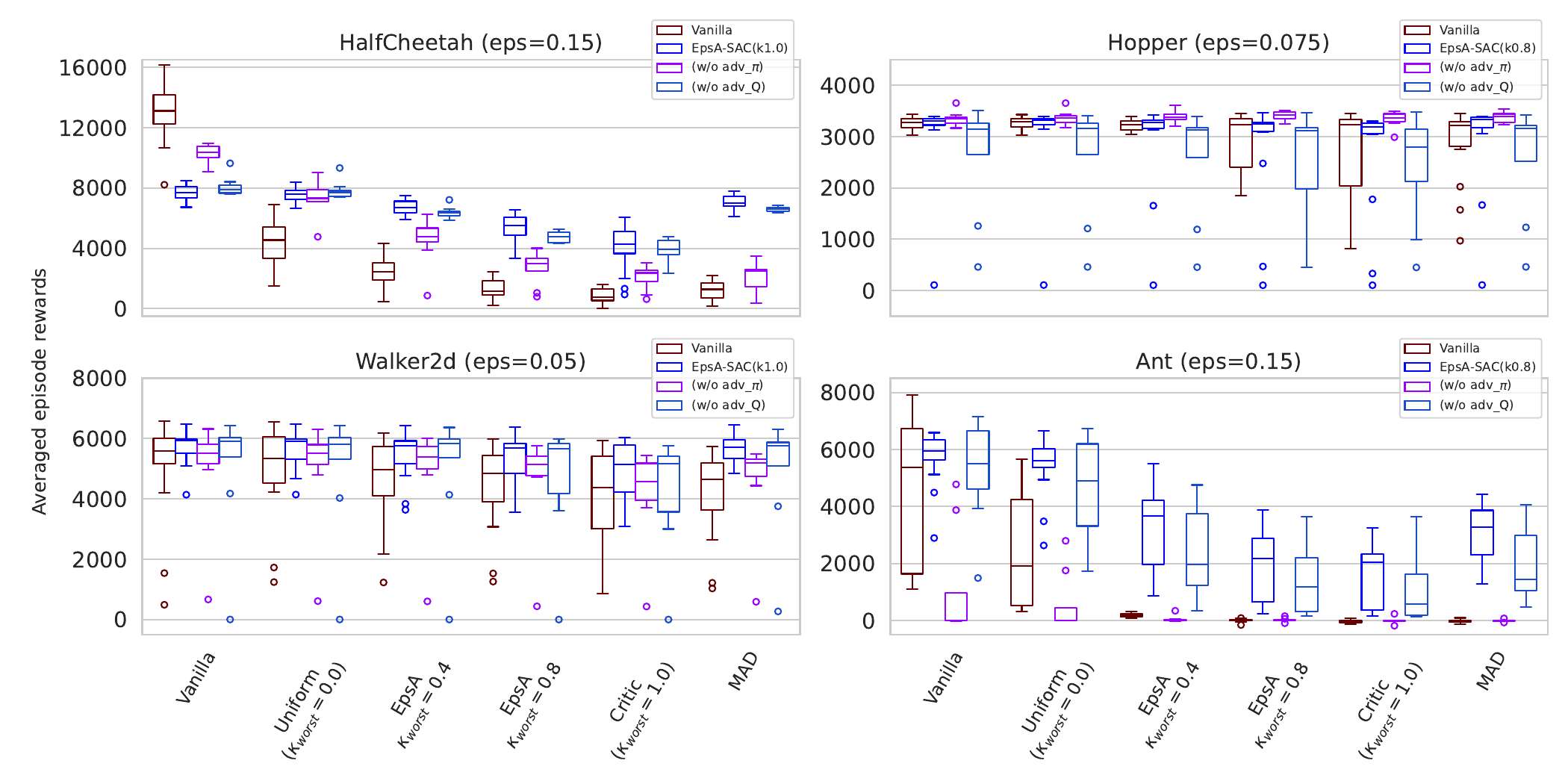}
\vspace{-5truemm}
\caption{
Ablation studies of EpsA-SAC under the $L_{\infty}$-norm based attack.
We tested versions including: 
omitting adversaries during policy improvement (w\slash o adv\_$\pi$) and 
omitting adversaries during the updating of Q-values (w\slash o adv\_Q).
}
\label{fig_appendix_epsa_ablation_boxplot_softa_procedure}
\end{figure}

\begin{figure}
\centering
\includegraphics[keepaspectratio, width=1\linewidth]{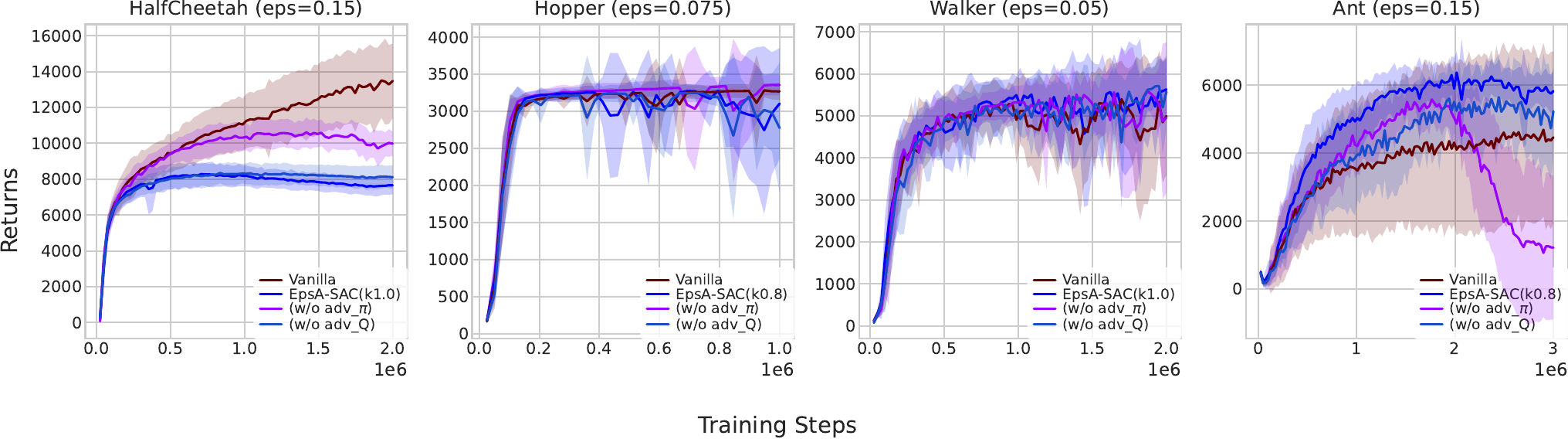}
\vspace{-5truemm}
\caption{
Learning curves of ablation studies for EpsA-SAC.
We trained versions including: 
omitting adversaries during policy improvement (w\slash o adv\_$\pi$) and 
omitting adversaries during the updating of Q-values (w\slash o adv\_Q).
}
\label{fig_appendix_epsa_ablation_procedure_learning_curve}
\end{figure}

For EpsA-SAC, we also tested the absence of adversaries during training procedures. 
Fig. \ref{fig_appendix_epsa_ablation_boxplot_softa_procedure} and \ref{fig_appendix_epsa_ablation_procedure_learning_curve} 
illustrate the learning curves and robust evaluation results for ablation studies that omit the adversary during policy improvement and Q-update. 
Although we observe tendencies similar to those in the SofA-SAC ablation study (detailed in Appendix \ref{subsecApp_SofA_additional_results_ablation_procedures}), 
the absence of an adversary during policy improvement typically results in reduced robustness, 
and its absence during Q-update leads to higher variance scores. 
In Hopper and Walker2d, the differences are minimal, 
but the collapse in Ant is severe.
We assume that this is related to the observation dimension and perturbation scales that is utilized during training.
As described in Appendix \ref{subsecApp_epsa_sac_contraction_policy_improvement}, 
EpsA-SAC theoretically integrates the training framework with the assumed adversary to maintain and update a unified action-value function. 
However, if the attack scale increases and the observation dimensions are large, 
the policy can no longer produce valid actions for subsequent action values. 
Consequently, the target action value decreases progressively, 
leading to training collapse.

\subsection{Sensitivity Analysis of Coefficient Parameters for EpsA-SAC and SA-SAC}

\label{subsecApp_EpsA_additional_results_epsa_coefficient}
\begin{figure}
\centering
\includegraphics[keepaspectratio, width=1\linewidth]{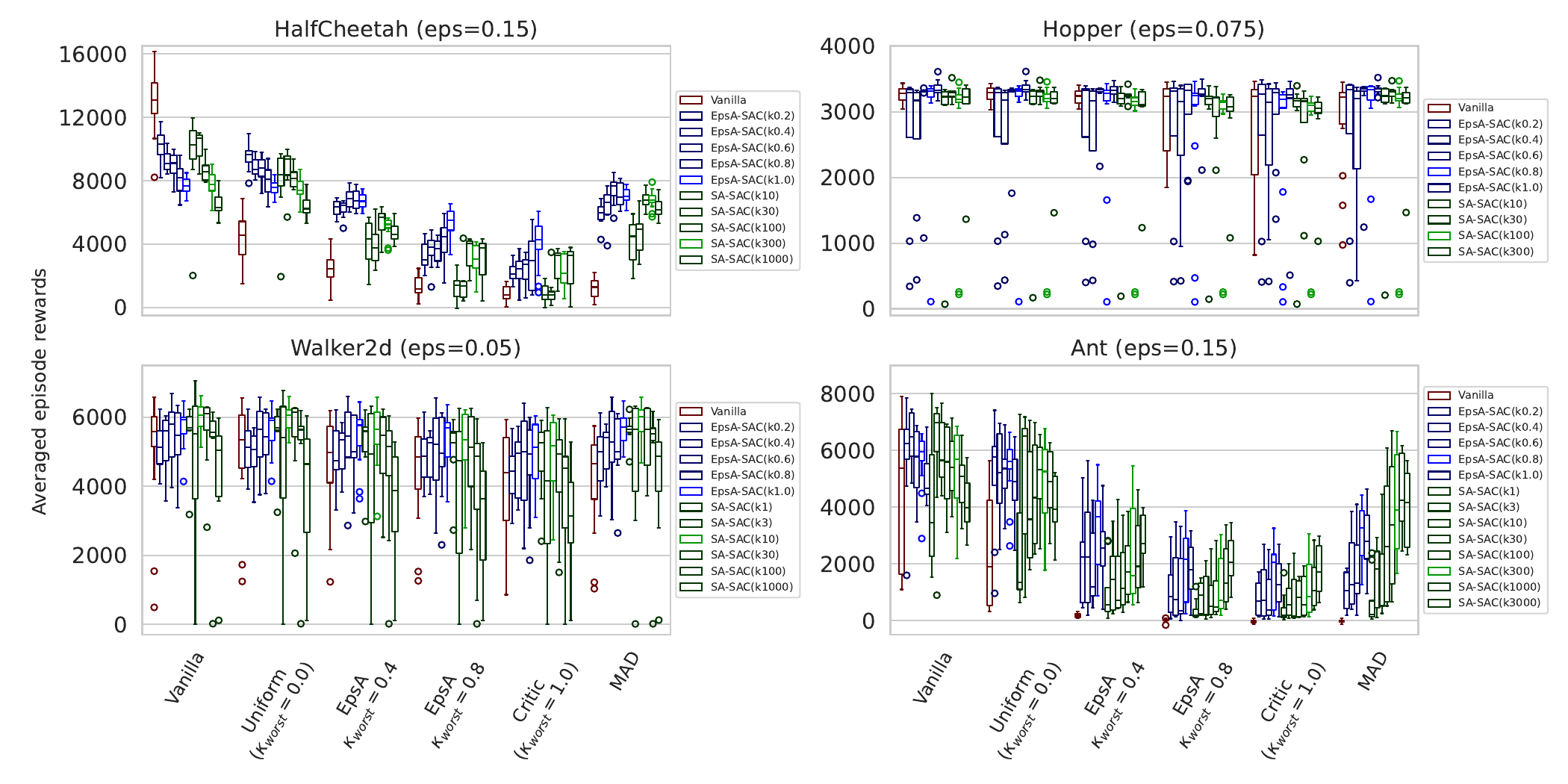}
\vspace{-5truemm}
\caption{
Box plots represent the performance of EpsA-SAC and SA-SAC under $L_\infty$-norm attacks across trainings with various coefficient parameters, 
including $\kappa_{worst}$ for EpsA-SAC and $\kappa_{reg}$ for SA-SAC. 
We use the same colors for the main settings: Vanilla (brown), EpsA-SAC (blue), and SA-SAC (green). 
Variants of EpsA-SAC and SA-SAC with different parameters are represented by dark blue and dark green, respectively.
}
\label{fig_appendix_epsa_coefficient_boxplot}
\end{figure}

Fig. \ref{fig_appendix_epsa_coefficient_boxplot} displays evaluation scores from trainings with various coefficient parameters, $\kappa_{worst}$ for EpsA-SAC and $\kappa_{reg}$ for SA-SAC. 
As discussed in Appendix \ref{subsecApp_SoftA_additional_results_softa_coefficient}, 
the trade-off between performance under attack-free evaluation and robustness is evident in HalfCheetah. 
In Ant, however, certain parameters optimally balance regularization, stability, robustness, and performance without attacks, 
demonstrating a sweet spot.

\section{Hyperparameter and Settings}
\label{secApp_hyperparams_settings}

In this section, we describe experiments settings and hyperparameter 
that are used in Section \ref{secExperiments}, Appendix \ref{secApp_SofA_additional_results}, and \ref{secApp_Epsilon_additional_results}.
We use SAC \citep{haarnoja2018softa,haarnoja2018softb} as the base algorithm. 
All variants, including SofA-SAC, EpsA-SAC, SA-SAC, and WocaR-SAC, 
generally adhere to the settings and parameters outlined in Appendix \ref{subsecApp_SAC_settings}, 
except for their specific adjustments.

\subsection{Vanilla SAC (Common settings)}
\label{subsecApp_SAC_settings}

Almost all settings follow those in the second SAC paper \citep{haarnoja2018softb}; 
however, it is important to note that we added procedures and modified parameters 
to ensure stable evaluation results across all four tasks. 
Specifically, we set the training steps at 1M for Hopper, 2M for HalfCheetah, 2M for Walker2d, and 3M for Ant. 
Table \ref{tab:sac_hyperparameters} lists all other settings, 
with underlined parts indicating the additions or modifications we made to achieve stable experimental outcomes.
We explain the reasons for the modifications made, as detailed below.

\paragraph{Target Entropy for Hopper-v2}

We found that the target entropy 
$\mathcal{H}_{\text{ent}} = -\text{dim}|\mathcal{A}| = -3$, 
as used for Hopper-v2 in the original SAC \citep{haarnoja2018softb}, 
is too low to get stable final results. 
Consequently, scores during training oscillated drastically, 
a phenomenon observed not only in our SAC implementation but also in a well-known open-source implementation \citep{stable-baselines3}.
Though a fixed temperature parameter $\alpha_{ent}=0.2$, as in the first SAC paper \citep{haarnoja2018softa}, 
or decreasing learning rate for $\alpha_{ent}$ also work well, 
we decide to tune the target entropy only for Hopper-v2 so as to keep stable final results.

\paragraph{Prioritized Experience Replay}
We use Prioritized Experience Replay (PER) \citep{schaul2015prioritized} to accelerate learning outcomes. 
While its effect is only slightly better in some tasks, 
we do not observe any disadvantages to using PER. 
Therefore, we decide to continue to use this method.

\paragraph{Normalizer}

We normalize state inputs by recording running statistics as the agent receives information from environments. 
Subsequently, when the SAC agent samples a mini-batch for learning or calculating output from observation inputs, 
we use the state normalizer to set the mean and standard deviation to 0.0 and 1.0, respectively.

We attempt to normalize rewards by episodic returns as SA-PPO \citep{zhang2020robust} and other recent on-policy methods \citep{zhang2021robust,oikarinen2021robust, sun2021strongest,liang2022efficient} did. 
However, we observe that the agent's learning became critically slow in HalfCheetah due to delays in updating the running statistics. 
Therefore, we decided to abandon this normalizer.

\begin{table}
\caption{Common SAC Hyperparameter Settings}
\label{tab:sac_hyperparameters}
\begin{tabular}{@{}lc@{}}
\toprule
Hyperparameter & Value \\ \midrule
Optimizer & Adam \citep{kingma2014adam} \\
Learning Rate(Actor, Critic, $\alpha_{ent}$) & 0.0003 \\
Discount Factor ($\gamma$) & 0.99 \\
Target Smoothing Coefficient ($\tau$) & 0.005 \\
Optimization Interval per Steps & 1 \\
Number of Nodes & 256 \\
Number of Hidden Layers & 2 \\
Activation Function & ReLU \\
\underline{Prioritized Experience Replay ($\beta_{per}$)}  & \underline{start from 0.4 to 1.0 at the end} \\
Replay Buffer Size & 1,000,000 \\
Batch Size & 256 \\
Temperature Parameter ($\alpha_{ent}$) & Auto-tuning \\
Target Entropy ($\mathcal{H}_{target}$) & -dim$|\mathcal{A}|$, \underline{only for Hopper-v2: $\mathcal{H}_{target}=0.2$} \\
Normalizer  & \underline{state normalizer only} \\
\bottomrule
\end{tabular}
\end{table}

\subsection{SofA-SAC settings}
\label{subsecApp_SofA-SAC_settings}

SofA-SAC utilizes temperature parameter $\alpha_{\text{attk}}$, 
which determines the strength of the adversary, 
and a sample approximation number $N$ as its unique parameters. 
Typically, we set $N=64$ during Q-update and policy improvement 
except for ablation studies. 
We find that setting $\alpha_{\text{ent}}$ from 4 to $\infty$ works well across four tasks 
without significantly compromising performance under attack-free evaluation, except in HalfCheetah.
In Hopper and Ant, 
starting training with $\alpha_{\text{attk}}=8$ 
and gradually decreasing it to $\alpha_{\text{attk}}=4$ 
results in stable scores and enhanced robustness. 
This approach likely prevents pessimistic exploration during the initial training phase. 

When scheduling $\alpha_{\text{attk}}$, 
we maintain the initial value until $t=2.5 \cdot 10^5$ steps, 
then linearly decrease it to reach the final target parameter at $t = \text{training-steps} - 2\cdot10^5 $. 
Subsequently, we maintain this final value until the end of the training.
Consequently, we denote the parameters in the legends of the graphs as '8to4' or '4to4' to clarify for the readers.

To avoid unintended unstable training in Walker2d-v2, 
we implement two special measures to exclude outlier values that could lead to divergence. 
The first measure involves percentile clipping, 
which restricts the next target value to the range between the 20\% percentile and the 80\% percentile across the perturbed next-state samples. 
The second measure is to disregard the importance weight during policy improvement, 
which, while potentially beneficial for incorporating the adversary, 
tends to amplify the impact of outlier values.

\subsection{EpsA-SAC settings}
\label{subsecApp_EpsA-SAC_settings}

EpsA-SAC combines uniform perturbations with an approximation of the worst attack using the mixture coefficient $\kappa_{\text{worst}}$. 
We employ Projected Gradient Descent (PGD) to approximate the worst attack, 
similarly to the approach used in the Critic attack \citep{pattanaik2017robust,zhang2020robust}, 
as detailed in Section \ref{subsubsec_epsa_rl}. 
Since estimating the worst-case states that consider the entropy term of SAC is challenging, 
we employ two approximations: (1) using the mean of the action outputs, and (2) ignoring the entropy term. As the SAC-based algorithm's Critic comprises two networks, we average their outputs for gradient calculations.

To balance computational resources and performance, 
we determine that five gradient steps with a random start during PGD suffice for our methods. 
To mitigate poor exploration during the initial training phase, 
we start $\kappa_{\text{worst}}=0$ and linearly increase it to the target value by $t=\text{training-steps} - 2\cdot10^5$, 
maintaining this level until the end of the training.

\subsection{SA-SAC settings}
\label{subsecApp_SA-SAC_settings}

Originally, SA-MDP \citep{zhang2020robust} was implemented for PPO, A2C, DDPG, and DQN. 
We have adapted this to create a SAC version, termed SA-SAC, by appropriately extracting key elements from both SA-PPO and SA-DDPG.
SA-SAC updates its policy with action consistency terms to account for cases where the attacker manipulates observations to induce actions that differ from the original ones. 
This loss function for policy is represented as:
\begin{equation}
\label{eq_app_sa-sac_loss_for_policy}
L(\pi) = 
\mathbb{E}_{s \sim D(\cdot)}
\left[
  \mathbb{E}_{a \sim \pi}
  \left[
  \alpha_{ent} \log \pi (a|s) - Q^{\pi}(s, a)
  \right]
  + \kappa_{reg} 
  \max_{\tilde{s}\in \mathcal{B}_{\epsilon_{p}}} D_{KL}(\pi(\cdot | s) \parallel \pi(\cdot | \tilde{s} ))
\right],
\end{equation}
where $\kappa_{reg}$ is a coefficient term to balance the original SAC loss and consistency of actions.
To solve the maximization term in Eq. \eqref{eq_app_sa-sac_loss_for_policy}, 
there are two versions for computing: 
one using convex relaxation and the other using Stochastic Gradient Langevin Dynamics (SGLD). 
We adopt SGLD due to the higher scores reported in SA-PPO \citep{zhang2020robust}. 
We carefully extract the essence from SA-DDPG and SA-PPO and set the configurations as follows: 
The attack scale increases linearly from 0 to the final target value according to a predefined schedule. 
This increase starts from step $t=2.5\cdot10^5$ and reaches the final value at step $t=\text{training-steps} - 2\cdot10^5$. 
Additionally, we use five gradient steps in our calculations. 
Given the trade-off between robustness and performance under attack-free evaluation, as reported in \cite{liang2022efficient}, 
we conduct a comprehensive parameter search for the coefficient term $\kappa_{\text{reg}}$ among the values $\{1,3,10,30,100,300,1000\}$.


\subsection{WocaR-SAC settings}
\label{subsecApp_WocaR-SAC_settings}

WocaR \citep{liang2022efficient} is originally implemented only for PPO, A2C, and DQN. 
We have adapted the essence of this method into our SAC implementation.
In WocaR, 
the agent learns an additional action-value function and its target network, 
which estimate the worst-case scenario induced by the policy's action.
This action-value function for the worst-case scenario is represented as the Bellman equation 
associated with the Worst-Case Bellman operator $\underline{\mathcal{T}}$ as:
\begin{equation}
\label{eq_app_wocarsac_bellman_for_worstq}
(\underline{\mathcal{T}} Q ) (s_t,a_t) 
= 
r(s_t,a_t) + 
\gamma
\mathbb{E}_{s_{t+1} \sim \mathcal{F}}
\left[
  \min_{\tilde{s}_{t+1} \in \mathcal{B}_{\epsilon}}
    Q(s_{t+1}, \mu(\tilde{s}_{t+1}))
\right].
\end{equation}
To calculate the next worst-case action value during updates practically, 
they calculate action bounds using convex relaxation (IBP) \citep{gowal2018effectiveness} 
and determine the worst action using PGD within these bounds. 
We employ \textit{auto\_LiRPA} \citep{xu2020automatic} for the convex relaxation 
and perform five iterations of PGD to ensure efficient computation within realistic time constraints.

During policy improvement, 
they enhance the gradient ascent objective of the policy by incorporating the worst-case action value, 
aiming to balance performance under attack-free evaluation with robustness in the worst-case attack scenario. 
For our SAC implementation, we implement two methods: 
mixing the worst action value with the analytical target value (soft-max of the Q) as:
\begin{equation}
\label{eq_app_wocarsac_loss_for_sac_policy1}
L(\pi) = 
\mathbb{E}_{s \sim D(\cdot)}
\left[
  \mathbb{E}_{a \sim \pi}
  \left[
  \alpha_{ent} \log \pi (a|s) - \left( 
    (1 - \kappa_{worst}) Q^{\pi}(s, a)  + \kappa_{worst} \underline{Q}^{\pi}(s, a)
    \right)
  \right]
\right],
\end{equation}
 combining DDPG's policy loss with the worst action-value function by using the mean of the action, $\mu(s)$, as:
\begin{equation}
\label{eq_app_wocarsac_loss_for_sac_policy2}
L(\pi_{\phi}) = 
\mathbb{E}_{s \sim D(\cdot)}
\left[
  (1 - \kappa_{worst}) 
  \mathbb{E}_{a \sim \pi}
  \left[
  \alpha_{ent} \log \pi (a|s) - 
      Q^{\pi}(s, a)
  \right]
  - \kappa_{worst}
    \underline{Q}^{\pi}(s, \mu(s))
\right].
\end{equation}

To mitigate poor exploration and pessimistic learning behaviors, 
they schedule both the attack scale and the mixture rate of the worst-case objective throughout the policy optimization process. 
We initiate the attack scale at 0.0 and maintain it until $2.5 \cdot 10^5$ steps, 
then linearly increase it to the final scale by $t = \text{training-steps} - 2 \cdot 10^5$, 
maintaining this value until the end of the training. 
Similarly, we schedule $\kappa_{worst}$ to increase from 0.0 to the final target value, 
testing among \{0.2, 0.4, 0.6\}.

Both methods in Eq. \eqref{eq_app_wocarsac_loss_for_sac_policy1} and Eq. \eqref{eq_app_wocarsac_loss_for_sac_policy2} 
learn well in smaller tasks (Pendulum, InvertedPendulum) and at the initial stages of all four tasks. 
However, as the attack scale increases, 
the performance drastically declines, leading to the collapse of training, even when we use more tighter bound method \citep{zhang2019towards} than IBP used in WocaR. 
We hypothesize that this is because the policy fails to learn from the worst-case scenarios due to the absence of adversaries 
during policy improvement. 
Consequently, the policy cannot produce valid actions for the next target action value, 
causing the worst action value to estimate increasingly poorer values as the attack scale enlarges.

A similar phenomenon is observed in the ablation experiment of EpsA-SAC in Ant-v2, 
which lacks an adversary during policy improvement. 
Therefore, our findings indicate that directly applying WocaR techniques to off-policy RL is challenging. 

\subsection{Evaluation settings}
\label{subsecApp_Eval_settings}

SAC requires more computational resources (GPUs) compared to PPO. 
Therefore, for evaluation, 
we plot the mean of 20 episodic returns, 
using data from sixteen different training seeds for the main result parameters (Fig. \ref{fig_main_softa_boxplot}, Fig. \ref{fig_main_epsa_boxplot}). 
For other parameters and ablation studies, 
we conducted trainings with eight different seeds for each setting and omitted the learning type attacker evaluations (SA-RL/PA-AD). 
For attacker learning methods (SA-RL/PA-AD), 
we employ the same state normalizer learned by the agent to normalize the attacker scale. 
Although this approach is somewhat akin to white-box attack settings, 
we ensure that these settings are treated consistently, 
thus guaranteeing that the results are evaluated in a uniform manner.

We use box plots to represent the average evaluation scores, mitigating the variance across different training seeds. 
The box in each plot shows the 25\%, 50\% (median), and 75\% percentile values, 
while the whiskers extend to the minimum and maximum scores within the range that is not considered outliers. 
Outliers are depicted as hollow circles and are defined as values beyond $1.5 \times \text{IQR}$ from the quartiles, 
where $\text{IQR}$ (Interquartile Range) is the difference between the 75\% and 25\% percentiles.


\paragraph{Max Action Deifference (MAD) attack}

We basically follow as SA-PPO and SA-DDPG \citep{zhang2020robust}. 
We utilize SGLD and confirm that ten gradient iterations after random initial start are enough for the evaluations. 

\paragraph{Critic attack}

We apply the modified version of Critic attack \citep{zhang2020robust}, rather than the one originally proposed in \citet{pattanaik2017robust}. 
We utilize Projected Gradient Decent (PGD) and confirm that ten gradient iterations after random initial start are enough for the evaluations.
As we mentioned in Appendix \ref{subsecApp_EpsA-SAC_settings}, SAC's Critic comprises two networks, 
then we use the mean of these two action values to calculate the gradient descent.

\paragraph{SA-RL / PA-AD}

Because the network structure and the procedures (especially normalization) of SAC are quite different from PPO, 
we implement the SA-RL \citep{zhang2021robust} and PA-AD \mbox{\citep{sun2021strongest}} adversaries as the SAC agent.
We use the same hyperparameter and settings as detailed in Appendix \ref{subsecApp_SAC_settings}, 
with the exception of flipping the reward term $r\rightarrow-r$ and 
setting the target entropy as $\mathcal{H}_{target}=-dim|\mathcal{S}|$ only for SA-RL.
We reuse the same state normalizer that was learned for the policy to simplify the process 
and reduce additional variance terms when learning adversaries.

\paragraph{Soft Worst Attack (SofA)}

As described in Appendix \ref{subsecApp_softa_procedure} and Algorithm \ref{alg:softa-sampling}, 
we sample 64 perturbed states in parallel from the prior distribution (Gaussian distribution in this study). 
Subsequently, we calculate the probability weights for these samples. 
We then select one sample based on these probability weights.
When we set $\alpha_{\text{attk}} \rightarrow 0$, 
we deterministically choose the sample that has the maximum probability weight.

\paragraph{Epsilon Worst Attack (EpsA)}

As described in Appendix \ref{subsecApp_epsa_procedure} and Algorithm \ref{alg:epsa-sampling}, 
we employ the uniform distribution as the prior distribution and use the Critic attack \citep{pattanaik2017robust,zhang2020robust} for approximating the worst-case attack. 
We utilize PGD for gradient descent and adopt ten gradient iterations, 
consistent with the methodology used in the Critic attack evaluation.

\section{Computer Resources}
\label{secApp_computer_resources}


To clarify the time and resources required to calculate our algorithms and other baselines, 
we present Table \ref{tab:computer_resoures}, which shows the total training time for each algorithm. 
To ensure fair comparison, we use \textit{Tesla V100 32GB} GPUs for all the trainings.

\begin{table}
\centering
\caption{Computation times for each method}
\label{tab:computer_resoures}
\begin{tabular}{c|ccccc}
\toprule
\multirow{2}{*}{Model} 
& Hopper & HalfCheetah & Ant \\
& \multicolumn{3}{c}{
 Time (hour)
} \\

\midrule
Vanilla-SAC & 7.2 & 16.7 & 23.4 &\\
SA-SAC & 9.3 & 20.6 & 34.7 &\\
SofA-SAC & 12.1  & 23.5 & 32.6 &\\
EpsA-SAC & 17.2 & 35.6 & 53.8 &\\
\bottomrule
\end{tabular}
\end{table}

\section{Potential Impact}
\label{secApp_potential_impact}

Our research primarily focuses on proposing robust RL methods and evaluation techniques, 
which we believe will significantly advance the practical applications of RL. 
However, it is important to note that robustness methods, including those proposed in our methods, 
require additional computational resources compared to the original RL algorithms. 
Consequently, utilizing these methods in scenarios where robustness is not a critical requirement may result in increased computational costs. 
Furthermore, as with RL in general, there exists a potential risk of misuse in applications involving automation and optimization. 
Nonetheless, we believe that the positive utility of assisting humanity in creating a better society far outweighs these concerns.

\end{document}